\documentclass{article}
\usepackage[preprint]{corl_2026} 
\usepackage{times}
\usepackage{multicol}
\usepackage{graphicx}
\usepackage{algorithm,algorithmic}
\usepackage{amsfonts,amsmath,amssymb}

\usepackage{booktabs}       
\usepackage{nicefrac}       
\usepackage{microtype}      
\usepackage[table]{xcolor}
\usepackage{acronym}
\usepackage{xspace}
\usepackage{enumitem}
\usepackage{tcolorbox}
\usepackage{multirow}
\usepackage{float}
\usepackage{titletoc}
\usepackage{makecell}
\tcbuselibrary{skins}
\usepackage{caption}
\usepackage{minitoc}
\usepackage{etoc}
\etocsettocstyle{\section*{Contents}}{}

\usepackage{xcolor}
\usepackage{url}           
\usepackage[export]{adjustbox}

\usepackage{subfiles}
\usepackage{subcaption}

\usepackage{pifont}

\definecolor{bestgait}{RGB}{205,215,194}   
\definecolor{besttrack}{RGB}{188,207,224} 

\usepackage{xcolor}
\definecolor{hotpink}{RGB}{255, 105, 180} 

\usepackage{hyperref}
\hypersetup{
    colorlinks=true,
    allcolors=hotpink,
}
\definecolor{mypurple}{RGB}{148,103,189} 

\def\sname{GaitSpan}

\begin{document}

\newcommand{\papertitle}{%
GaitSpan: Growing Humanoid Locomotion from Walking to Running
}

\title{\papertitle}
\author{
  Kwan-Yee Lin {\thanks{Equal contribution and joint first authorship}}\\
  University of Michigan\\
  \texttt{junyilin@umich.edu}
  \And
  Zilin Wang$^{\color{pink}{*}}$\\
  University of Michigan\\
  \texttt{zilinwan@umich.edu}
  \AND
  Janelle J. Liu\\
  Skyline High School\\
  \texttt{liujanellej@gmail.com}
  \And
  Stella X. Yu\\
  University of Michigan, UC Berkeley\\
  \texttt{stellayu@umich.edu}
}

\maketitle

\etocdepthtag.toc{mtchapter}
\etocsettagdepth{mtchapter}{subsection}
\etocsettagdepth{mtappendix}{none}
\faketableofcontents

\begin{abstract} 
A humanoid that can walk should not relearn locomotion from scratch to jog or run. Yet
current approaches often obtain gait diversity by prescribing gait schedules, imitating motion clips, training experts to switch between or distilling skills into one policy. These strategies can produce impressive behaviors, but offer limited flexibility across continuous speed commands, terrains, and morphologies. 
We study skill growth with \textbf{\sname}, a framework that expands a pretrained, basic {\it walking} policy into faster locomotion. It treats {\it walking} as a seed skill: reusable motor structure for balance, support, body coordination, and contact transition that can be regenerated at new rhythms, extended into longer/higher strides, and corrected by residual adaptation. This expansion has three aspects: {\bf 1) rhythm generation}, which modulates the frozen {\it walking} policy with multiple internal clocks and learns command-conditioned combinations of the resulting canonical actions; {\bf 2) stride shaping}, which rewards dynamic locomotion patterns appropriate for higher commanded speeds using a physically grounded objective inspired by spring-loaded inverted pendulum dynamics; and {\bf 3) residual adaptation}, which captures motion details not accounted for by rhythm generation or stride shaping. 
\sname{} is the first to deliver a single command-conditioned humanoid policy that spans {\it walking}, {\it jogging}, and {\it running}-like regimes covering a continuous speed range, transfers across morphologies, and deploys zero-shot on unseen sim-to-sim, and real-world terrains. Compared with baselines either trained with multi-experts or imitation from humans, it learns faster and achieves stronger gait performance. These results make {\it walking}-to-{\it jogging}-to-{\it running} a concrete demonstration of skill growth in humanoid learning, where existing skills serve not only as endpoints, but as seed priors for behavior families, pointing to a higher-performing, more efficient and scalable route for humanoid skill acquisition. Videos are available at~\url{https://gaitspan2026.github.io/}.
\end{abstract}


\def\figTeaser{
\begin{figure}[t]\centering
 \includegraphics[width=1\linewidth]{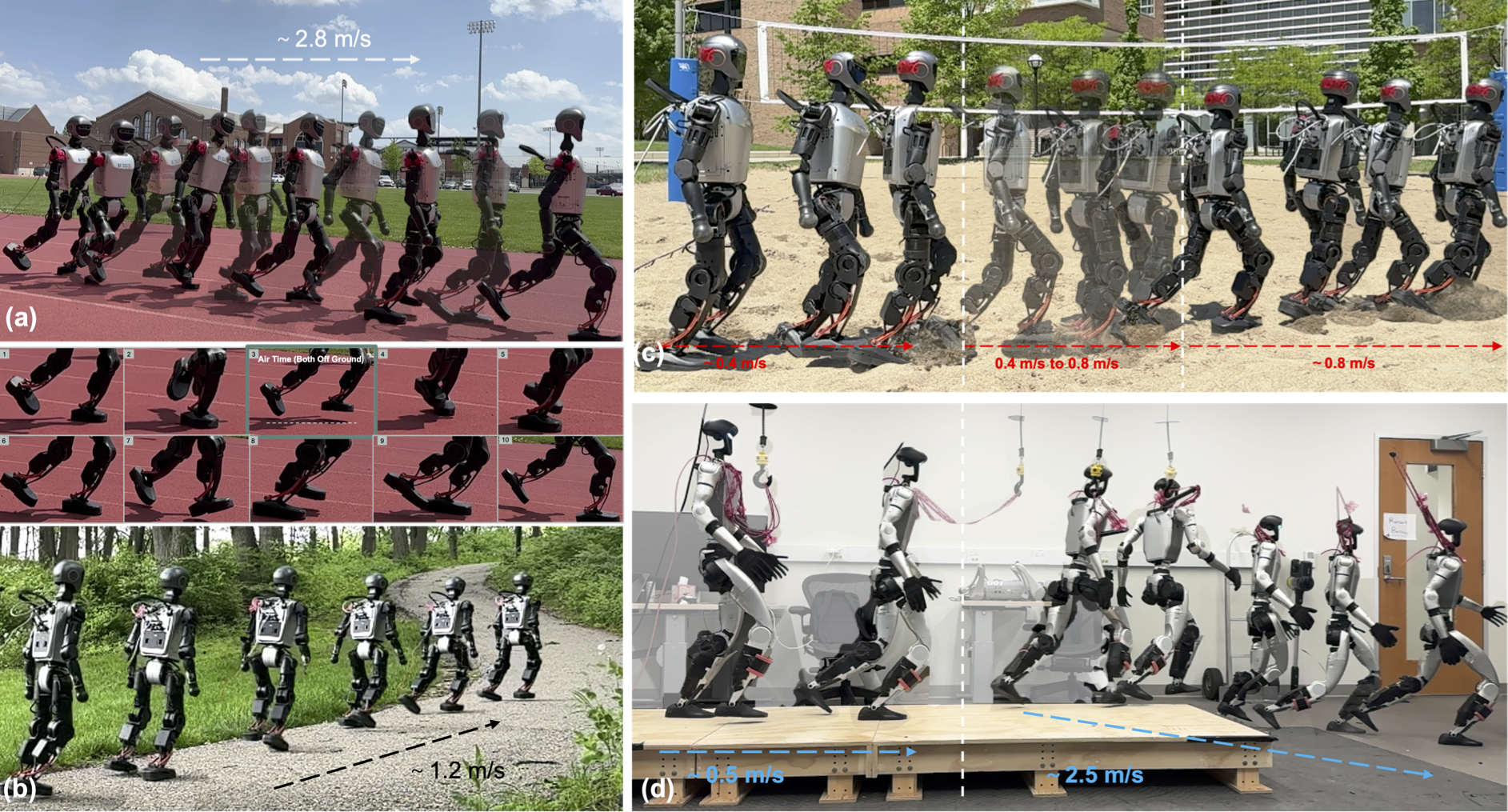}
    \caption{{\textbf{GaitSpan grows gaits from walking.} It demonstrates a single policy can produce diverse gaits such as \textbf{a)} running and \textbf{b)} jogging on outdoor terrain, \textbf{c)} walking-speed transitions on sand, and \textbf{(d)} walking-to-running transitions. Results demonstrate robustness across terrains and generalization across embodiments.}}
   \label{fig:teaser}
\end{figure}
}

\section{Introduction}

\figTeaser

Locomotion is how a body carries itself through the world by regulating contact, balance, and momentum. In humans, {\it walking}, {\it jogging}, and {\it running} are related regimes of a growing motor repertoire. As speed increases, a person does not discard {\it walking} and learn {\it running} from scratch. They reuse motor structure for balance, support, center-of-mass shifting, foot placement, body coordination, and contact transition, while changing rhythm, stride, contact timing, and foot clearance. This paper studies how a humanoid that has learned to {\it walk} can expand that acquired skill into broader locomotion regimes such as {\it jogging} and {\it running}, without learning each gait independently.

This skill-growth view contrasts with the current state of humanoid learning. Recent systems show impressive behaviors, like {\it fast locomotion}~\cite{DBLP:journals/corr/abs-2409-16611}, {\it athletic motions}~\cite{he2025asap}, {\it table tennis}~\cite{DBLP:journals/corr/abs-2508-21043,DBLP:conf/humanoids/HuangBKS016}, {\it martial arts}~\cite{xie2025kungfubot,han2025kungfubot2}, and {\it choreographed dances}~\cite{DBLP:journals/corr/abs-2511-04131}. Yet these behaviors are usually trained, optimized, or demonstrated as separate competencies. A humanoid that learns one athletic motion does not necessarily acquire command-conditioned locomotion across {\it walking}, {\it jogging}, and {\it running}. Even when humanoid {\it running} is demonstrated, the underlying models are often unavailable, and the behavior may be trained for a narrow speed range or task setting. What is missing is not impressive motion, but a learning principle by which one acquired skill can expand into a family of related behaviors.

Existing approaches offer different routes toward adaptive humanoid locomotion, but none show how a single learned humanoid policy can span {\it walking}, {\it jogging}, and {\it running} as the commanded speed rises.  Model-based control provides structure and robustness, but often depends on hand-designed templates, phase variables, contact schedules, and morphology-specific controllers~\cite{DBLP:journals/ral/SovuklukEO24,DBLP:conf/icra/ParkKLP25}. Imitation learning can reproduce rich motions from human or motion-capture data, but the learned policy is tied to the demonstrated distribution and may not generalize reliably across terrains, embodiments, or command ranges\cite{2021-TOG-AMP,DBLP:journals/corr/abs-2508-08241}. Reinforcement learning is more flexible across terrain and embodiment, but diverse-speed locomotion remains hard to obtain: Reward terms that stabilize {\it walking} may suppress {\it jogging} or {\it running}, while rewards for dynamic motion are difficult to craft, balance, and interpret as speed increases~\cite{Amazon_FAR_and_Abbeel_Holosoma}. A common workaround is to train discrete skills or experts and then distill or switch among them in one policy. This is inefficient for a continuous speed range and can introduce cross-skill interference, averaged behaviors near regime boundaries, and brittle transitions between gait styles. No published open model or codebase demonstrates a single learned humanoid policy that covers {\it walking}, {\it jogging}, and {\it running} as a continuous command-conditioned family.

Our goal is not to prescribe, imitate, distill, or switch among separate gaits, but to expand {\it walking} into faster locomotion (as shown in Fig~\ref{fig:teaser}). The premise is simple: {\it walking} can serve as a seed skill. A humanoid that walks has a reusable motor structure for balance, support, body coordination, and contact transition. To move faster, it should not relearn them from scratch. It should regenerate them at new rhythms, extend them into longer strides, and add residual corrections where generated motion falls short.

We introduce \textbf{\sname{}}, a framework that expands a pretrained {\it walking} policy into faster locomotion.  Starting from a low-speed {\it walking} policy, \sname{} grows a broader gait family through three aspects.
{\bf 1) Rhythm generation.} \sname{} modulates the frozen {\it walking} policy with multiple internal clocks. Querying the policy at different temporal scales produces a family of walking-derived actions, and command-conditioned coefficients learn how to combine them so cadence, stride timing, direction, and speed vary continuously with the command.
{\bf 2) Stride shaping.} \sname{} rewards dynamic locomotion patterns appropriate for higher commanded speeds. A physically grounded objective inspired by spring-loaded inverted pendulum dynamics~\cite{DBLP:conf/iros/WensingO13} shapes compression, rebound, touchdown, and flight events, encouraging the policy to lengthen stride and enter {\it jogging}- and {\it running}-like regimes as speed increases.
{\bf 3) Residual adaptation.} A residual policy captures motion details not accounted for by rhythm generation or stride shaping, resulting in a comprehensive policy.

\sname{} is the first to deliver a single command-conditioned humanoid policy that spans {\it walking}, {\it jogging}, and {\it running}-like regimes across five embodiments. It covers a continuous speed range, transfers across morphologies, and deploys zero-shot on diverse unseen real-world terrains, despite training only on flat and mildly uneven terrains. Compared with various baselines, \sname{} learns diverse speed-related motions and achieves stronger gait performance, demonstrating that existing skills need not remain endpoints, but can serve as generative priors for families of related behaviors.  This points to a higher-performing, more efficient, and more scalable route for humanoid skill acquisition.

\section{Related Work}
\label{related_work}
\noindent\textbf{Humanoid Locomotion.} 
Practical locomotion is a fundamental and long-standing challenge in humanoid robotics, requiring stable balance, whole-body coordination, and motion adaptation across commands, terrains, and environments. Recent years have witnessed remarkable progress in both academia and industry~\cite{unitree_robot_running_10ms,boston_dynamics_atlas_walk_run_crawl}, with increasingly agile walking, running, and whole-body behaviors being demonstrated. Yet, achieving a single controller that remains stable and adaptable across diverse speed regimes remains difficult.

Early progress in humanoid locomotion was largely driven by classical control theories for legged robots~\cite{raibert1986legged}, which achieved robust walking and running through model-based optimization and carefully designed heuristics. Representative examples include ZMP-based control for dynamic balance~\cite{DBLP:journals/ijrr/Goswami99}, linear inverted pendulum (LIP) models for walking pattern generation~\cite{kajita20013d}, spring-loaded inverted pendulum (SLIP) models for high-speed legged locomotion like running~\cite{DBLP:journals/siamrev/HolmesFKG06,DBLP:journals/tac/PoulakakisG09}, and trajectory optimization or whole-body control frameworks built upon such models~\cite{kuindersma2016optimization,ding2022orientation,khazoom2024tailoring,DBLP:conf/icra/SovuklukSEO25,DBLP:conf/iros/Dosunmu-OgunbiS24}. The literature is vast (we refer to~\cite{Humanoid_Locomotion_Survey,DBLP:journals/corr/abs-2501-02116} for comprehensive survey), shared characteristics behind these approaches are that they encode strong structural priors for stability; however, they typically rely on explicit model assumptions for dynamics, rigid gait schedules, or task-specific controller design, which can limit flexibility when adapting continuously across locomotion speeds and conditions.

Learning-based approaches have demonstrated impressive capabilities for humanoid robots, enabling more agile motion behaviors that were previously difficult to achieve with hand-designed controllers. 
Recent reinforcement learning (RL) works fall broadly into the categories of narrowing the sim-to-real gap (\textit{e.g.,} domain randomization~\cite{radosavovic2024real}, motor adaptation upon teacher-student distilling~\cite{DBLP:conf/iros/KumarLZPSM22,gu-RSS-24}, and training RL wrt real-world actuator feedback~\cite{singh2023learning}), 
and reward shaping for specific locomotion skills (\textit{e.g.,} walking, running, and parkour~\cite{siekmann2021sim,van2024revisiting,Amazon_FAR_and_Abbeel_Holosoma,crowley2023optimizing,zhuang2024humanoid}). These methods typically treat different gaits as distinct skills, requiring either separate policies or multi-stage training and distilling pipelines to induce the emergence of a family of locomotion skills.  Although impressive behaviors can be obtained, the relation among skill regimes remains largely implicit. As a result, motion transitions can be fragile and difficult to control, particularly under high-speed contacts and whole-body dynamics. Recent work~\cite{DBLP:journals/corr/abs-2509-19573} incorporates control-theoretic ideas and optimized dynamic reference trajectories into RL reward design, aiming to combine the robustness of model-based control with the flexibility of learning. However, its reliance on reference trajectories still prescribes the target motion structure. Research on imitation learning (IL) leverages human motion data to bypass sophisticated controller design and reward engineering, enabling humanoid robots to reproduce natural and expressive behaviors from demonstrations~\cite{tang2023humanmimic,radosavovic2024humanoid,cheng2024expressive,fu2024humanplus,li2025bfmzeropromptablebehavioralfoundation,luo2025sonic}. Yet, human motion is not a robot-native prior, as it may mismatch the robot's morphology, actuator limits, and energy landscape, often requiring careful retargeting and additional stabilization. Moreover, IL imports motion diversity from mimicking external demonstrations rather than explaining how a robot's own motor skill can develop. This dependence on demonstration data overlooks the adaptation and generalization where commands, previous actions, or environments move beyond the demonstrated distribution in robot's real-world executions.

Our work focuses on emergent whole-body locomotion transitions from a single policy, without gait labels, complex switching logic, or human demonstrations. We treat a learned walking controller as a seed prior, preserving its reusable locomotion structure while learning structured, physically guided deviations that grow it into more dynamic gait regimes. This combines robustness from the stable walking seed with flexibility from command-conditioned skill expansion.

\noindent\textbf{Curriculum and Continual Skill Learning in Robotics.} Curriculum learning~\cite{DBLP:conf/icml/BengioLCW09} is a widely used optimization strategy in RL for robotics, where training tasks are organized in a predefined order of complexity, typically progressing from easier to harder stages in terms of task objectives~\cite{DBLP:conf/icml/FlorensaHGA18}, command ranges~\cite{DBLP:conf/humanoids/PengBZ25,Amazon_FAR_and_Abbeel_Holosoma,DBLP:conf/corl/CuiLHQZHZTHNCJ24}, terrain complexity~\cite{DBLP:conf/cvpr/LinY25,DBLP:conf/icra/ChengSAP24}, or environmental difficulty~\cite{DBLP:conf/iros/WangXSX25}. These curricula stabilize exploration and thus improve training efficiency. However, they often require manual stage design, which can introduce stage-wise dependency, where failures or biases in early stages affect later learning. More importantly, curriculum learning does not inherently preserve earlier skills; without explicit anchoring or regularization, adapting to later stages may overwrite or weaken previously learned behaviors~\cite{DBLP:journals/jmlr/NarvekarPLSTS20}. Continual learning~\cite{DBLP:journals/ras/ThrunM95,DBLP:journals/inffus/LesortLSMFR20}, by contrast, studies how robots acquire, retain, and reuse knowledge over extended experience.  Existing approaches reuse prior knowledge in different ways (\textit{e.g.,} direct fine-tuning of pretrained policies for new locomotion~\cite{smith2021legged,11155209} or manipulation tasks ~\cite{DBLP:conf/corl/JulianSSLFH20}, reusing skill libraries or modular policies at inference time~\cite{AUDDY2023104427}, and learning latent skill priors that encode reusable behaviors into a continuous latent space for downstream RL~\cite{DBLP:conf/corl/PertschLL20}). Our work is related to continual learning's goal of reusing prior motor knowledge, but differs in the level at which reuse is formulated. Rather than treating walking as a policy to be fine-tuned, a module to be selected, or a behavior prior to be sampled from, \sname{} explicitly reuses the internal locomotion structures of a walking skill (\textit{e.g.,} rhythm and stability) and grows them into faster and more dynamic motion regimes. With the shared locomotion structure as the basis of expansion, it supports continuous transfer across gait regimes while improving generalization to broader speed commands, environments, and humanoid embodiments.

\section{Method}
\label{sec:method}

\begin{figure*}[!t]
    \centering
    \includegraphics[width=\linewidth]{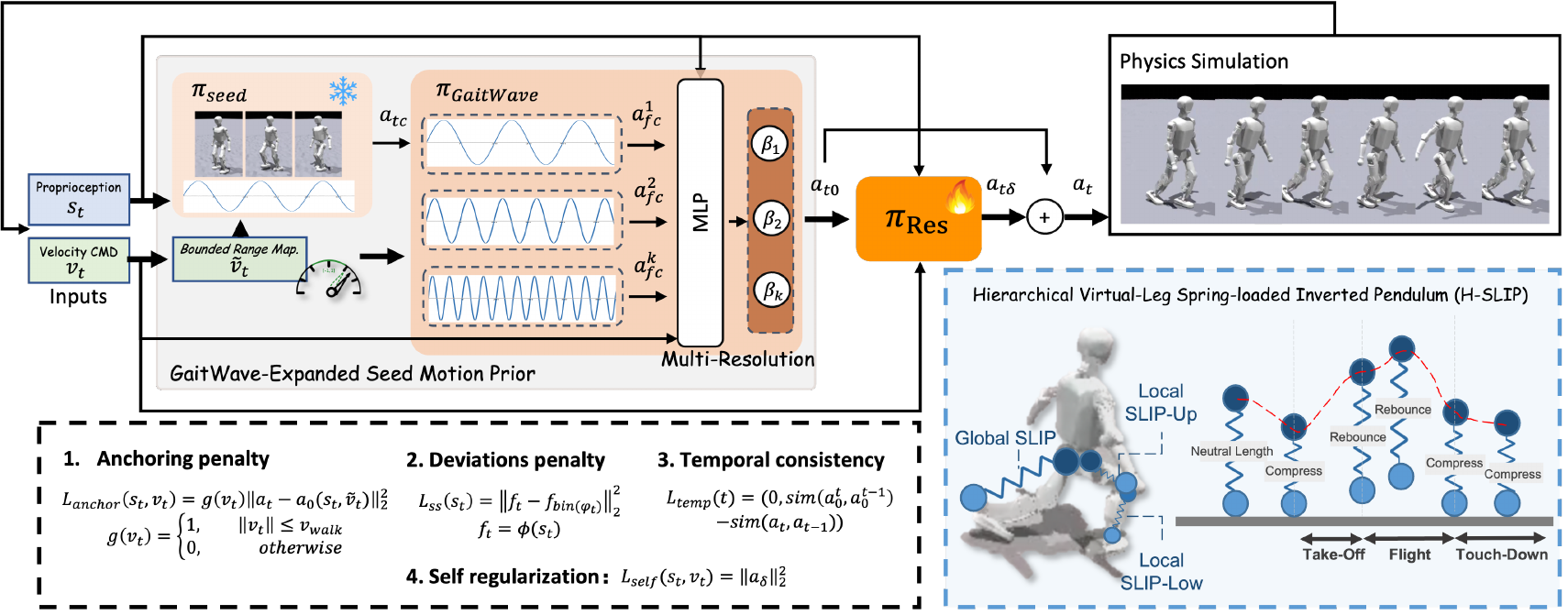}
    \caption{\textbf{Framework of GaitSpan}. Starting from a frozen walking policy as the seed, GaitWave grows rhythm through hierarchical composition of seed-derived action waves, while H-SLIP shapes dynamic stride events. A residual branch supplies additional adaptation, yielding a single policy spanning walking-, jogging-, and running-like regimes.}
 
    \label{fig:framework}
\end{figure*}

\textbf{Problem Formulation.} We formulate the humanoid gait emergence in this work as a whole-body skill expansion problem under varying velocity commands. At each time step $t$, the robot receives a proprioceptive observation
$\mathbf{s}_t \in \mathcal{S}$ and a velocity command
$\mathbf{v}_t \in \mathcal{V}$, and outputs a joint-level action
$\mathbf{a}_t \in \mathcal{A}$. Our goal is to learn a single command-conditioned policy $\pi_{\theta}$ that smoothly spans diverse gait regimes, such as walking- to jogging- and running-like motion, as well as the continuous transitions between them.  We assume access to a pretrained walking policy, denoted as the \emph{seed policy} $\pi_{\mathrm{seed}}$, which is frozen throughout training. The key challenge is to preserve the stable locomotion structure already captured by walking while allowing sufficient deviation for faster and more dynamic behaviors. To this end, GaitSpan expands this seed skill through a seed-residual adaptation architecture, where the frozen walking policy is treated not as a static base action as in conventional residual methods, but as a generator of reusable locomotion motor structure:
\begin{equation}
\begin{aligned}
    \mathbf{a}_t
    &=
        \underbrace{
        \mathcal{G}_{\psi}
        \left(
            \left\{
            \mathbf{a}^{\mathrm{seed}^{(k)}}_t
            \right\}_{k=0}^{K},
            \boldsymbol{\alpha}^{\psi}_t
        \right)
        }_{
        \displaystyle
        \mathbf{a}^{\mathrm{wave}}_t
        }
        +
        \mathbf{a}^{\mathrm{res}}_t \quad 
        \text{clipped to } (
        \mathbf{a}_{\min},
        \mathbf{a}_{\max})                                          \\
\end{aligned}
\label{eq:overall_action}
\end{equation}
Here, $    \mathbf{a}^{\mathrm{seed}^{(k)}}_t
    =
    \pi_{\mathrm{seed}}
    \left(
        \mathcal{T}_{k}(\mathbf{s}_t),
        \tilde{\mathbf{v}}_t
    \right),
    \boldsymbol{\alpha}^{\psi}_t
    =
    \operatorname{SoftBlend}
    \left(
        \mathcal{M}_{\psi}^{l},
        \mathbf{v}_t
    \right)$. $\mathcal{T}_{k}(\cdot)$ denotes a family of internal rhythm transformations applied when querying the same frozen seed policy $\pi_{\mathrm{seed}}$, $\mathcal{M}_{\psi}^{l}$ denotes a learnable hierarchical memory that produces command-conditioned composition coefficients, and $\mathcal{G}_{\psi}$ combines seed-derived action waves using learned coefficients. Thus, $\mathbf{a}^{\mathrm{wave}}_t$ is not generated by separate gait experts, but by expanding the robot's own seed skill through learnable composition of seed-derived action waves. The residual term $\mathbf{a}^{\mathrm{res}}_t$ captures additional adaptation beyond this structured expansion. To encourage physically meaningful dynamic regimes, we introduce \textit{H-SLIP}. It is baked into the training objectives as a relaxed, learning-compatible dynamic prior inspired by the classic SLIP model. The overall framework is illustrated in Fig~\ref{fig:framework}. We unfold the method details in the following subsections.

\subsection{Vanilla Architecture: Seed-Residual Adaptation}
\label{sec:canonical_residual}

We begin with the simplest formulation for growing locomotion from an acquired skill. Given the current proprioceptive observation $\mathbf{s}_t$ and commanded velocity $\mathbf{v}_t$, the seed branch produces $ \mathbf{a}^{\mathrm{seed}{(0)}}_t
    =
    \pi_{\mathrm{seed}}
    \left(
        \mathbf{s}_t,
        \tilde{\mathbf{v}}_t
    \right)$, where $\tilde{\mathbf{v}}_t=\mathcal{R}_{\mathrm{seed}}(\mathbf{v}_t)$ denotes a bounded-range mapping that projects the target command into the command domain covered by the pretrained walking policy. In practice, the planar velocity commands $v_x$ and $v_y$ are mapped to the normalized interval $[-1,1]$ expected by $\pi_{\mathrm{seed}}$. In this way, the seed branch always provides a stable, closed-loop walking action within its acquired locomotion range. To adapt the seed skill toward broader command regimes, we use a trainable residual policy $
    \mathbf{a}^{\mathrm{res}}_t
    =
    \delta_{\theta}
    \left(
        \mathbf{s}_t,
        \mathbf{v}_t,
        \mathbf{a}^{\mathrm{seed}}_t
    \right),$ which receives the original target command and predicts state-dependent corrections beyond the walking capability of the seed policy. The final action of the vanilla architecture is
$
    \mathbf{a}^{\mathrm{van}}_t
    =
    \operatorname{clip}
    \left(
        \mathbf{a}^{\mathrm{seed}}_t
        +
        \mathbf{a}^{\mathrm{res}}_t,
        \mathbf{a}_{\min},
        \mathbf{a}_{\max}
    \right).
$ Although this architecture reuses an acquired walking skill, it exposes a fundamental limitation. $\pi_{\mathrm{seed}}$ is queried only within its walking domain and nominal rhythm, so all growth toward dynamic regimes must be encoded by the residual, which is insufficient, as demonstrated in Fig.~\ref{fig:ablation}b in the experiment section.

\subsection{GaitWave: Hierarchical Rhythm Growth from Seed Walking}
\label{sec:gaitwave}

We therefore introduce \textit{GaitWave}, which grows the seed policy into a family of command-conditioned locomotion rhythms. It first constructs phase-scaled canonical action waves, and then composes them through a hierarchical multi-resolution coefficient memory bank.

\noindent\textbf{Phase-scaled Canonical Action Waves.}
The seed walking policy adopts a standard sinusoidal phase encoding \cite{DBLP:journals/ral/ShaoJLHWY22} in its observation~\cite{Amazon_FAR_and_Abbeel_Holosoma}, where $\phi_t$ denotes the locomotion phase. We reuse this existing phase input to query the same frozen seed policy under multiple phase scales.  We define a set of phase scale
$
    \mathcal{P}
    =
    \left\{
        \rho_0, \rho_2, \ldots, \rho_K
    \right\},$
where $\rho_0$ corresponds to the original walking phase and the remaining scales expose alternative rhythm structures inherited from the same policy. For each scale $\rho_k$, we construct a phase-modulated observation
$\mathbf{s}^{(k)}_t$ by replacing the original phase features with $
    \left(
        \sin \phi_t,
        \cos \phi_t
    \right)
    \longrightarrow
    \left(
        \sin (\rho_k \phi_t),
        \cos (\rho_k \phi_t)
    \right).$ The frozen seed policy is then queried at each phase scale to obtain the \emph{canonical action waves} $ \mathbf{a}^{\mathrm{seed}^{(k)}}_t$. These waves are not motion demonstrations, hand-designed gait templates, or separately trained experts. They are self-generated action structures obtained by re-querying robot's own foundational walking policy under alternative internal rhythms. Refer to Sec~\ref{subsec:seed_policy_supp} in the Supp for the illustration of those action waves.

\noindent\textbf{{Hierarchical Multi-Resolution Coefficient Memory.}} The appropriate combination of canonical action waves changes continuously with the velocity command. A single global mapping from command to wave coefficients would be too coarse to capture transitions over a broad locomotion range, whereas hard command partitioning would introduce abrupt gait switches, as shown in Supp (Sec~\ref{app:wave_composition}). We instead introduce a hierarchical memory bank that learns command-conditioned wave coefficients at multiple resolutions. Specifically,  we construct a hierarchy of memory resolutions
$ \mathcal{H} = \{2,4,\cdots, l \}.$ At each resolution $L \in \mathcal{H}$, the command interval is represented by $L$ learnable memory units $
    \mathcal{M}^{(L)}
    =
    \left\{
        m^{(L)}_{\psi,1},
        m^{(L)}_{\psi,2},
        \ldots,
        m^{(L)}_{\psi,L}
    \right\}, $
where each unit predicts coefficients over the $K$ canonical action waves:
$
    \mathbf{\beta}^{(L)}_{t,j}
    =
    m^{(L)}_{\psi,j}
    \left(
        \mathbf{s}_t,
        \mathbf{v}_t
    \right)
    \in
    \mathbb{R}^{K}.$
Such a design ensures both global coherence and local expressiveness in gait growth, as demonstrated in Supp. Importantly, these units do not represent any named gaits; they jointly parameterize a continuous coefficient field over speed. For example, for commands in $[0,2.2]$\,m/s, the layer of two-cell bank models the broad split between $[0,1.1]$ and $(1.1,2.2]$\,m/s, while the layer of four- and eight-cell banks provide increasingly fine refinements within these intervals. For the boundary conditions, GaitWave softly blends the neighboring memory outputs:$
    \boldsymbol{\alpha}^{(L)}_t
    =
    \left(
        1-\lambda^{(L)}_t
    \right)
    \mathbf{\beta}^{(L)}_{t,j^-}
    +
    \lambda^{(L)}_t
    \mathbf{\beta}^{(L)}_{t,j^+}.$
This continuous retrieval avoids discontinuities at bin boundaries and allows gait changes to emerge smoothly with commanded speed. The coefficient predictions from different resolutions are aggregated as $
    \boldsymbol{\alpha}_t
    =
    \sum_{L \in \mathcal{H}}
    \omega^{(L)}_t
    \boldsymbol{\alpha}^{(L)}_t,$ with $
    \sum_{L \in \mathcal{H}}
    \omega^{(L)}_t = 1$. This ensures GaitWave preserves smooth transitions across commands while allowing walking-derived rhythms to grow into faster locomotion regimes without the risk of unnatural posture under conventional explicit gait switching, as demonstrated in Supp. With the above, we obtain 
  $  \mathbf{a}^{\mathrm{wave}}_t
    =
    \mathbf{a}^{\mathrm{seed}}_t
    +
    \sum_{k=1}^{K}
    \alpha_{t,k}
        \mathbf{a}^{\mathrm{seed}}_t
.$ At low commanded speeds, GaitWave can recover behavior close to the seed walking policy by assigning smaller weights to its expanded rhythm components. As the commanded speed increases, the hierarchical memory learns how alternative seed-derived rhythms should reshape the walking action.

\subsection{H-SLIP: Hierarchical Virtual-Leg Dynamic Shaping}
\label{sec:hslip}

GaitWave expands locomotion rhythms inherited from walking, but rhythmic expansion alone does not guarantee the emergence of physically meaningful dynamic gaits at high speeds (Fig.~\ref{fig:ablation}a). In particular, faster locomotion requires coordinated dynamic events, such as stance loading and compression, rebound, flight, and touchdown~\cite{DBLP:journals/siamrev/HolmesFKG06}. We thus introduce \textit{H-SLIP}, a hierarchical virtual-leg objective inspired by the spring-loaded inverted pendulum (SLIP)\cite{jindrich2002dynamic,DBLP:conf/iros/WensingO13} abstraction.

\noindent\textbf{Hierarchical Virtual Legs.} We define three levels of virtual-leg abstractions for each foot $i$. The global virtual leg connects the robot root to the foot, while the local virtual legs connect the root to knee, and knee to foot, as illustrated in Fig~\ref{fig:framework}c:
$
    \ell^{\mathrm{root}}_{t,i}
    =
    \left\|
        \mathbf{p}^{\mathrm{root}}_t
        -
        \mathbf{p}^{\mathrm{foot}}_{t,i}
    \right\|_2,
    \ell^{\mathrm{upper}}_{t,i}
    =
    \left\|
        \mathbf{p}^{\mathrm{root}}_{t,i}
        -
        \mathbf{p}^{\mathrm{knee}}_{t,i}
    \right\|_2, 
        \ell^{\mathrm{lower}}_{t,i}
    =
    \left\|
        \mathbf{p}^{\mathrm{knee}}_{t,i}
        -
        \mathbf{p}^{\mathrm{foot}}_{t,i}
    \right\|_2,$
Their temporal variations are approximated by
$
    \dot{\ell}^{x}_{t,i}
    =
    \frac{
        \ell^{x}_{t,i}
        -
        \ell^{x}_{t-1,i}
    }{\Delta t},
    x \in
    \left\{
        \mathrm{root},
        \mathrm{upper},
        \mathrm{lower}
    \right\}.$
Let $c_{t,i} \in \{0,1\}$ indicate whether foot $i$ is in ground contact. The root--foot virtual leg captures whole-body support and center-of-mass dynamics, while the knee--foot virtual leg captures local lower-limb compression and rebound near contact.

\noindent\textbf{Hierarchical Dynamics Shaping.} Given the virtual-leg representation, H-SLIP organizes dynamic locomotion through four key stages: compression, rebound, touchdown, and flight. Intuitively, compression and rebound regulate how energy is absorbed and released through the virtual-leg hierarchy, while touchdown and flight shape contact transitions that become increasingly important as locomotion enters jogging- and running-like regimes. For each virtual-leg level $x$, we define the compression and rebound terms using the temporal change in virtual-leg length:
$
    r^{x}_{\mathrm{comp},t}
    =
    \sum_i
    c_{t,i}
    \operatorname{ReLU}
    \left(
        -\dot{\ell}^{x}_{t,i}
    \right),
    r^{x}_{\mathrm{reb},t}
    =
    \sum_i
    c_{t,i}
    \operatorname{ReLU}
    \left(
        \dot{\ell}^{x}_{t,i}
    \right).$
The global root--foot level shapes whole-body loading and propulsion, while the upper- and lower-leg levels encourage coordinated articulation of local limb segments during support and release. Since rebound should contribute to commanded locomotion rather than induce task-irrelevant bouncing, we modulate it by a velocity-tracking gate:
$
    g^{\mathrm{track}}_t
    =
    \exp
    \left(
        -
        \frac{
            \left\|
                \mathbf{v}^{\mathrm{base}}_{t,xy}
                -
                \mathbf{v}_{t,xy}
            \right\|_2^2
        }{
            \sigma_{\mathrm{track}}^2
        }
    \right).
$ We further shape contact transitions through touchdown and flight. Let
$d_{t,i}=\mathbf{1}[c_{t,i}=1 \land c_{t-1,i}=0]$
indicate that foot $i$ establishes a new contact. We reward controlled touchdown by discouraging large planar foot velocity at contact:
$
    r_{\mathrm{td},t}
    =
    \sum_i
    d_{t,i}
    \exp
    \left(
        -
        \frac{
            \left\|
                \mathbf{v}^{\mathrm{foot}}_{t,i,xy}
            \right\|_2^2
        }{
            \sigma_{\mathrm{td}}^2
        }
    \right),
    r_{\mathrm{flight},t}
    =
    \mathbf{1}
    \left[
        \sum_i c_{t,i}=0
    \right].$ Touchdown encourages stable support formation after swing, while flight encourages the policy to leave quasi-static walking when sufficiently dynamic motion is required. In practice, we additionally penalize undesirable stance slips that are command-orthogonal, as it destabilizes contact and can lead to hardware damage during real-world deployment. H-SLIP combines these event terms across the virtual-leg hierarchy:
\begin{equation}
\begin{aligned}
    r^{\mathrm{H\text{-}SLIP}}_t
    =
    g^{\mathrm{dyn}}_t
    \Bigg[
        &
        \sum_{x}
        \left(
            w^{x}_{\mathrm{comp}}
            r^{x}_{\mathrm{comp},t}
            +
            g^{\mathrm{track}}_t
            w^{x}_{\mathrm{reb}}
            r^{x}_{\mathrm{reb},t}
        \right)+
        w_{\mathrm{td}} r_{\mathrm{td},t}
        +
        w_{\mathrm{flight}} r_{\mathrm{flight},t}
    \Bigg]
    +
    w_{\mathrm{slip}} r_{\mathrm{slip},t}.
    \label{eq:hslip_reward}
\end{aligned}
\end{equation}
H-SLIP provides a gait-agnostic physical bias that preserves the event-level physical structure of SLIP without inheriting the hard trajectory-level constraints of a simplified spring-mass template. Thus, it guides expanded motions toward jogging- and running-like regimes while retaining flexibility across speeds and embodiments.

The total reward to train GaitSpan combines the standard locomotion objective~\cite{Amazon_FAR_and_Abbeel_Holosoma}, the proposed H-SLIP dynamic-shaping reward, and auxiliary self-anchored regularization:
\begin{equation}
    r_t^{\mathrm{total}}
    =
    r_t^{\mathrm{task}}
    +
    \lambda_{\mathrm{H\text{-}SLIP}}
    r_t^{\mathrm{H\text{-}SLIP}}
    -
    \lambda_{\mathrm{SA}}
    \mathcal{L}_{\mathrm{SA},t}.
    \label{eq:total_objective}
\end{equation}
Here, $\mathcal{L}_{\mathrm{SA},t}$ includes low-speed seed anchoring, residual compactness, phase-aware self-similarity, and temporal consistency terms (Fig~\ref{fig:framework}b), which preserve the inherited walking structure without preventing command-driven dynamic gait growth. Refer to the Supp for details.

\section{Experiments}
\label{sec:exp}

\noindent\textbf{Robots.}
We evaluate our method on five humanoid robot configurations: Unitree G1 (29\,DoF, 23 \,DoF), Booster T1(29\,DoF, 23\,DoF), and Booster K1 (22\,DoF). We use official URDF models provided by the manufacturers. For each robot, we use the same training pipeline. 

\noindent\textbf{Implementations.}
Observations cover proprioception (angular velocity, projected gravity, joint positions and velocities, the previous action, and a sinusoidal phase encoding) and commanded velocities. Policy outputs target joint positions, which are converted to torques through a PD controller. Unless specified, policies are trained with \textit{FastSAC}. Policies are trained with $4096$ parallel environments in IsaacGym~\cite{DBLP:conf/nips/MakoviychukWGLS21}, and sim-to-sim evaluation is trained on IsaacGym, evaluated on MujoCo\cite{todorov2012mujoco}. {\textit{Refer to Supp for more details.}}
\begin{figure}[t]
    \centering
    \includegraphics[width=\textwidth]{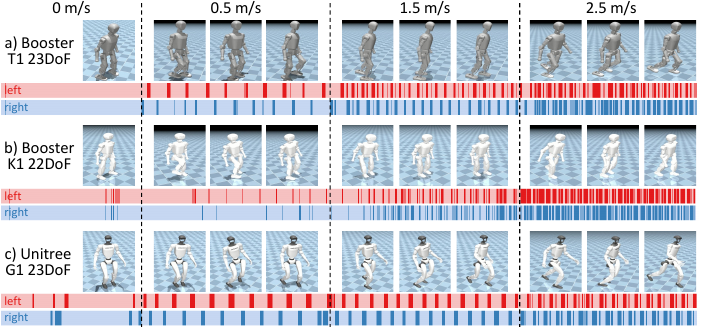}
    \vspace{-0.4cm}
\caption{\textbf{Emergent gaits across speeds and embodiments.} Foot-contact plots for the left (red) and right (blue) feet reveal smooth transitions from standing and walking contacts to increasingly dynamic, flight-producing patterns as commanded speed increases. At the same commanded speed, different humanoids exhibit distinct emergent contact patterns (bold color means the corresponding foot is in contact with the ground and the light color means the foot is in the air), demonstrating that GaitSpan adapts gait growth to each robot's body structure.}
\label{fig:footcontact}

\end{figure}

\noindent\textbf{Metrics.}
We evaluate command tracking and dynamic gait emergence in both in-distribution and out-of-distribution settings. \textcolor{green!50!black}{\textbf{Tracking Error}} measures the discrepancy between commanded and executed base velocity; \textcolor{blue!65!black}{\textbf{Flight Time}} measures the duration of aerial phases with no foot contact; and \textcolor{mypurple}{\textbf{Energy}} measures motion cost by integrating joint torque--velocity products over time.

\subsection{Beyond Walking - Emergence of Jogging, Striding, and Running}
\label{sec:emergent_gaits}

We first examine the behaviors that emerge in humanoid robots to understand how a seed walking skill grows in response to commanded speed and embodiment. 

\noindent\textbf{Across speeds.}
As the commanded speed increases, GaitSpan exhibits a continuous progression from stable walking to longer-stride, jogging-, and running-like behaviors. In the foot-contact plots of Fig.~\ref{fig:footcontact}, low-speed commands produce regular alternating support with little or no flight, whereas higher-speed commands exhibit reorganized contact frequencies and denser contact--flight transitions, revealing the emergence of increasingly dynamic gait patterns. The time-lapse sequences in Fig.~\ref{fig:teaser} further show that these transitions remain stable on real-world terrains. In particular, Fig.~\ref{fig:teaser}(c,d) visualizes smooth motion transitions from walking toward faster dynamic regimes, supporting our central claim that new gaits emerge by expanding the seed walking skill, rather than through explicit gait switching or separately trained behaviors, which often yield more discrete and less flexible transitions.

\noindent\textbf{Across embodiments.}
Gait emergence is not identical across robot bodies. At the same commanded speed, Booster T1, Booster K1, and Unitree G1 exhibit distinct foot-contact patterns in Fig.~\ref{fig:footcontact}, reflecting embodiment-specific adaptation to differences in morphology and actuation. Importantly, GaitSpan neither imposes rigid embodiment-specific design nor a singular gait schedule over all robots; it provides a shared skill-growth principle that allows each embodiment to develop its own command-appropriate locomotion patterns in a flexible manner.

\begin{figure*}[t]
\small\centering
\newcommand{\squareimage}[1]{\includegraphics[width=0.165\linewidth, height=0.165\linewidth]{#1}}
\newcommand{\colspace}{\hspace{1pt}}
\newcommand{\colspaceclose}{\hspace{0pt}}
\newcommand{\colspacenone}{\hspace{0pt}}
\newcommand{\tablerow}[1]{
\squareimage{figs/imgs/velocity_tracking_acc/ablation_robots_#1_abs.png} &
\squareimage{figs/imgs/velocity_tracking_acc/ablation_robots_#1_flight.png} 
}
\resizebox{\textwidth}{!}{
\renewcommand{\arraystretch}{0.3}
\begin{tabular}
{@{}c@{\colspaceclose}c@{\colspace}c@{\colspaceclose}c@{\colspace}c@{\colspaceclose}c@{}}
\tablerow{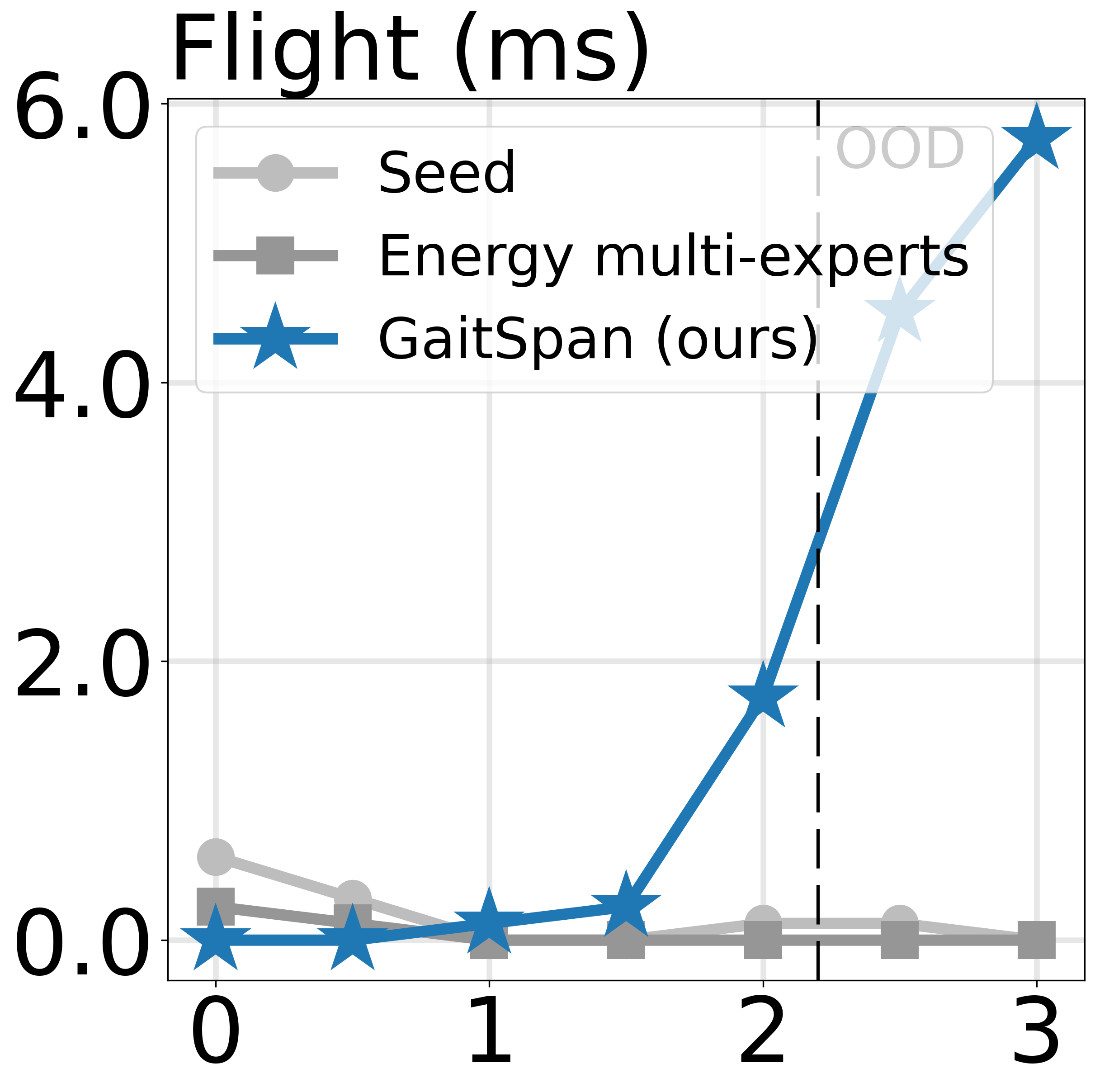} & \tablerow{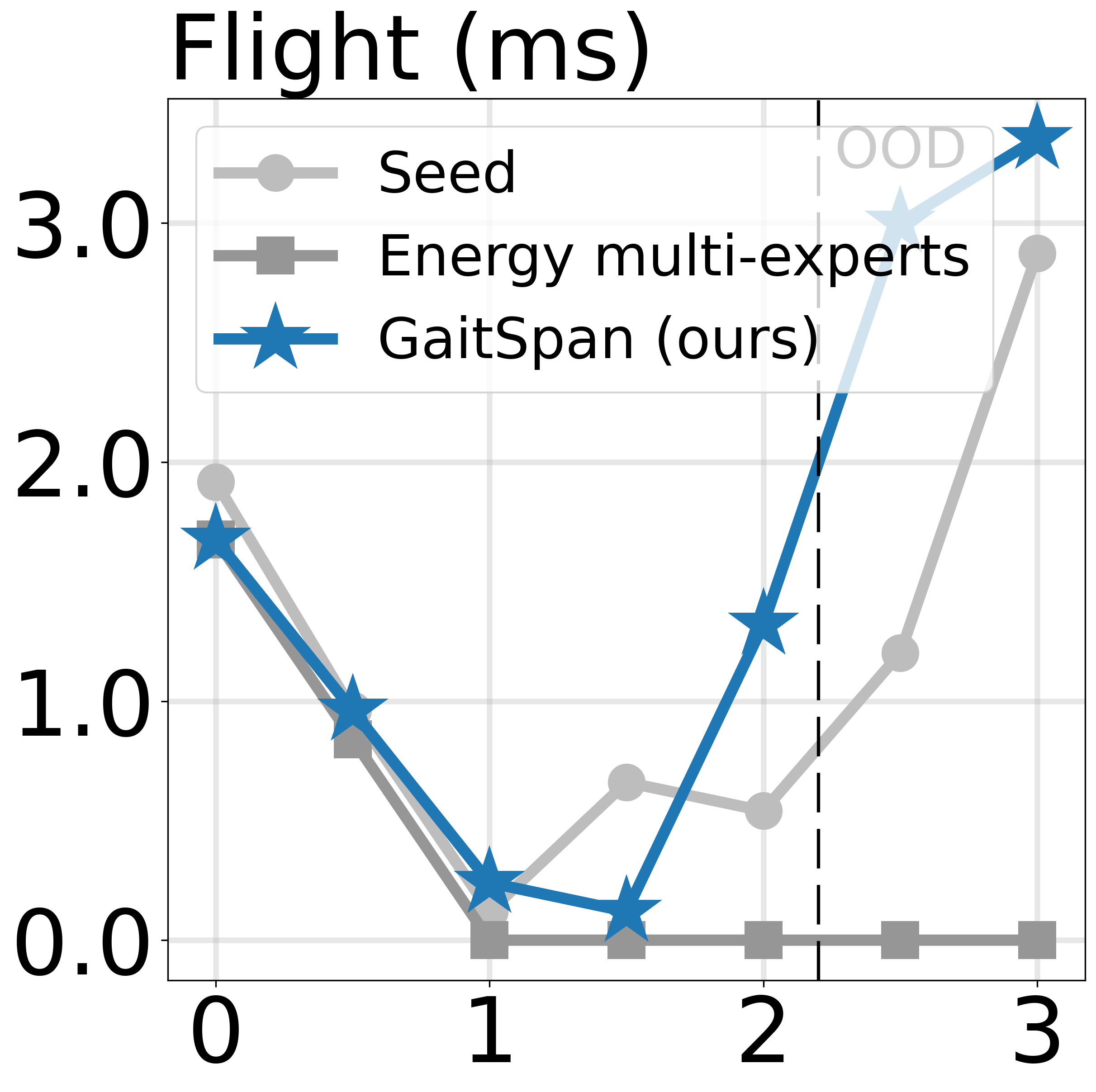} & \tablerow{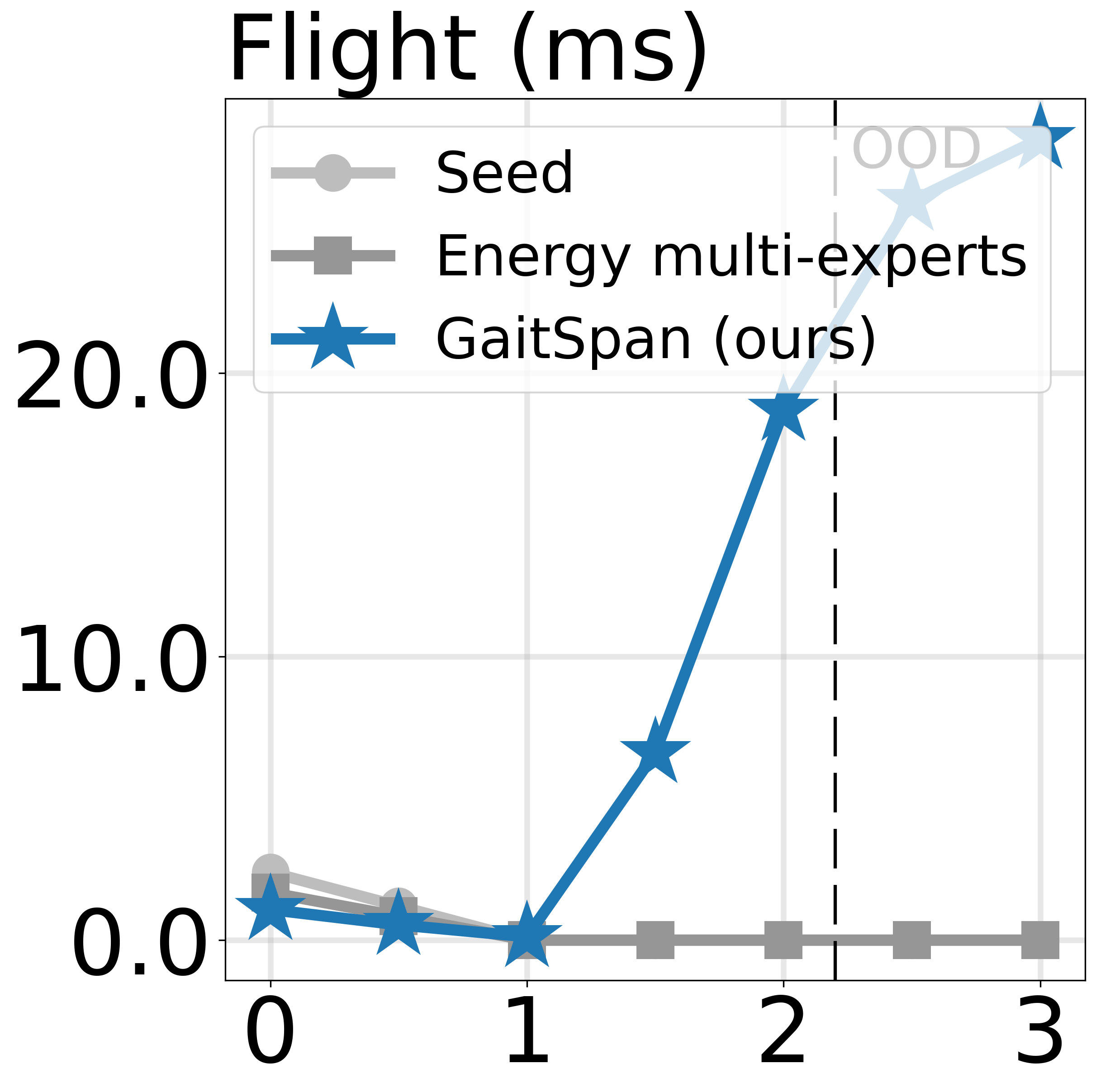} \\
\multicolumn{2}{c}{a) Booster T1 23DoF} & \multicolumn{2}{c}{b) Booster K1 22DoF} & \multicolumn{2}{c}{c) Unitree G1 23DoF} \\
\end{tabular}
}
\vspace{-0.1cm}
\caption{\textbf{Tracking accuracy and dynamic gait emergence across speeds and embodiments.} Across Booster T1, K1, and Unitree G1, GaitSpan maintains lower tracking error while developing substantial flight time at high speed under both in-distribution and out-of-distribution (OOD) commands. In contrast, the walking seed and energy-based multi-expert baseline exhibit degraded tracking and fail to produce command-appropriate flight.}
\label{fig:vel_tracking_acc}
\end{figure*}

\begin{figure}[t]
\newcommand{\close}{\hspace{1pt}}
\centering\small
\begin{minipage}[t]{0.32\linewidth}
\newcommand{\squareimage}[1]{\includegraphics[width=0.49\textwidth, height=0.49\textwidth]{#1}}
\vspace{0pt}
\centering\small
\resizebox{\textwidth}{!}{
\begin{tabular}{@{}c@{\close}c@{}}
\squareimage{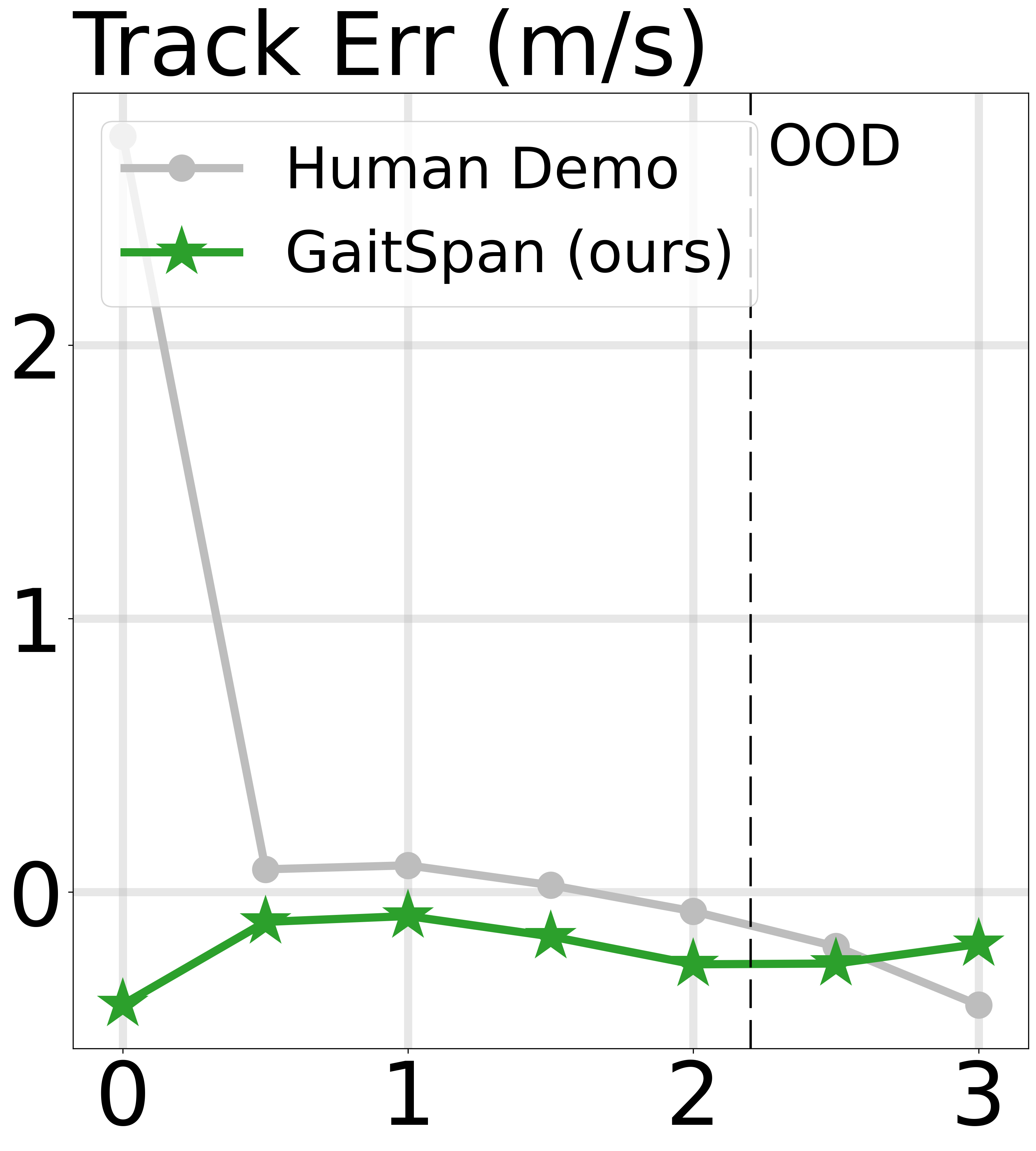} &
\squareimage{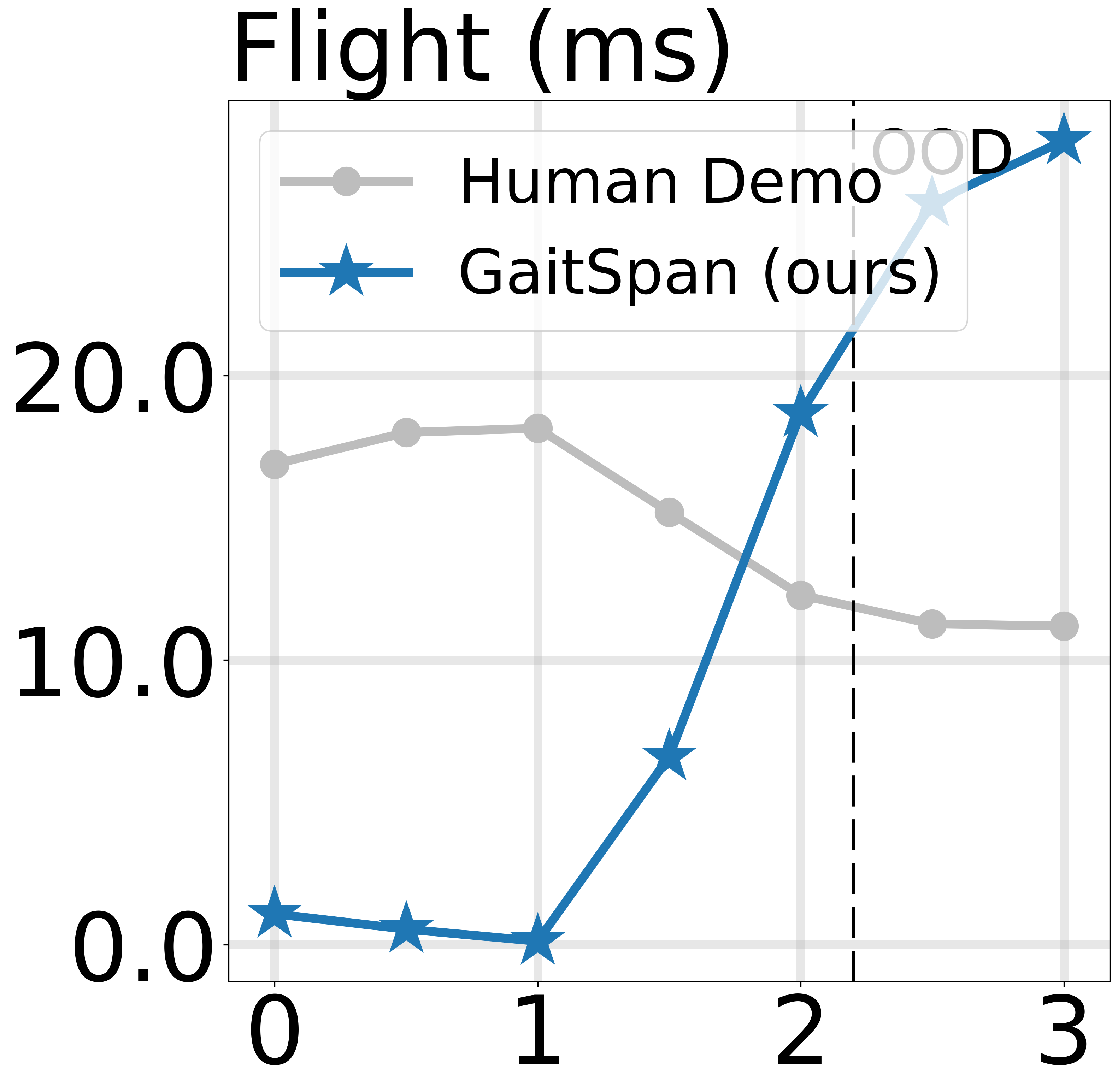}
\end{tabular}
}

\captionof{figure}{\textbf{Comparison with a human-demonstration baseline on Unitree G1 (23 DoF).} Tracking error and flight time are evaluated.}

\label{fig:humandemo}
\end{minipage}
\hfill
\begin{minipage}[t]{0.65\linewidth}
\newcommand{\squareimage}[1]{\includegraphics[width=0.24\textwidth, height=0.24\textwidth]{#1}}
\newcommand{\tablerow}[1]{
\squareimage{figs/imgs/ablation/#1_abs.png} &
\squareimage{figs/imgs/ablation/#1_flight.png} 
}
\vspace{0pt}
\centering\small
\resizebox{\textwidth}{!}{
\begin{tabular}{@{}c@{\close}c@{\close}c@{\close}c@{}}
\tablerow{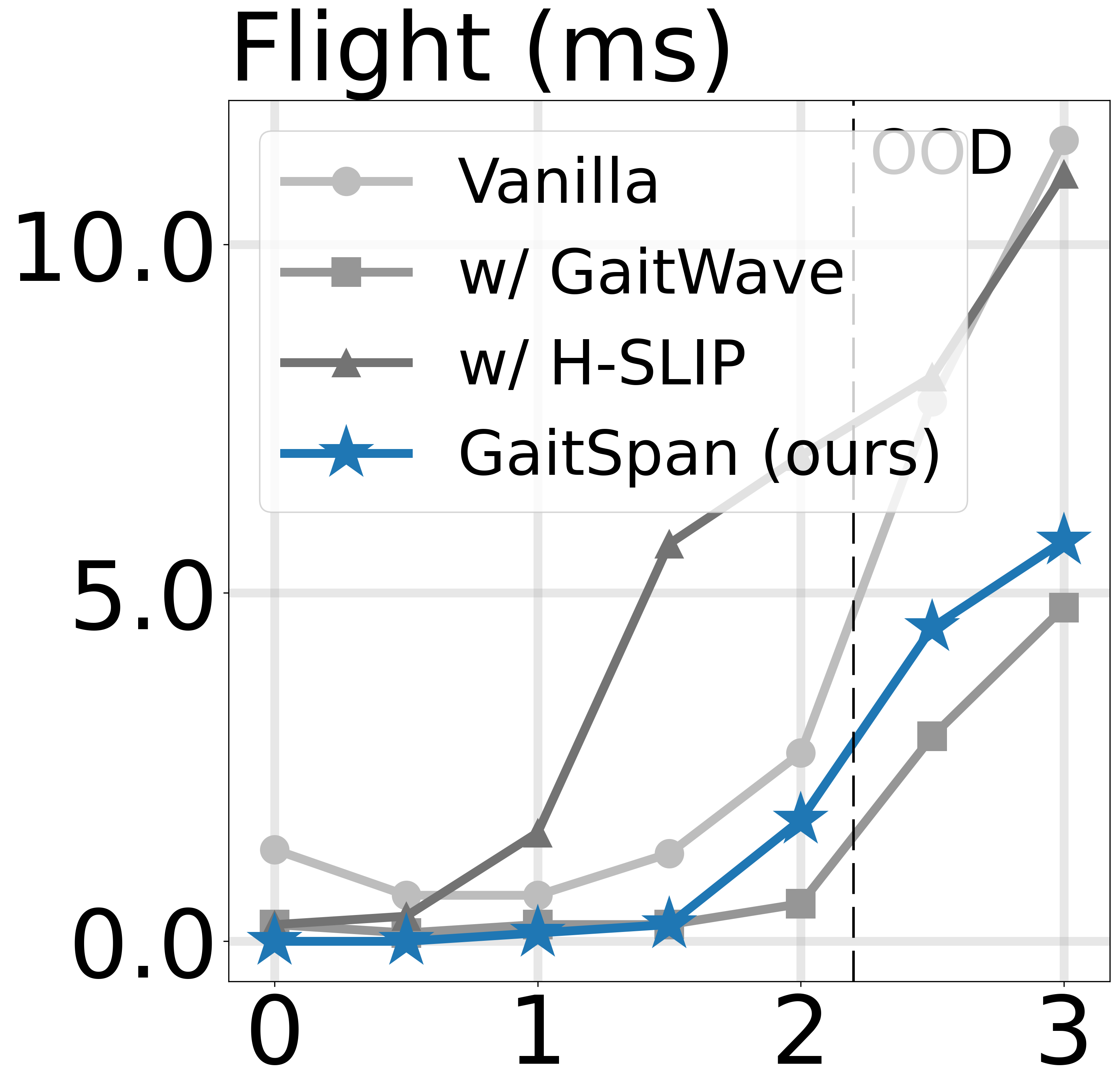} & \tablerow{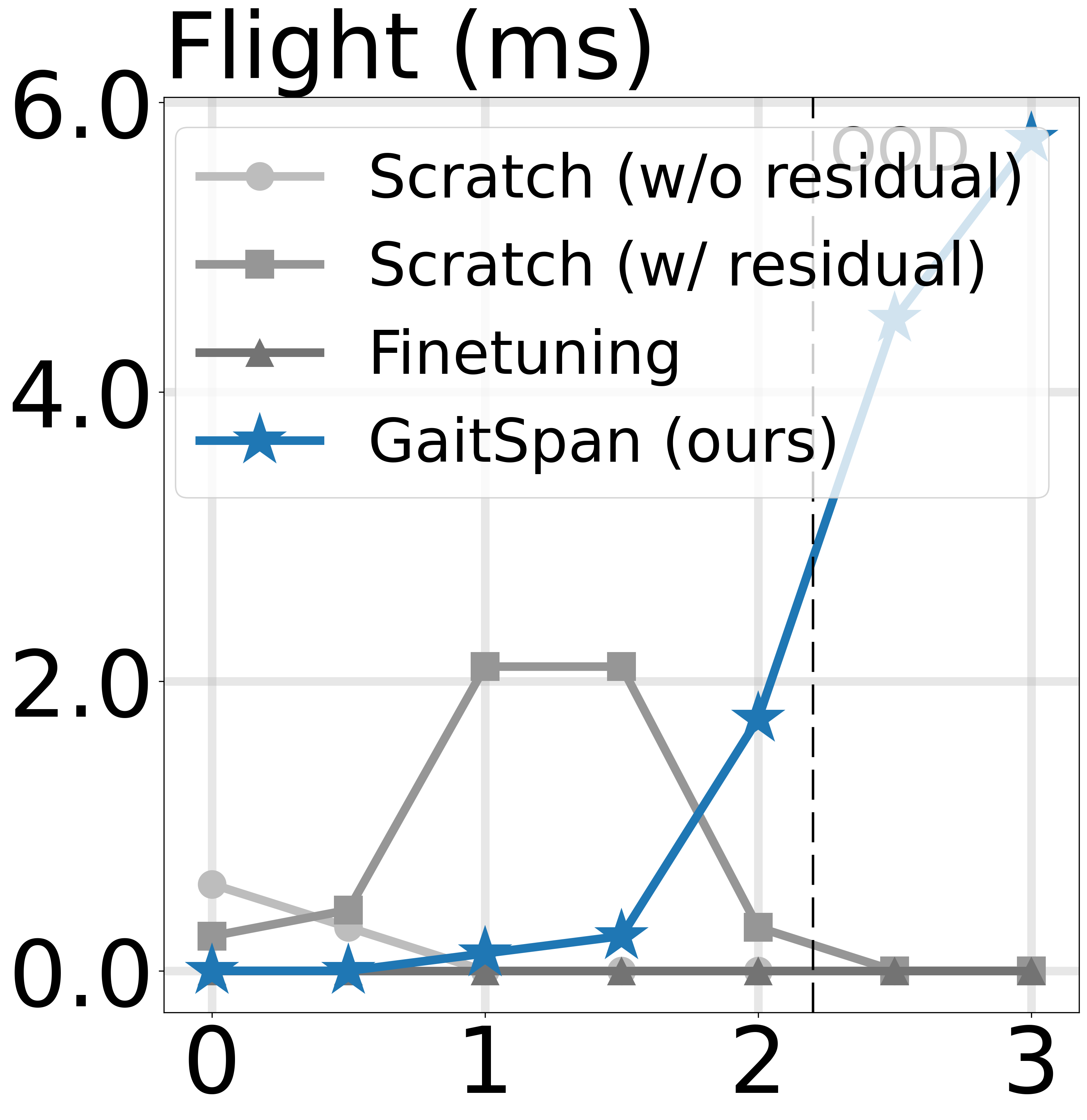} \\
\multicolumn{2}{c}{a) main components} & \multicolumn{2}{c}{b) training strategies}
\end{tabular}
}
\vspace{-0.1cm}
\captionof{figure}{\textbf{Ablation study of key components and training strategies on Booster T1.} We evaluate the contributions of main components, and compare GaitSpan with from-scratch training and direct fine-tuning.}

\label{fig:ablation}

\end{minipage}
\vspace{-0.1cm}
\end{figure}

\subsection{Baseline Comparison}
\label{sec:baseline}

Having shown that GaitSpan produces diverse gait regimes, we next ask whether this skill growth preserves control quality and efficiency. We examine three questions: \textbf{1)} Does {\it jogging}- and {\it running}-like emergence sacrifice command tracking? \textbf{2)} Is a continuous gait repertoire better grown from one seed skill than assembled from energy-shaped experts? \textbf{3)} Can skill growth produce dynamic, energy-efficient locomotion without human-motion demonstrations?

\noindent\textbf{Settings.} Since current research does not directly support the task, we compare \textit{GaitSpan} with the following designs that adapting them to this task: \textit{1) Seed}, the frozen walking policy directly evaluated over the extended command range, testing whether walking alone can extrapolate to faster regimes; \textit{2) Energy Multi-Experts}~\cite{fu2021minimizing}, a set of separately trained energy-shaped gait policies combined for different speed regimes, testing skill assembly against skill growth; and \textit{3) Human Demonstration}, we use AMP~\cite{2021-TOG-AMP}, a locomotion policy trained with human-motion priors, testing whether external demonstrations are necessary to produce dynamic and energetically efficient gaits.

\noindent\textbf{Results.} Fig.~\ref{fig:vel_tracking_acc} and Fig.~\ref{fig:humandemo} in the main paper, and Fig.~\ref{fig:humandemo_w_energy_compare},~\ref{fig:energy_based_multi_expert_compare},~\ref{fig:baseline_sidebyside},~\ref{fig:ours_vs_humandemo_g1_29_qualitative} in the Supp show key three findings. \textbf{$1)$} {\textit{Gait emergence does not compromise command following}}. Across all three embodiments and both in-distribution and OOD commands, GaitSpan keeps low tracking error while flight time increases with speed, indicating that {\it jogging}- and {\it running}-like behaviors emerge without losing tracking accuracy (Fig.~\ref{fig:vel_tracking_acc}). \textbf{$2)$} {\textit{Growing gaits from one {\it walking} skill is more effective than assembling energy-shaped experts}}. The multi-expert baseline, despite combining separately trained behaviors, shows degraded high-speed tracking and almost no flight phases due to its energy-oriented objectives favoring conservative motion (Fig.~\ref{fig:energy_based_multi_expert_compare},~\ref{fig:baseline_sidebyside}), suggesting that expert composition does not yield a coherent continuous gait family.~\textbf{$3)$} {\textit{Seed-skill growth can be more command-responsive than human-motion priors.}}
Compared with the human-demonstration baseline, GaitSpan tracks commands more accurately across the evaluated range and exhibits a clearer speed-dependent reorganization of contact dynamics, where it preserves little or no flight at walking speeds, but develops increasingly pronounced simultaneous no-contact intervals under high-speed and OOD commands (Fig.~\ref{fig:humandemo}). GaitSpan also consumes less energy over most of the command range while remaining competitive at the highest speeds (Fig~\ref{fig:humandemo_w_energy_compare}). These results also suggest that human-like imitation is not always necessary for learning dynamic, energy-conscious, and command-adaptive humanoid gaits, as the most human-like motion (Fig.~\ref{fig:ours_vs_humandemo_g1_29_qualitative}) is not necessarily the most effective motion for a humanoid robot.

\subsection{Ablation Study}
\label{sec:ablation}

\noindent\textbf{Settings.}
In the main paper, we conduct two groups of ablations. \textbf{1)} Component contributions. We compare the vanilla seed--residual baseline, variants using only \textit{GaitWave} or only \textit{H-SLIP}, and the full \textit{GaitSpan} model. \textbf{2)} Skill-expansion strategies. We compare training from scratch with and without a residual branch, direct fine-tuning of the seed policy, and our self-anchored expansion formulation. More fine-grained analyses and additional experiments are provided in the Supp (Sec~\ref{appendix:ablation}).

\noindent\textbf{Results.}
1)~{\textit{GaitWave and H-SLIP play complementary roles.} In Fig.~\ref{fig:ablation}a, neither the vanilla seed--residual baseline nor either single-component variant simultaneously achieves accurate command tracking and meaningful dynamic gait emergence across the whole range. GaitWave alone remains close to the seed walking solution and therefore attains slightly lower tracking error at several moderate-speed commands. However, it produces only limited flight and becomes less expressive at the most challenging OOD command, indicating that rhythm expansion alone is insufficient to induce robust and diverse high-speed locomotion behaviors. Conversely, H-SLIP alone generates substantial flight at higher speeds, but its tracking deteriorates sharply under OOD commands, suggesting overly aggressive dynamic motion without sufficient command-conditioned structure. The full \textit{GaitSpan} model achieves the best overall balance, as it maintains accurate tracking across the command range, develops flight progressively as speed increases, and remains substantially more robust at the extreme OOD command. This reveals a synergistic coupling between the two components: \textit{GaitWave} provides structured, command-conditioned expansion of the seed rhythm, while \textit{H-SLIP} supplies the physical bias needed to organize that expansion into controlled dynamic strides.

2)~{\textit{How the walking skill is reused matters.}
In Fig.~\ref{fig:ablation}b, training from scratch, with or without a residual branch, fails to recover a coherent gait family across the full command range. These variants either lose tracking accuracy at higher speeds or produce transient flight at inappropriate commands that does not persist as speed increases. Direct fine-tuning shows rapidly increasing tracking error with commanded speed while producing little sustained flight, indicating that adapting the seed policy directly can overwrite its stable walking structure without reliably developing running-like behavior. In contrast, \textit{GaitSpan} preserves accurate low-speed locomotion while introducing increasingly dynamic contact patterns only as commanded speed grows. This demonstrates that structured expansion of an acquired walking skill is more effective than either relearning locomotion from scratch or directly overwriting the pretrained policy.

\section{Conclusion}\label{sec:conclusion}

We presented \textbf{GaitSpan}, a framework that grows a humanoid's foundational walking skill into a broader locomotion repertoire. Instead of prescribing gait labels, imitating human motions, or assembling separately trained experts, GaitSpan expands an acquired walking policy through rhythm generation with \textit{GaitWave}, stride shaping with \textit{H-SLIP}, and residual adaptation. Our findings show that a learned motor skill can be more than an endpoint of training, when treated as a seed for structured expansion, it can generate richer families of related humanoid behaviors.

\section{Limitations}
GaitSpan depends on the quality and coverage of its {\it walking} seed; a weak or narrow seed can constrain the behaviors that emerge. It also demonstrates one-step skill growth, from {\it walking} to {\it jogging} and {\it running}, rather than continual self-expansion that folds newly acquired behaviors back into the seed. A staged, closed-loop recursion could address this limitation; choosing when and how to reseed requires a thorough investigation.

\clearpage

\bibliography{example}

\appendix

\onecolumn

\title{\papertitle}

\begin{center}{\bf \LARGE \textls[0]{\papertitle}}\end{center}
\begin{center}{\Large Supplementary Materials}\end{center}

\hypersetup{linkcolor=black}
\etocdepthtag.toc{mtappendix}
\etocsettagdepth{mtchapter}{none}
\etocsettagdepth{mtappendix}{subsection}

\definecolor{hotpink}{RGB}{255, 105, 180} 

\hypersetup{
    colorlinks=true,
    allcolors=hotpink,
}
\tableofcontents
\clearpage
\hypersetup{linkcolor=red}
\begin{abstract}
In this supplementary material, we provide a more detailed elaboration of \textit{GaitSpan}. We first recap the positioning of {\textit{GaitSpan}} and highlight how it differs from current humanoid locomotion methods (Sec.~\ref{appendix:position}). We then present additional baseline comparisons to further analyze the role of the seed walking policy, the limitations of direct phase scaling, the energy behavior of human-demonstration baselines, and qualitative differences across methods (Sec.~\ref{appendix:baseline}). We also provide additional ablation studies that examine the detailed designs of GaitWave and H-SLIP, including wave composition, speed-dependent coefficient usage, hierarchical SLIP shaping, and component generalization across robot embodiments, as well as additional emergent behavior analyses (Sec. ~\ref{appendix:ablation}). Finally, we include additional implementation details (Sec. ~\ref{appendix:implementation}).

\end{abstract}
\section{The Positioning of GaitSpan}
\label{appendix:position}

\begin{figure}[h]
    \centering
    \includegraphics[width=\textwidth]{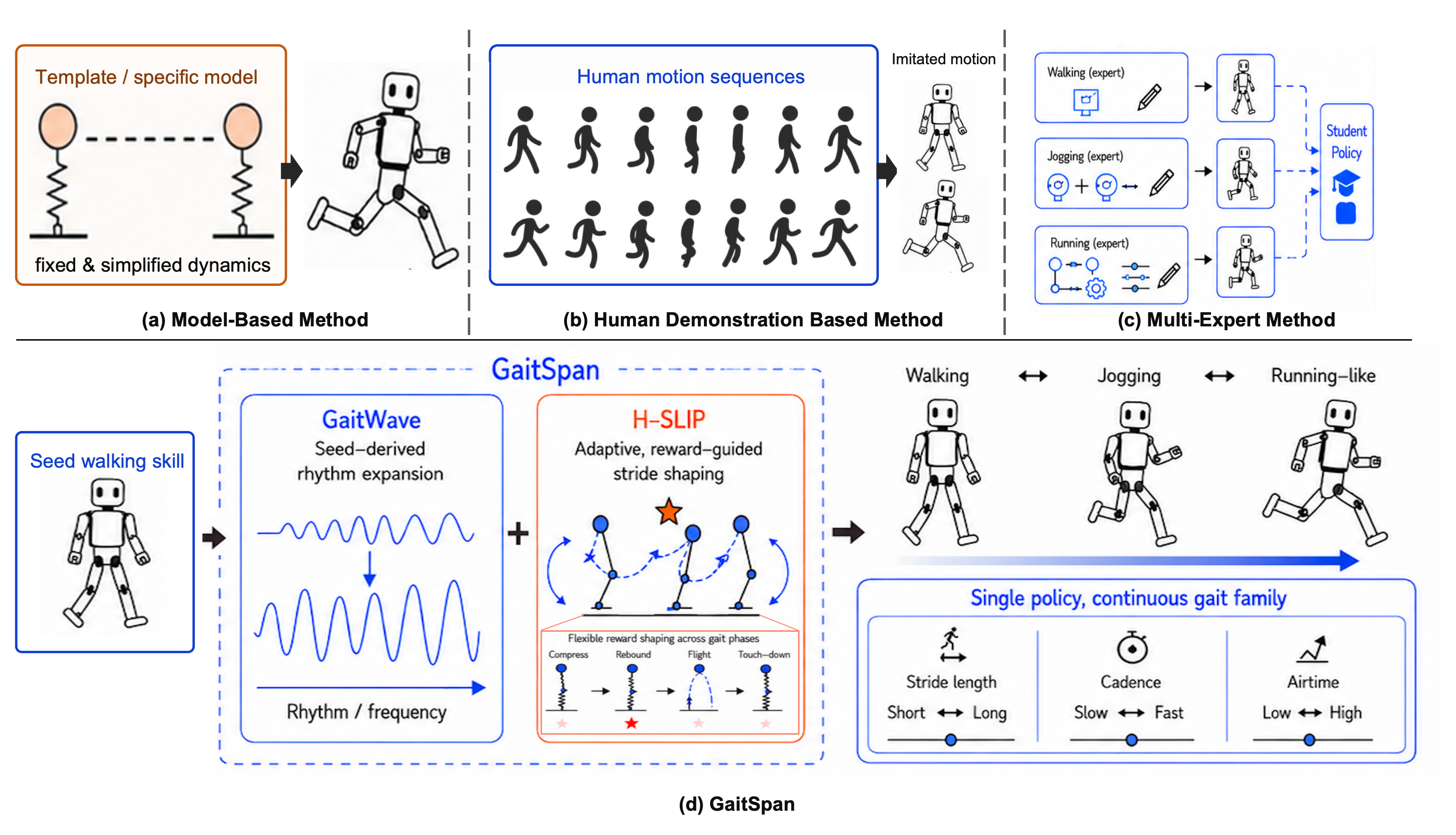}
    \vspace{-0.4cm}
\caption{\textbf{The conceptual framework differences.} We summarize the key conceptual level differences between our work and current humanoid robot trends on locomotion for better positioning of GaitSpan.}
\label{fig:positioning}
\end{figure}

\label{sec:positioning}

To better understand \sname{}'s positioning, we present a conceptual framework comparison in Fig.~\ref{fig:positioning}. Existing approaches to diverse humanoid locomotion typically fall into three paradigms. Model-based methods (Fig.~\ref{fig:positioning} a) construct locomotion through simplified dynamics or specific templates, providing useful physical structure but often limiting flexibility. Human-demonstration-based methods (Fig.~\ref{fig:positioning} b) import rich motion priors without explicitly modeling robot dynamics, but rely on external cross-embodiment data that may mismatch the robot's morphology, actuation, contact dynamics, and energy objectives, while remaining constrained by the demonstrated motion distribution. Multi-expert methods (Fig.~\ref{fig:positioning} c) train separate controllers for skills like walking, jogging, and running, then distill them into one policy. This treats gait diversity as the assembly of independently acquired skills rather than the growth of one skill into another. Such designs can be costly to scale, sensitive to expert boundaries, and prone to discontinuous or conflicting behavior during composition.

In contrast, \sname{} (Fig.~\ref{fig:positioning}d) views walking as a seed skill from which a broader family of related locomotion regimes can grow. Starting from a frozen walking policy, GaitWave generates seed-derived rhythm variations to modulate cadence and timing, while H-SLIP provides reward-guided stride shaping through encouraging representative dynamic events such as compression, rebound, flight, and touchdown. A residual branch captures remaining command- and state-dependent adaptations. The resulting policy spans walking-, jogging-, and running-like regimes with continuous gait changes. This makes \sname{} a skill-growth framework, where walking is not a terminal behavior or one expert among many, but the seed from which a continuous family of humanoid gaits emerges. By expanding rather than overwriting the seed skill, \sname{} preserves the reusable structure of walking while allowing new regime-specific dynamics to emerge upon it.

\section{Additional Baseline Comparisons and Analysis}
\label{appendix:baseline}
In this section, we provide additional results and analysis for the baselines discussed in the main paper. We focus on three questions: {\textbf{1)}} what locomotion structure is already encoded in the seed walking policy, what benefits and limitations arise from directly modifying its phase (Subsec ~\ref{subsec:seed_policy_supp})? {\textbf{2)}} How does GaitSpan compare with human-demonstration-based locomotion in energy and dynamic gait emergence (Subsec ~\ref{app:human_demo_energy})? and  {\textbf{3)}} how different baselines behave qualitatively across speeds and embodiments settings (Subsec ~\ref{app:qualitative_compare})? These analyses further clarify why GaitSpan grows gaits from walking rather than simply extrapolating, imitating, or assembling existing behaviors.

\subsection{What the Seed Walking Policy Encodes}
\label{subsec:seed_policy_supp}

\begin{figure}[h]
\centering\small
\newcommand{\squareimage}[1]{\includegraphics[width=0.49\textwidth, height=0.32\textwidth]{#1}}

\resizebox{\textwidth}{!}{
\begin{tabular}{@{}c@{\hspace{2pt}}c@{}}
\squareimage{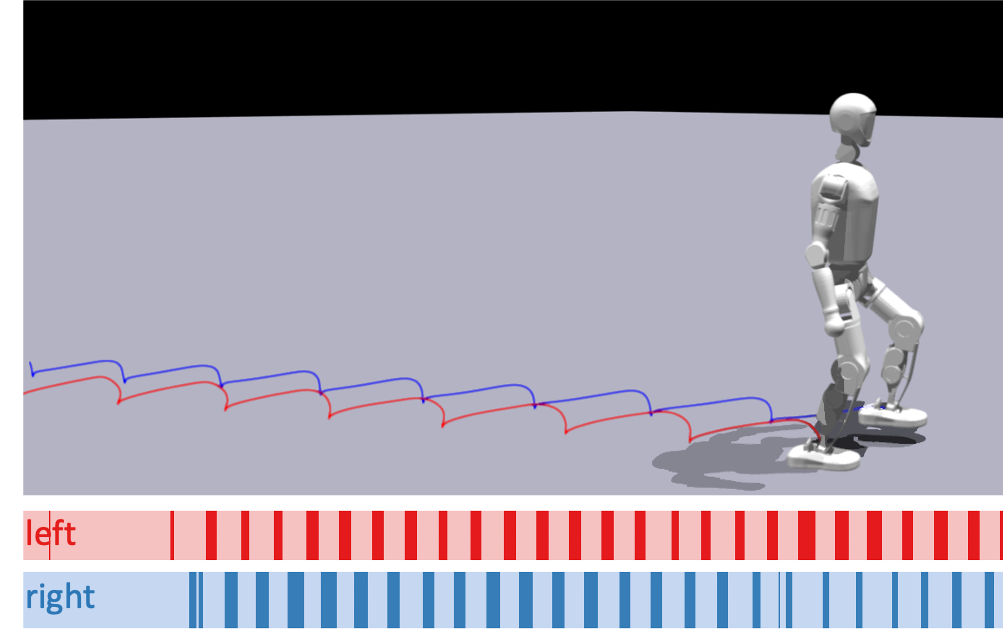} &
\squareimage{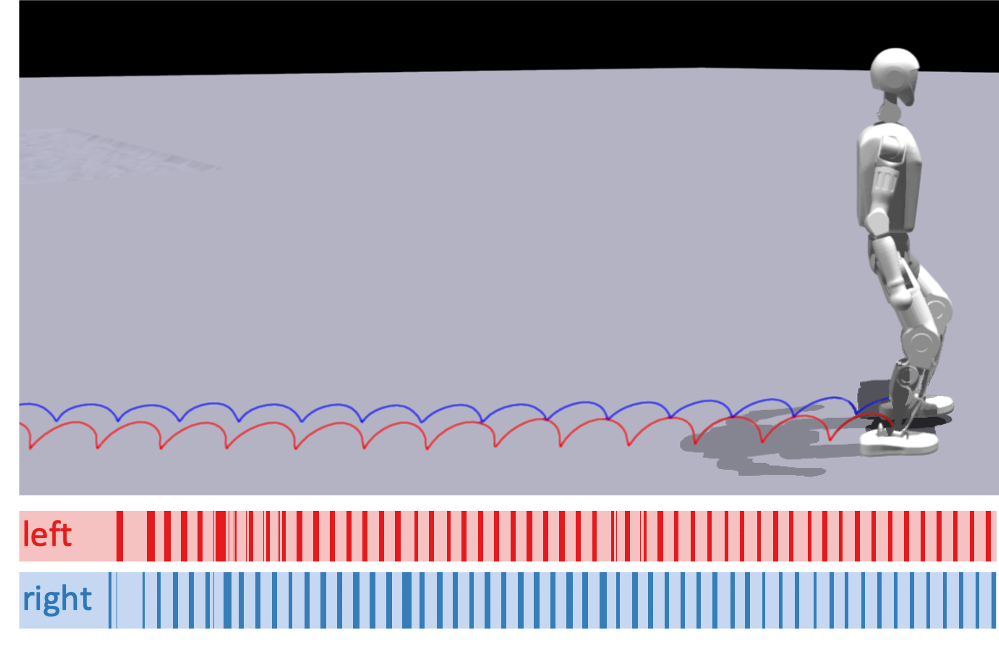} \\
a) seed policy at 1.0 frequency, 0.5m/s & b) seed policy at 2.0 frequency, 0.5m/s \\
\squareimage{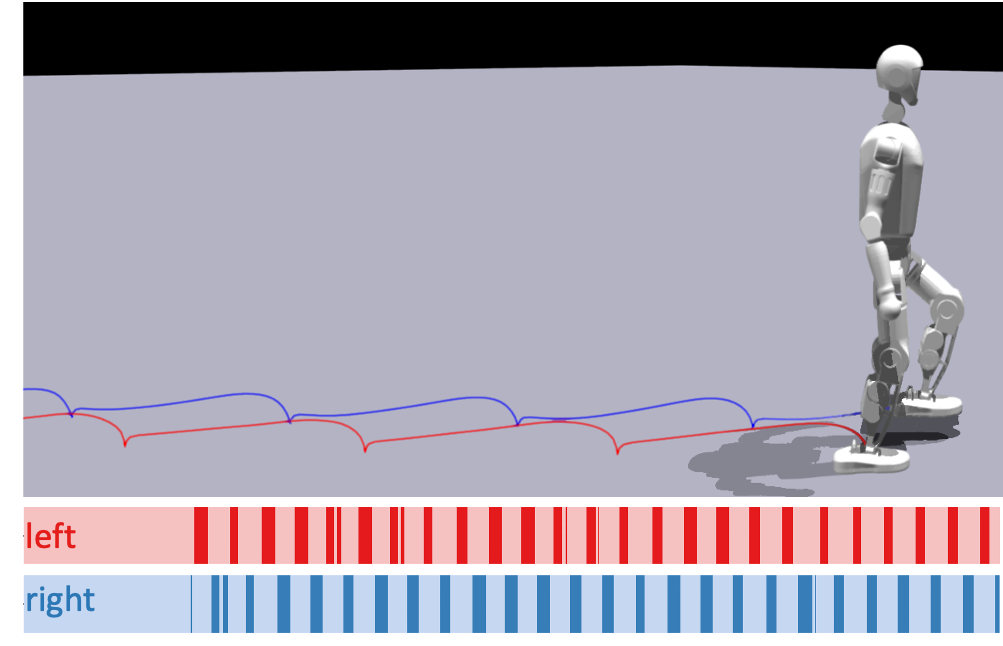} &
\squareimage{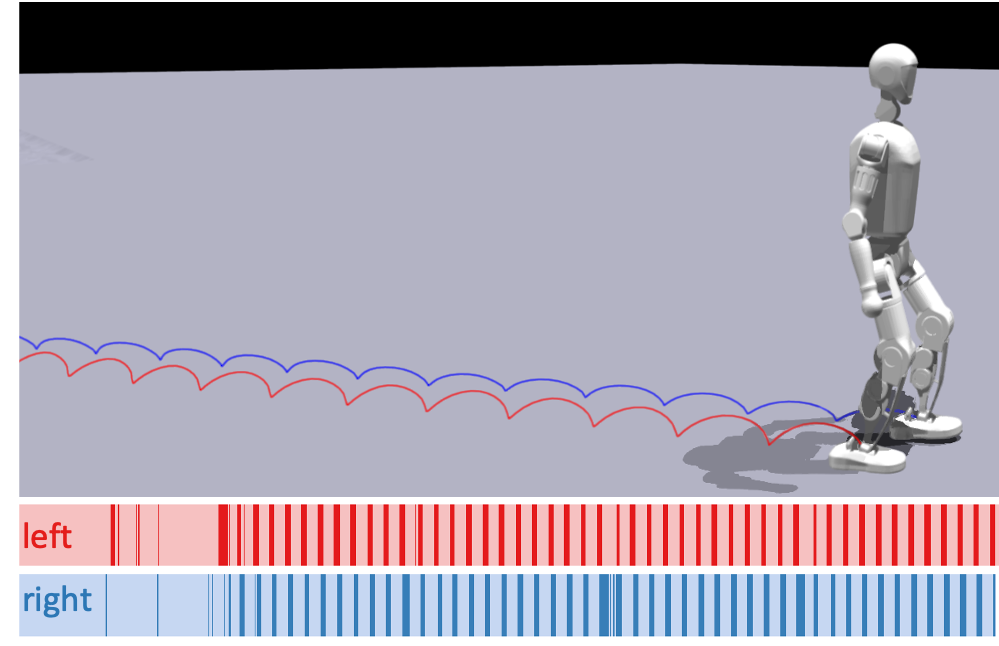} \\
c) seed policy at 1.0 frequency, 1.5m/s & d) seed policy at 2.0 frequency, 1.5m/s \\
\end{tabular}
}
\caption{\textbf{Phase scaling of the seed policy.} We vary the phase frequency and velocity command of the frozen seed policy on the Booster T1-23DoF robot. Foot trajectories and foot-contact plots are shown at $0.5$\,m/s and $1.5$\,m/s. From the trajectories and plots, we can see that phase scaling changes the contact rhythm and foot motion. This indicates that the seed walking policy contains reusable rhythmic structures. However, direct phase scaling reuses this structure only in a fixed manner and does not produce reliable high-speed gait growth beyond the seed policy's training range.}

\label{fig:fractal_scale}
\end{figure}

We first unfold the frozen seed walking policy to understand what locomotion structure it already contains. As shown in Fig.~\ref{fig:fractal_scale}, changing the phase frequency of the seed policy alters the rhythm and contact density of the resulting walking motion, indicating that the seed policy encodes reusable rhythmic structure. However, directly changing the phase frequency is not sufficient for command-adaptive gait growth. Associating with Fig.~\ref{fig:vel_tracking_acc} (a) in the main paper, when the target forward velocity increases beyond the seed policy's training range, such as $v_x=1.5$\,m/s compared with the training range of $[-1,1]$\,m/s, the phase-scaled seed still produces similar rhythmic and contact patterns but leads to large tracking error. These results suggest that the seed policy contains useful rhythmic primitives, but they must be used in a way that is adaptive rather than directly executed. This motivates the design philosophy of GaitWave, where the seed policy should serve as a stabilizing anchor that preserves its reusable rhythmic structure, while its seed-derived action waves can be adaptively recomposed in a certain way to move beyond the seed's original command range.

\subsection{Energy Comparison with Human Demonstration}
\label{app:human_demo_energy}

As discussed in the main paper, human-demonstration priors might provide rich motion structure in a snap, but they do not necessarily yield better energy behavior across the full command range. Here, we provide a closer comparison between GaitSpan and the human-demonstration baseline on Unitree G1 23DoF. As shown in Fig.~\ref{fig:humandemo_w_energy_compare}, GaitSpan consumes substantially less energy in low- and mid-speed regimes while remaining comparable near the OOD boundary. At higher OOD commands, both methods require increased energy, reflecting the higher physical demand of dynamic locomotion. These results suggest that human demonstrations are not always aligned with robot-specific objectives such as energy efficiency, which can be important for real-world deployment. Instead, growing gaits from the robot's own walking skill can produce energetically reasonable motion without relying on external motion priors.

\begin{figure}[t]
    \centering

    \begin{minipage}[t]{0.68\linewidth}
        \centering
        \vspace{0pt}

        \begin{subfigure}[t]{0.32\linewidth}
            \centering
            \includegraphics[height=0.12\textheight,width=\linewidth,keepaspectratio]{figs/imgs/humandemo/ablation_humandemo_abs.png}
            \caption{Tracking error}
            \label{fig:humandemo_abs}
        \end{subfigure}
        \hfill
        \begin{subfigure}[t]{0.32\linewidth}
            \centering
            \includegraphics[height=0.12\textheight,width=\linewidth,keepaspectratio]{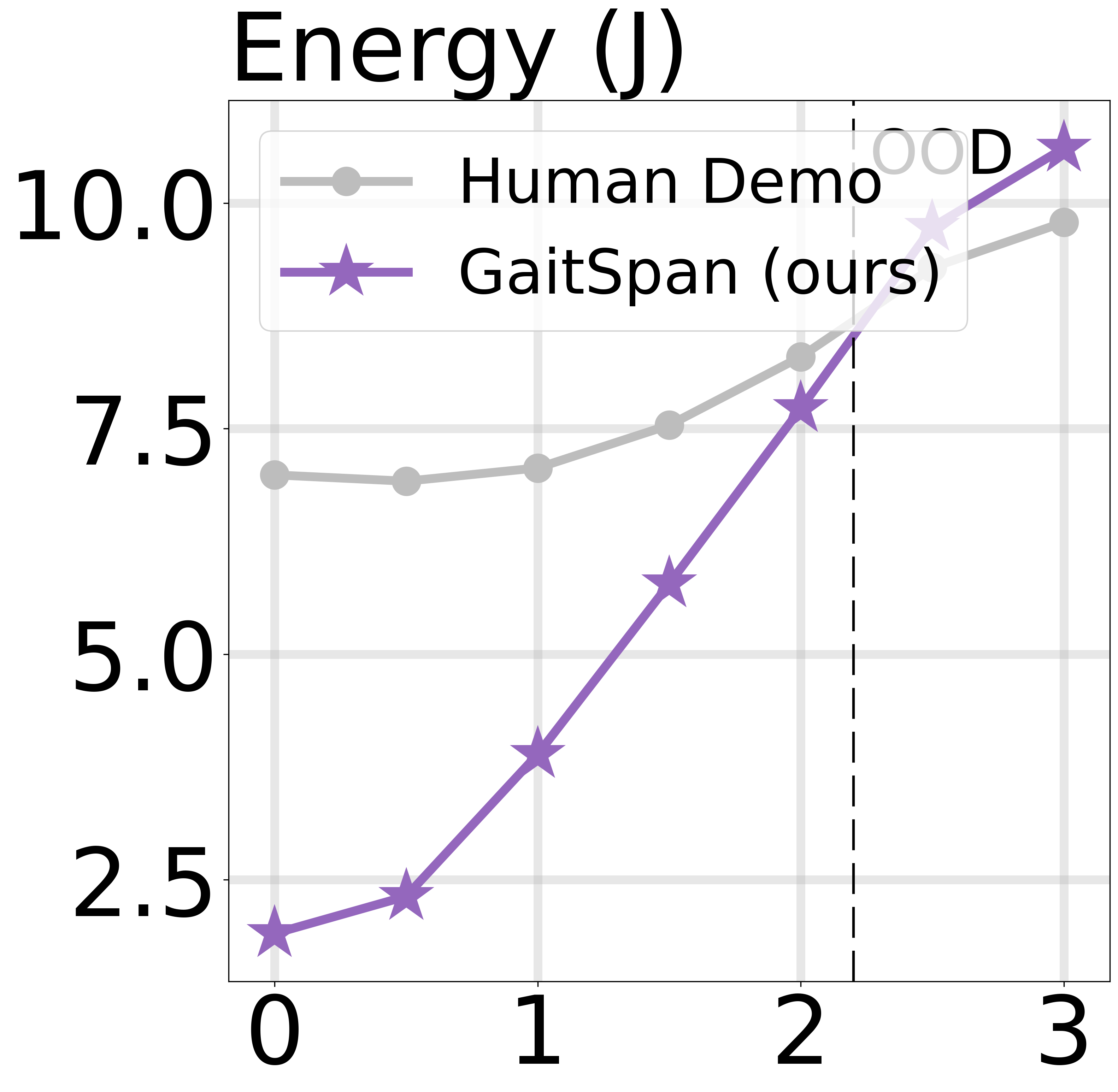}
            \caption{Energy}
            \label{fig:humandemo_energy}
        \end{subfigure}
        \hfill
        \begin{subfigure}[t]{0.32\linewidth}
            \centering
            \includegraphics[height=0.12\textheight,width=\linewidth,keepaspectratio]{figs/imgs/humandemo/ablation_humandemo_flight.png}
            \caption{Flight time}
            \label{fig:humandemo_flight}
        \end{subfigure}

        \captionof{figure}{
        \textbf{Comparison with the human-demonstration baseline on Unitree G1 robot.}
        We report tracking error, energy, and flight-time statistics for completeness, while the supplementary discussion mainly focuses on energy behavior.
        }
        \label{fig:humandemo_w_energy_compare}
    \end{minipage}
    \hfill
    \begin{minipage}[t]{0.28\linewidth}
        \centering
        \vspace{0pt}

        \includegraphics[height=0.12\textheight,width=\linewidth,keepaspectratio]{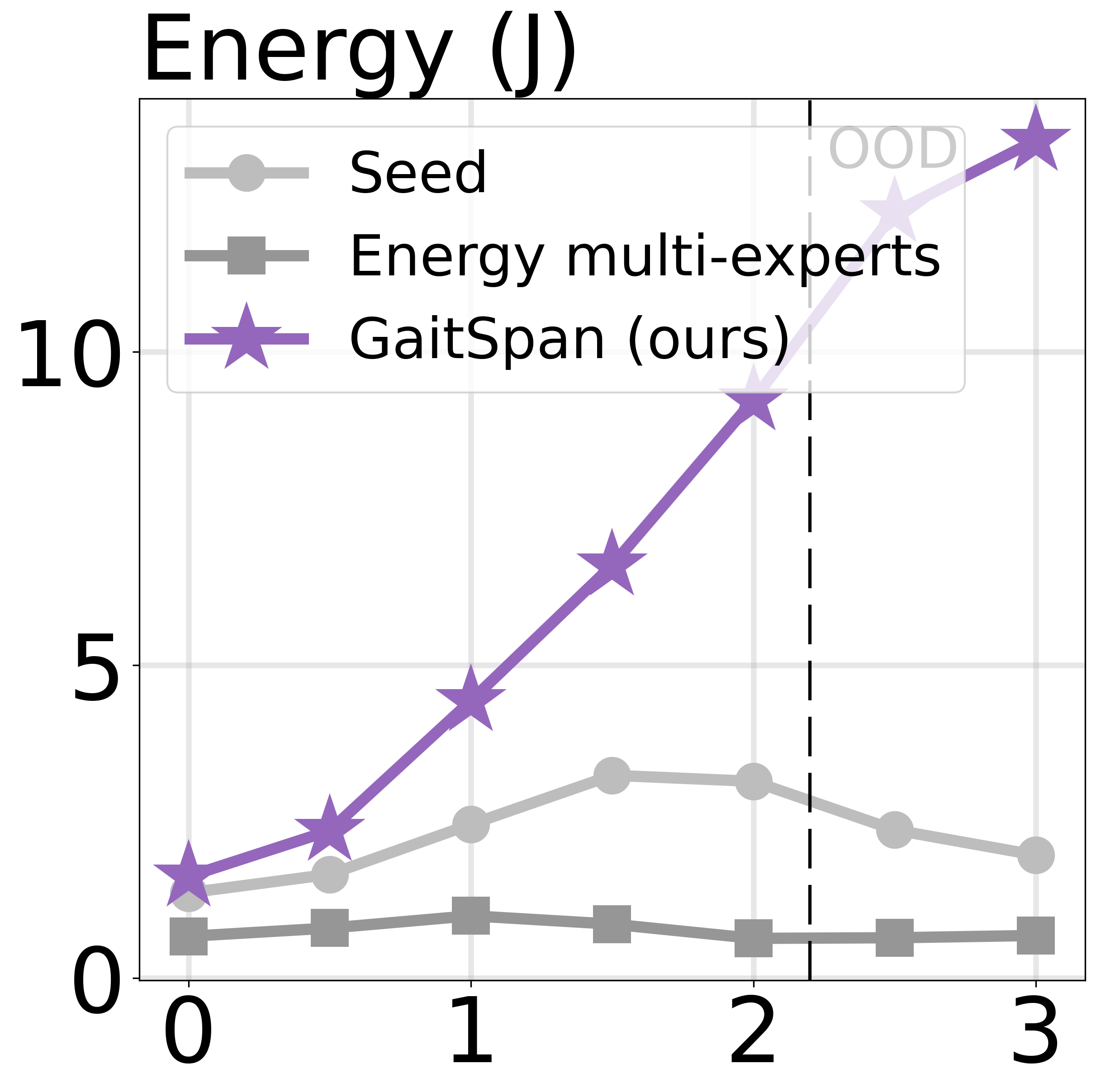}

        \captionof{figure}{
        \textbf{Energy comparison with the energy-based multi-expert baseline on Booster T1 robot.} GaitSpan achieves comparable energy consumption at low speeds.}
        \label{fig:energy_based_multi_expert_compare}
    \end{minipage}

\end{figure}


\subsection{Qualitative Motion Behavior Comparison}
\label{app:qualitative_compare}

\begin{figure}[h]
\centering\small
\newcommand{\squareimage}[1]{\includegraphics[width=\textwidth, height=0.182\textwidth]{#1}}

\resizebox{\textwidth}{!}{
\begin{tabular}{@{}l@{}}
a) GaitSpan (Ours) \\
\squareimage{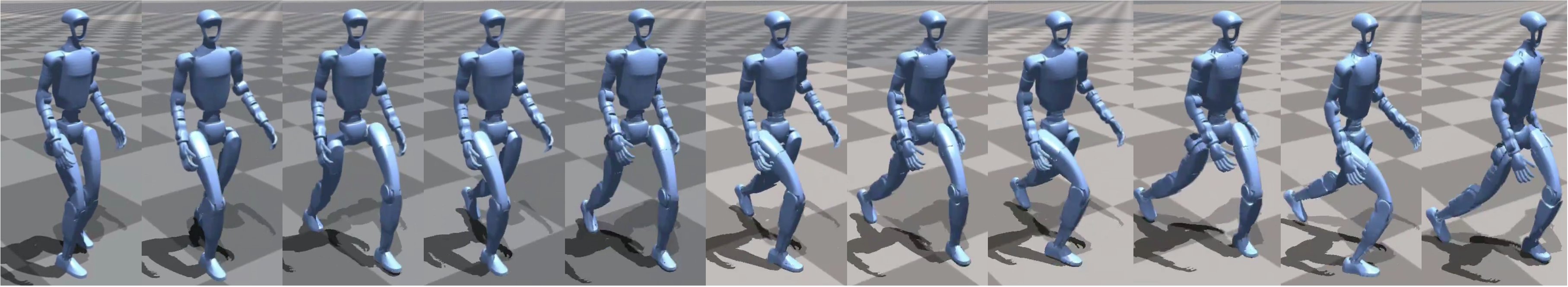} \\
b) Human Demonstration \\
\squareimage{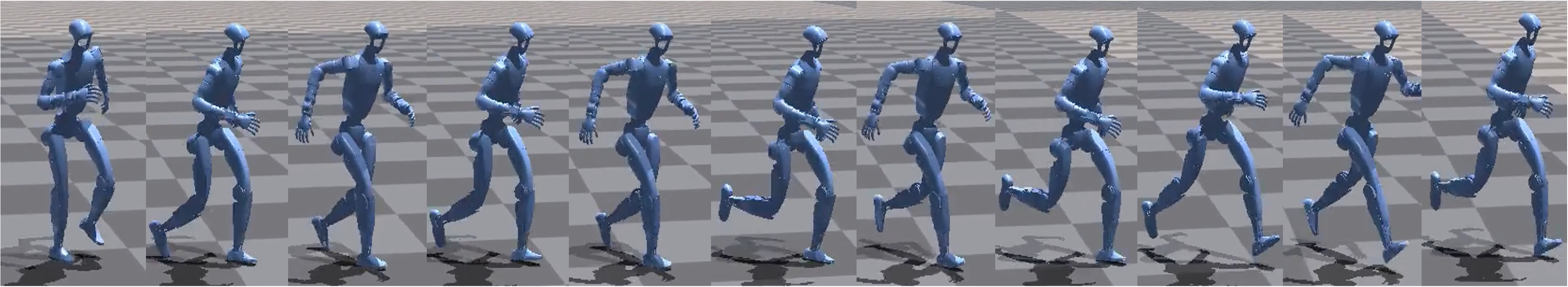}
\end{tabular}
}

\caption{\textbf{Qualitative comparison between GaitSpan and the human-demonstration baseline on Unitree G1 (29 DoF).} From left to right, the commanded speeds are $0.5$, $1.5$, and $2.5$\,m/s. Although the human-demonstration baseline exhibits more human-like posture and limb coordination, it largely preserves a similar running pattern across speeds, with limited command-dependent gait differentiation. This observation is consistent with Fig.~\ref{fig:energy_based_multi_expert_compare}, where the human-demonstration baseline incurs relatively high energy and substantial flight time even at low speeds. These results suggest that human-like imitation is not necessary for adaptive, speed-appropriate gait emergence.}
\label{fig:ours_vs_humandemo_g1_29_qualitative}
\end{figure}
\begin{figure}[t]
\centering\small
\newcommand{\squareimage}[1]{\includegraphics[width=\textwidth, height=0.182\textwidth]{#1}}
%
\resizebox{\textwidth}{!}{
\begin{tabular}{@{}l@{}}
a) GaitSpan (ours) \\
\squareimage{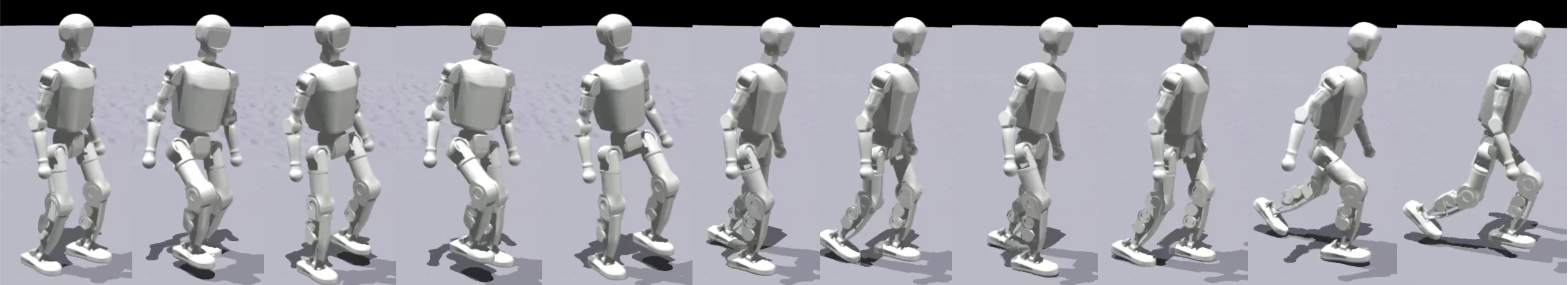} \\
b) Energy-based multi-experts \\
\squareimage{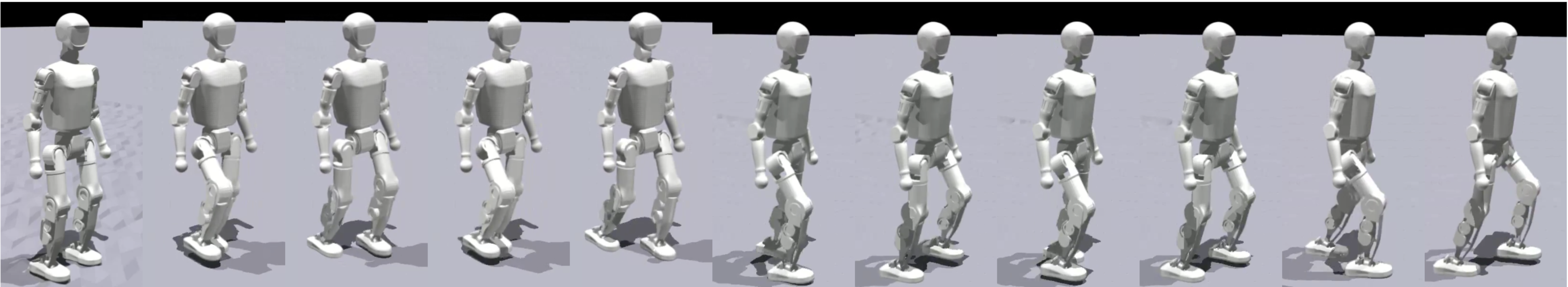} \\
c) Training from Scratch \\
\squareimage{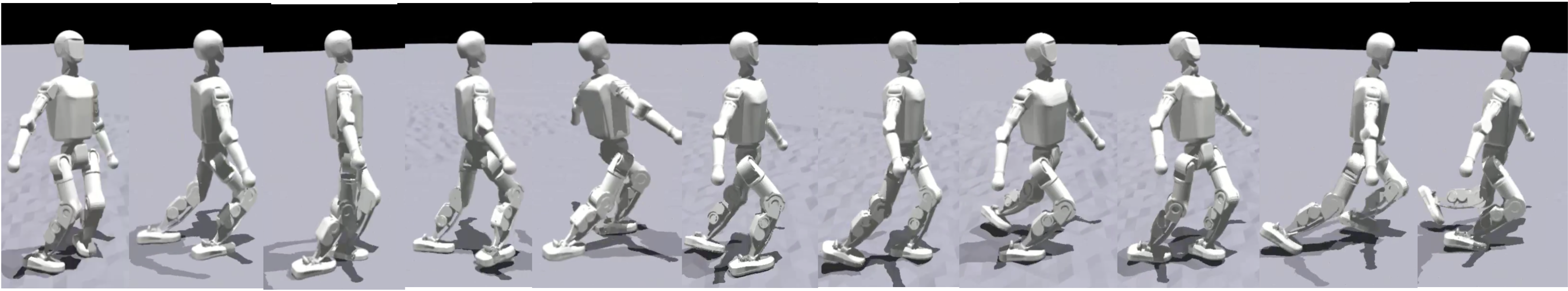} \\
d) Fine-Tuning \\
\squareimage{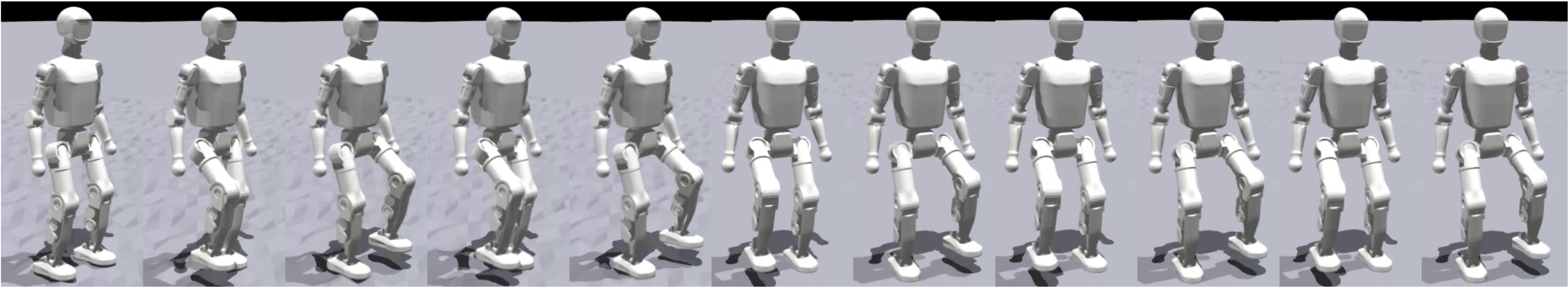}
\end{tabular}
}
\caption{\textbf{Qualitative baseline comparison.} Under increasing command speeds ($0 \rightarrow 0.5 \rightarrow 1.5 \rightarrow 2.5$\,m/s), GaitSpan produces smooth walking-to-dynamic gait transitions, while the energy-based multi-expert baseline remains conservative, training from scratch yields less coherent motion, and fine-tuning collapses into stepping-in-place behavior.}

\label{fig:baseline_sidebyside}
\end{figure}

We further provide qualitative comparisons (shown in Fig~\ref{fig:ours_vs_humandemo_g1_29_qualitative} and Fig~\ref{fig:baseline_sidebyside}) between GaitSpan and representative baselines to illustrate their behavioral differences beyond scalar metrics.

\noindent\textbf{GaitSpan vs Priors from Human Demonstration.}
As shown in Fig.~\ref{fig:ours_vs_humandemo_g1_29_qualitative}, human-demonstration-based policy (\textit{i.e.}, AMP~\cite{2021-TOG-AMP} in our experiments) can produce natural human-like posture and expressive limb coordination by importing motion priors from human data. However, the resulting motion pattern changes only modestly across commanded speeds, exhibiting limited command-dependent gait differentiation. AMP exhibits substantial flight time even at walking speeds; this is complementary with Fig.~\ref{fig:humandemo_w_energy_compare}'s finding, where the human-demonstration baseline incurs higher energy consumption over most of the evaluated range. In contrast, GaitSpan develops speed-appropriate locomotion from the robot's own walking seed, where it preserves grounded walking behavior at low speeds and progressively reorganizes rhythm, stride, and contact timing as commands enter jogging- and running-like regimes. These results suggest that visually human-like motion is not necessarily optimal for robot locomotion, as effective gaits must also reflect the robot's physical properties and command-dependent objectives.

\noindent\textbf{GaitSpan vs. Energy-Based Multi-Experts.}
Energy-shaped multi-experts can produce low-energy behaviors within individual speed regimes, but their motions are assembled from separately trained skills. As shown in Fig.~\ref{fig:baseline_sidebyside}, this baseline maintains low energy consumption across speeds, but remains conservative and produces limited dynamic flight. In contrast, GaitSpan grows a continuous gait family from one walking seed, producing smoother changes in rhythm, stride, and contact timing as speed increases. Meanwhile, from the energy cost aspect (as shown in the Fig~\ref{fig:energy_based_multi_expert_compare}),  at low speeds, GaitSpan maintains energy consumption close to the seed and energy-based baseline; at higher speeds, it uses additional energy to realize the dynamic motion required by the command, rather than suppressing gait emergence for the sake of energy minimization.

\noindent\textbf{GaitSpan vs. Training from Scratch.}
Policies trained from scratch over the full command range must simultaneously discover stable walking and dynamic high-speed locomotion. Qualitatively, this leads to motion spectrum collapse that fails to develop clear jogging- and running-like patterns (as shown in Fig.~\ref{fig:baseline_sidebyside}). GaitSpan avoids this difficulty by anchoring learning to a stable seed skill, while allowing exploration beyond walking without losing the structure needed for balance and support.

\noindent\textbf{GaitSpan vs. Fine-Tuning.}
As shown in Fig.~\ref{fig:baseline_sidebyside}, directly fine-tuning the walking policy adapts the original skill, but can overwrite the low-speed behavior that made the seed stable in the first place. This produces degraded walking behavior while still failing to reliably generate dynamic high-speed motion. GaitSpan instead freezes the seed policy and expands around it, preserving the reusable structure of walking while learning additional rhythm, stride, and residual adaptations for faster regimes.

\section{Additional Ablation Studies}

\label{appendix:ablation}
In this section, we provide additional ablation studies to better understand the detailed designs of GaitWave and H-SLIP that connect learned rhythm composition with physically guided stride shaping (Sec.~\ref{sec_ablation_details}). Additionally, we further examine the universality of GaitSpan across embodiments (Sec.~\ref{sec:universality}) beyond the results covered in the main paper.


\subsection{Looking Inside GaitWave and H-SLIP}
\label{sec_ablation_details}

In this subsection, we provide additional evidence for how GaitSpan grows a walking skill into richer locomotion regimes. We first examine the internal behavior of GaitWave, including how seed-derived action waves are combined, how their coefficients change with commanded speed, and how this composition preserves stable seed-like behavior in the low-speed regime while enabling expansion at higher speeds. We then analyze H-SLIP by comparing single-, double-, and three-level virtual-leg designs, showing why hierarchical dynamic shaping is needed to turn rhythm-expanded motions into more dynamic stride patterns.

\subsubsection{GaitWave: From Seed Actions to Wave Composition}
\label{app:wave_composition}

\noindent\textbf{Ablation on how seed action waves combine.}
\begin{figure*}[t]
\small\centering
\newcommand{\squareimage}[1]{\includegraphics[width=0.245\linewidth, height=0.245\linewidth]{#1}}
\newcommand{\colspace}{\hspace{8pt}}
\newcommand{\colspaceclose}{\hspace{1pt}}
\newcommand{\colspacenone}{\hspace{0pt}}
\newcommand{\tablerow}[1]{
\squareimage{figs/imgs/ablation/#1_abs.png} &
\squareimage{figs/imgs/ablation/#1_flight.png} 
}
\resizebox{\textwidth}{!}{
\renewcommand{\arraystretch}{0.3}
\begin{tabular}
{@{}c@{\colspaceclose}c@{\colspace}c@{\colspaceclose}c@{}}
%
%
\tablerow{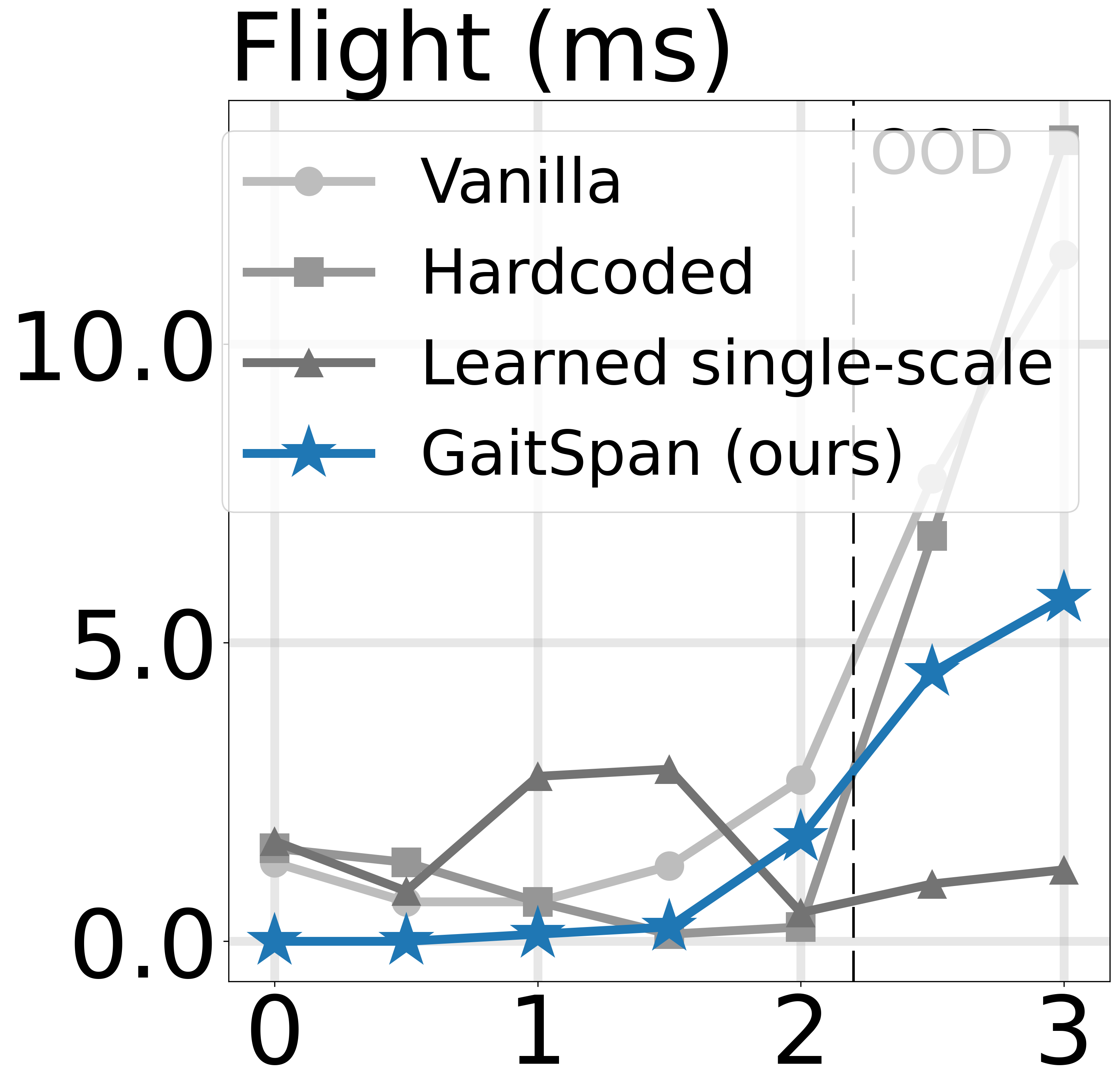} & \tablerow{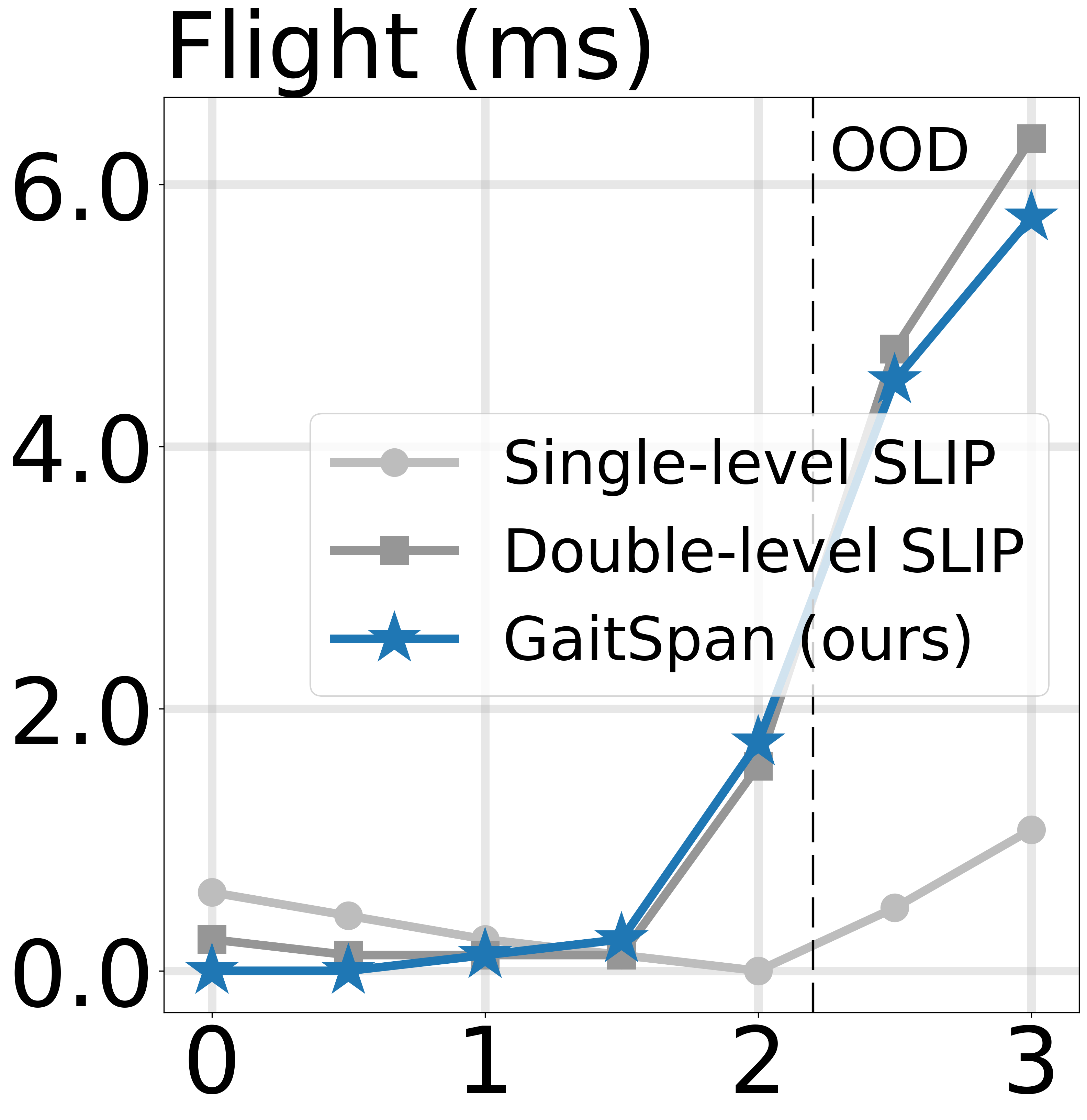} \\
\multicolumn{2}{c}{a) different combination of wave basis} & \multicolumn{2}{c}{b) SLIP reward}
\end{tabular}
}
\caption{\textbf{Ablation on details of GaitWave and H-SLIP designs on Booster T1 robot (23 DoF).} 
\textbf{(a)} Hierarchical learned wave composition improves tracking and produces controlled flight compared with the variants. 
\textbf{(b)} H-SLIP better balances tracking accuracy and dynamic flight with three-level virtual legs.}

\label{fig:suppl_ablation_combineway}
\end{figure*}
We compare with three variants: a vanilla, {\textit{i.e.,}}seed-residual model without phase-scaled wave composition, a hardcoded phase-scaling variant that changes with speed, and a learned single-scale variant with 16 bins. As shown in Fig.~\ref{fig:suppl_ablation_combineway} (a), directly using the seed policy or manually scaling its phase is insufficient for reliable high-speed gait growth, as these variants either degrade tracking under OOD commands or produce poorly controlled flight. Learning a single-phase scale improves flexibility, but still lacks the command-dependent expressiveness (low flight time in the figure) needed to span the full speed range. In contrast, GaitSpan maintains low tracking error while producing increasing, controlled flight time at high speeds, showing that hierarchical learned composition of seed-derived action waves is essential for growing the walking seed into dynamic gait regimes. We provide additional qualitative comparison to other alternatives in Fig.~\ref{fig:ablate_singlemlp_hardmixture}. Associating it with Fig~\ref{fig:baseline_sidebyside} (a) and Fig.~\ref{fig:suppl_ablation_combineway} (a) provides further insight. Under the surface, when one shared predictor is used over the entire command domain, learning is dominated by the larger action deviations required at high speeds, causing overly aggressive wave mixtures to leak into the walking regime. Dividing the command space into independent regions reduces this global bias, but removes continuity between neighboring regions and introduces visible boundary artifacts when combination is rigidly manual scaling. Our hierarchical multi-resolution memory resolves this trade-off by sharing coarse trends across speeds while retaining local command-dependent refinements, enabling distinct gait regimes to emerge without sacrificing smooth transitions.

\begin{figure}[h]
\centering\small
\newcommand{\squareimage}[1]{\includegraphics[width=\textwidth]{#1}}
%
\resizebox{\textwidth}{!}{
\begin{tabular}{@{}l@{}}
a) Single global mapping \\
\squareimage{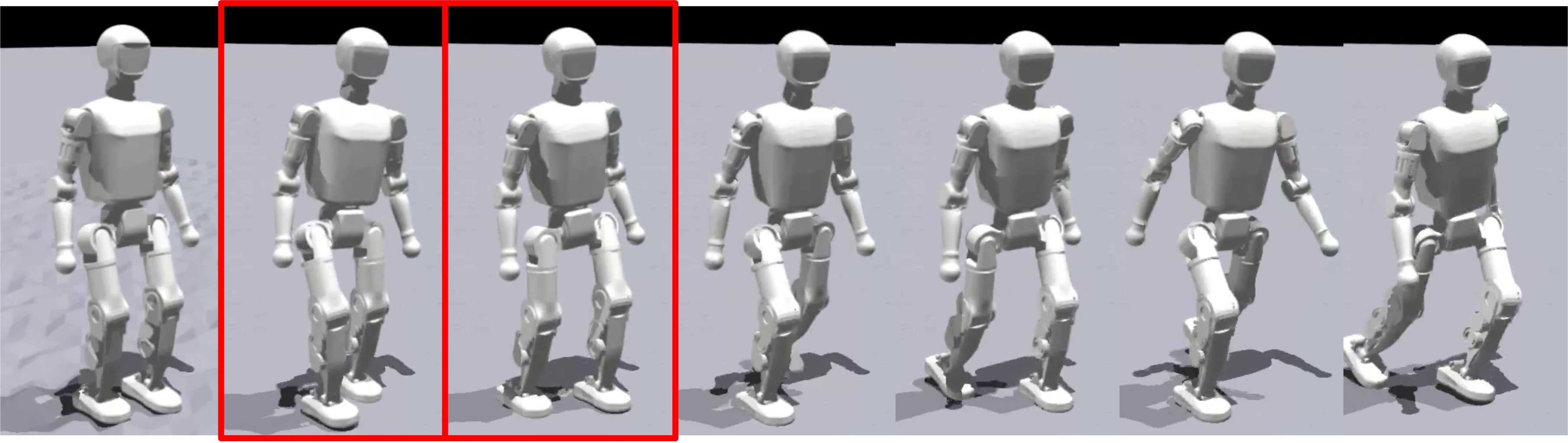} \\
b) Hard command partitioning \\
\squareimage{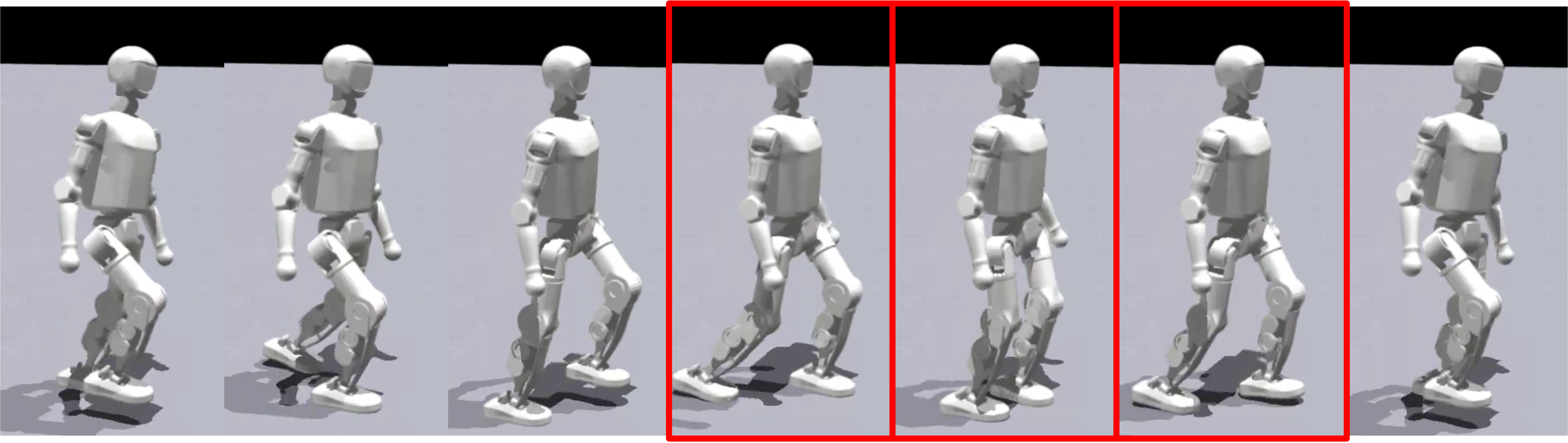} \\
\end{tabular}
}
\caption{\textbf{Qualitative ablation of our hierarchical multi-resolution coefficient memory.}
a) Using a single global mapping from command speed to coefficients leads to a bias toward high-velocity commands, ignoring walking gaits at lower speeds because the single mapping is too coarse to cover the broad command range, as marked by red boxes.
b) Partitioning the command range introduces hard transitions at partition boundaries, causing abrupt gait switches, as marked by red boxes.}

\label{fig:ablate_singlemlp_hardmixture}
\end{figure}

\begin{figure}[h]
\newcommand{\close}{\hspace{1pt}}
\newcommand{\heatmapimage}[1]{\includegraphics[width=0.62\textwidth, , height=0.16\textwidth, valign=m]{#1}}
\newcommand{\squareimage}[1]{\includegraphics[width=0.17\textwidth, height=0.17\textwidth, valign=m]{#1}}
\newcommand{\tablerow}[1]{
\heatmapimage{figs/imgs/heatmap/fractal_beta_heatmap_#1.png} &
\squareimage{figs/imgs/heatmap/heatmap_#1_seed.png} &
\squareimage{figs/imgs/heatmap/heatmap_#1_ours.png}
}
\centering\small
\resizebox{\textwidth}{!}{
\begin{tabular}{@{}l@{\close}c@{\close}c@{}}
a) Booster T1 23DoF & Seed & GaitSpan (ours) \\
\tablerow{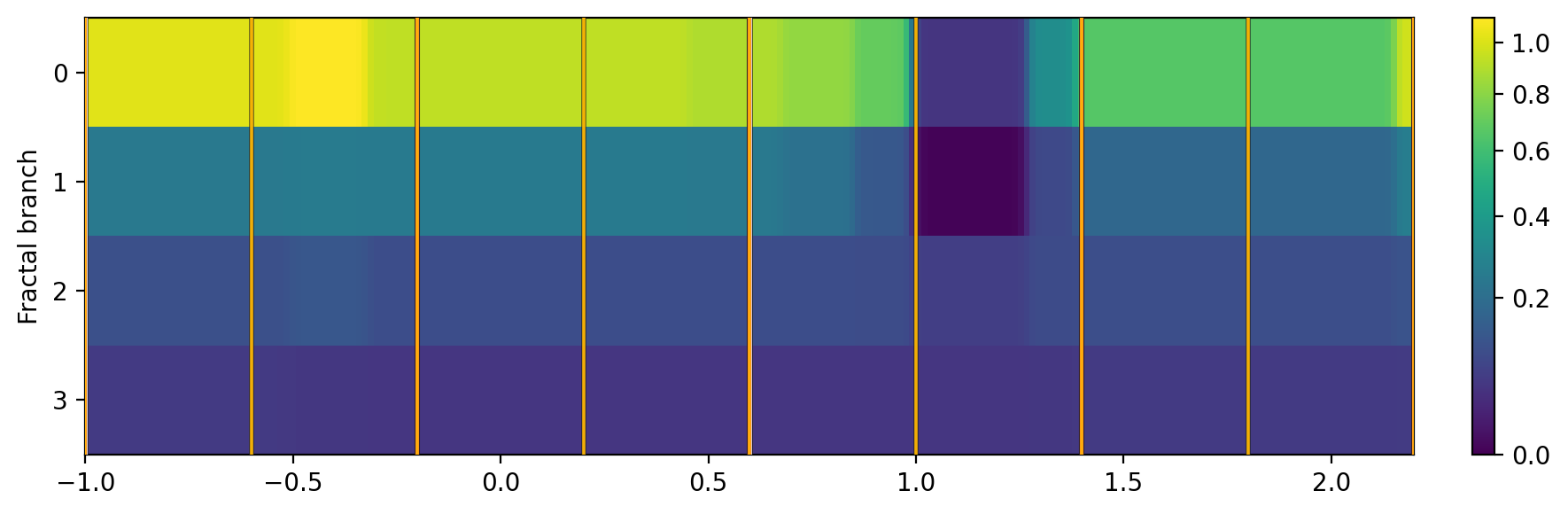} \\
\multicolumn{3}{@{}l}{b) Booster K1 22DoF} \\
\tablerow{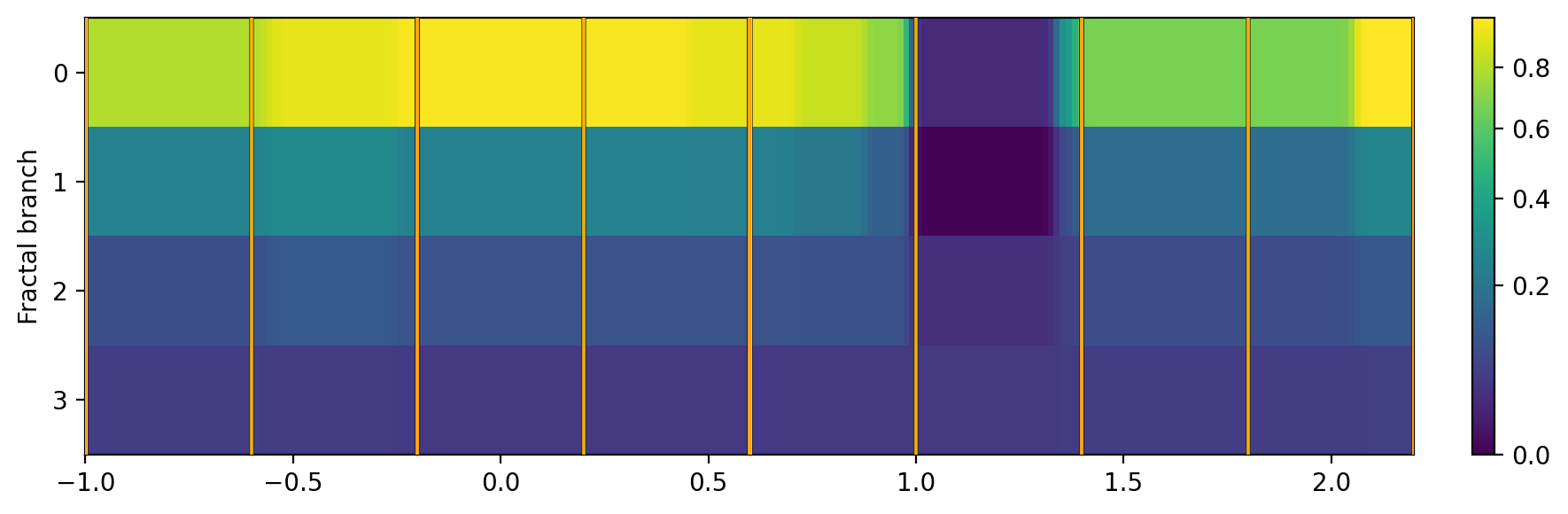} \\
\multicolumn{3}{@{}l}{c) Unitree G1 23DoF} \\
\tablerow{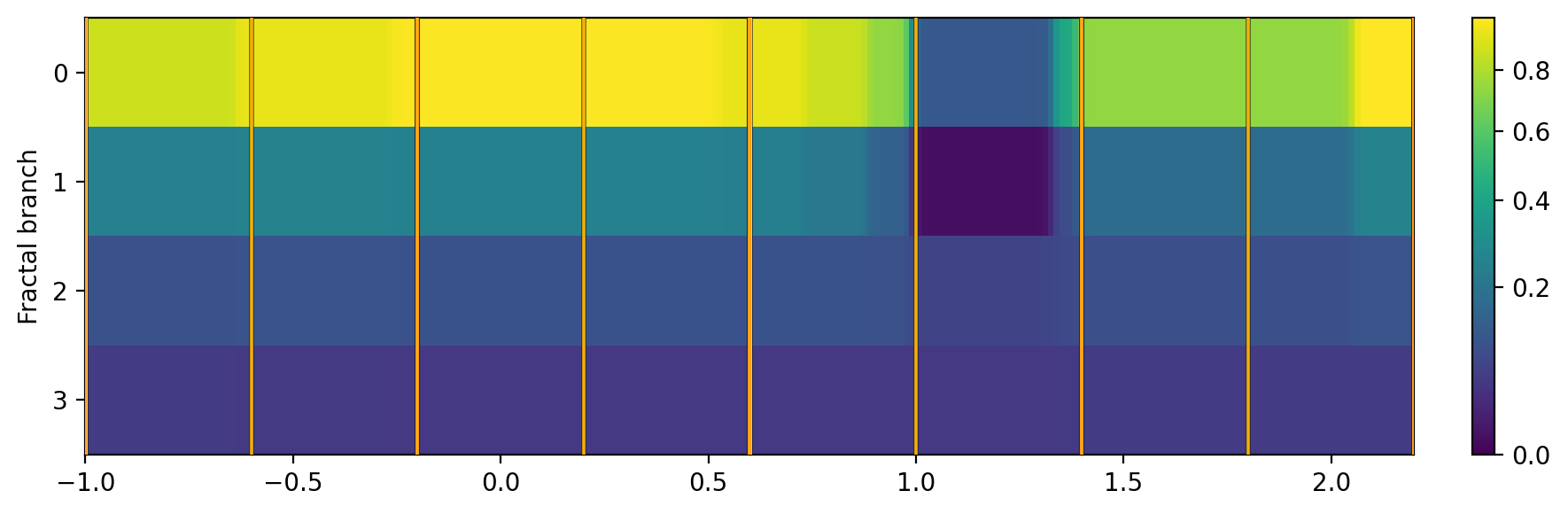}
\end{tabular}
}
\caption{\textbf{Heatmap of wave coefficients and seed behavior preservation at low speed.} Left: learned GaitWave coefficients across commanded speeds for Booster T1, Booster K1, and Unitree G1. Right: at $0.5$\,m/s, GaitSpan preserves walking behavior visually close to the frozen seed policy across embodiments.}

\label{fig:fractal_heatmap}
\end{figure}

\noindent\textbf{Speed-dependent use of wave coefficients.} Next, we further investigate how the learned wave coefficients vary with commanded speed. Fig.~\ref{fig:fractal_heatmap}(a-c) visualizes the normalized activation of different wave branches across the command range for three humanoid embodiments. The figure reveals several insights. \textbf{1)} Lower-order branches consistently dominate, especially in the low-speed regime, indicating that GaitWave strongly relies on seed-like rhythmic structure when walking behavior is sufficient.  This is further supported by the low-speed qualitative comparison in Fig.~\ref{fig:fractal_heatmap} (d) (right side of the figure), where GaitSpan preserves walking behavior visually close to the frozen seed policy at $0.5$\,m/s across different embodiments. For higher speed, the lower-order branches serve as stable anchors. \textbf{2)} The coefficient distribution is not fixed across speed. As the command moves toward faster regimes, branch activations reorganize, especially near the transition region around the upper walking range. This shows that GaitWave adaptively changes the composition of seed-derived waves rather than applying a uniform phase scaling. \textbf{3)} The overall trend is consistent across different embodiments, suggesting a general mechanism for speed-dependent gait growth. At the same time, the exact activation patterns differ across embodiments, reflecting morphology-specific choices in how each robot reallocates wave components to realize faster locomotion. Overall, these results support the view that GaitWave preserves seed-like behavior at low speeds while selectively reallocating wave contributions to support richer locomotion at higher speeds.

\subsubsection{H-SLIP: Why We Need Hierarchical SLIP Reward}
\label{app:hslip_hierarchy}
We further ablate the hierarchy design in H-SLIP by comparing single-level, double-level, and the final three-level virtual-leg shaping. As shown in Fig.~\ref{fig:suppl_ablation_combineway}(b), single-level SLIP shaping provides only a coarse root--foot dynamic prior. It can encourage limited flight, but ignores the articulated motion of the leg and therefore cannot sufficiently coordinate local compression and rebound across the limb, leading to weaker dynamic gait emergence. Double-level SLIP increases flight emergence by adding local structure to knee and foot, but does not explicitly constrain these local motions to form a coherent whole-body stride. Consequently, the two segments may satisfy their own objectives while producing poorly coordinated root motion, leading to degraded tracking under high-speed and OOD commands.

In contrast, the three-level H-SLIP  better balances global and local dynamics. The root--foot level captures whole-body support and aerial transition, while the upper- and lower-leg levels shape limb-level compression and rebound. This hierarchy allows the policy to develop flight-producing behaviors without sacrificing tracking accuracy. These results show that H-SLIP is not merely adding a running-style reward; its hierarchical structure is important for turning rhythm-expanded motions into physically meaningful dynamic strides.

\subsection{The Universality of GaitSpan}
\label{sec:universality}
\subsubsection{Main Component Generalization Across Robots}

\begin{figure*}[h]
\small\centering
\newcommand{\squareimage}[1]{\includegraphics[width=0.245\linewidth, height=0.245\linewidth]{#1}}
\newcommand{\colspace}{\hspace{8pt}}
\newcommand{\colspaceclose}{\hspace{1pt}}
\newcommand{\colspacenone}{\hspace{0pt}}
\newcommand{\tablerow}[1]{
\squareimage{figs/imgs/ablation_g1/#1_abs.png} &
\squareimage{figs/imgs/ablation_g1/#1_flight.png} 
}
\resizebox{\textwidth}{!}{
\renewcommand{\arraystretch}{0.3}
\begin{tabular}
{@{}c@{\colspaceclose}c@{\colspace}c@{\colspaceclose}c@{}}
\tablerow{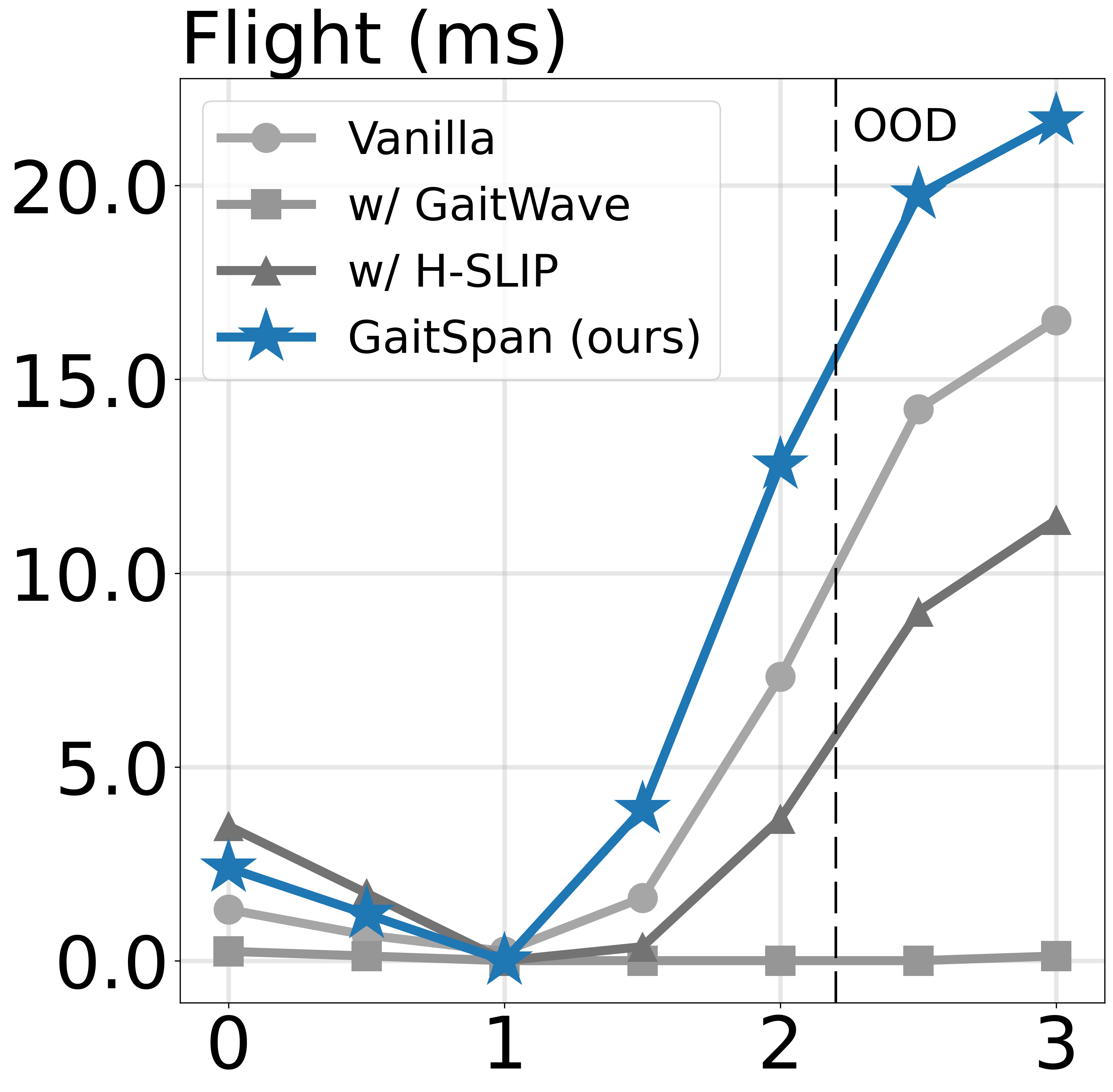} & \tablerow{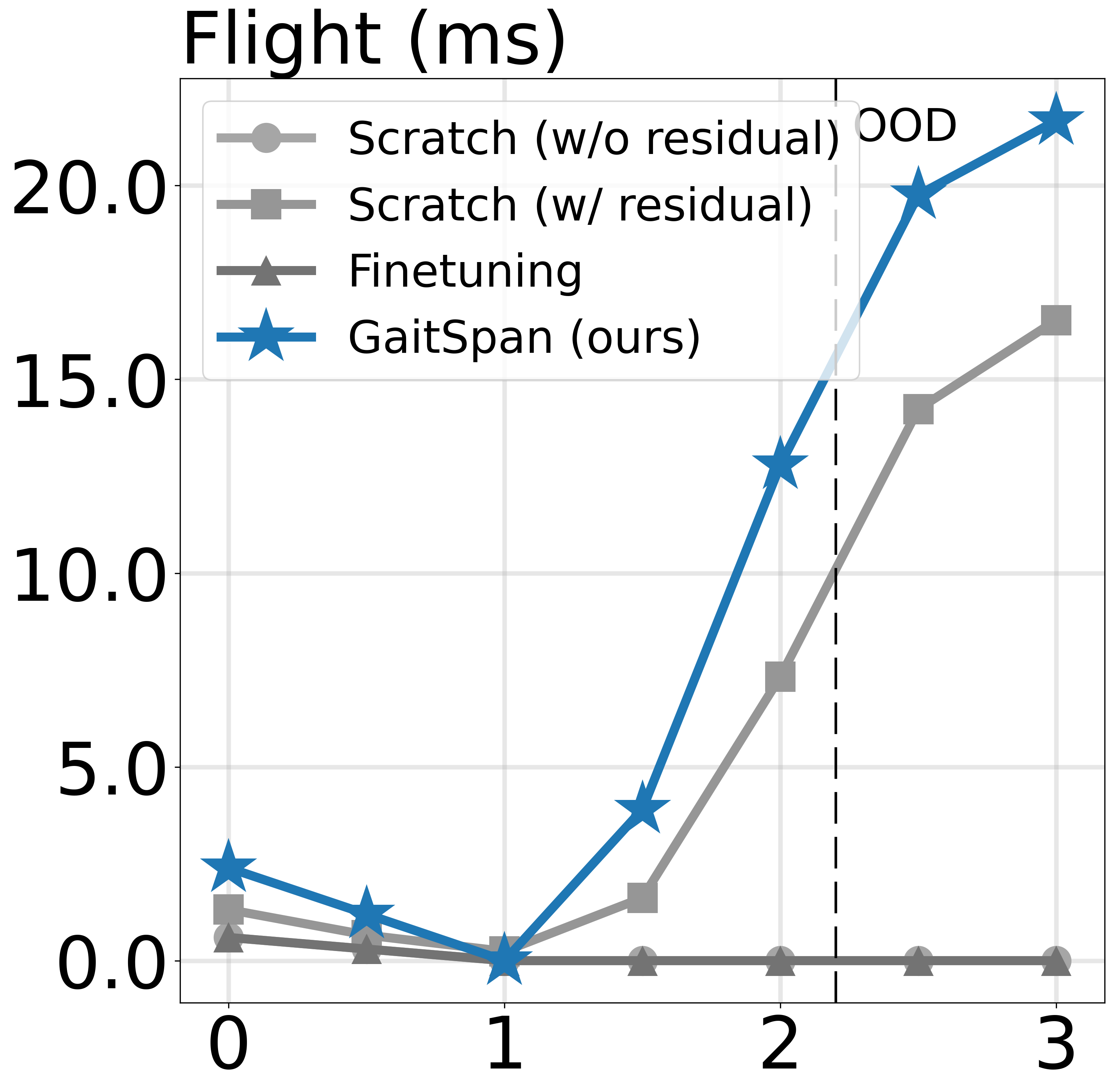} \\
\multicolumn{2}{c}{a) main components} & \multicolumn{2}{c}{b) training strategies} \\ [8pt]
\tablerow{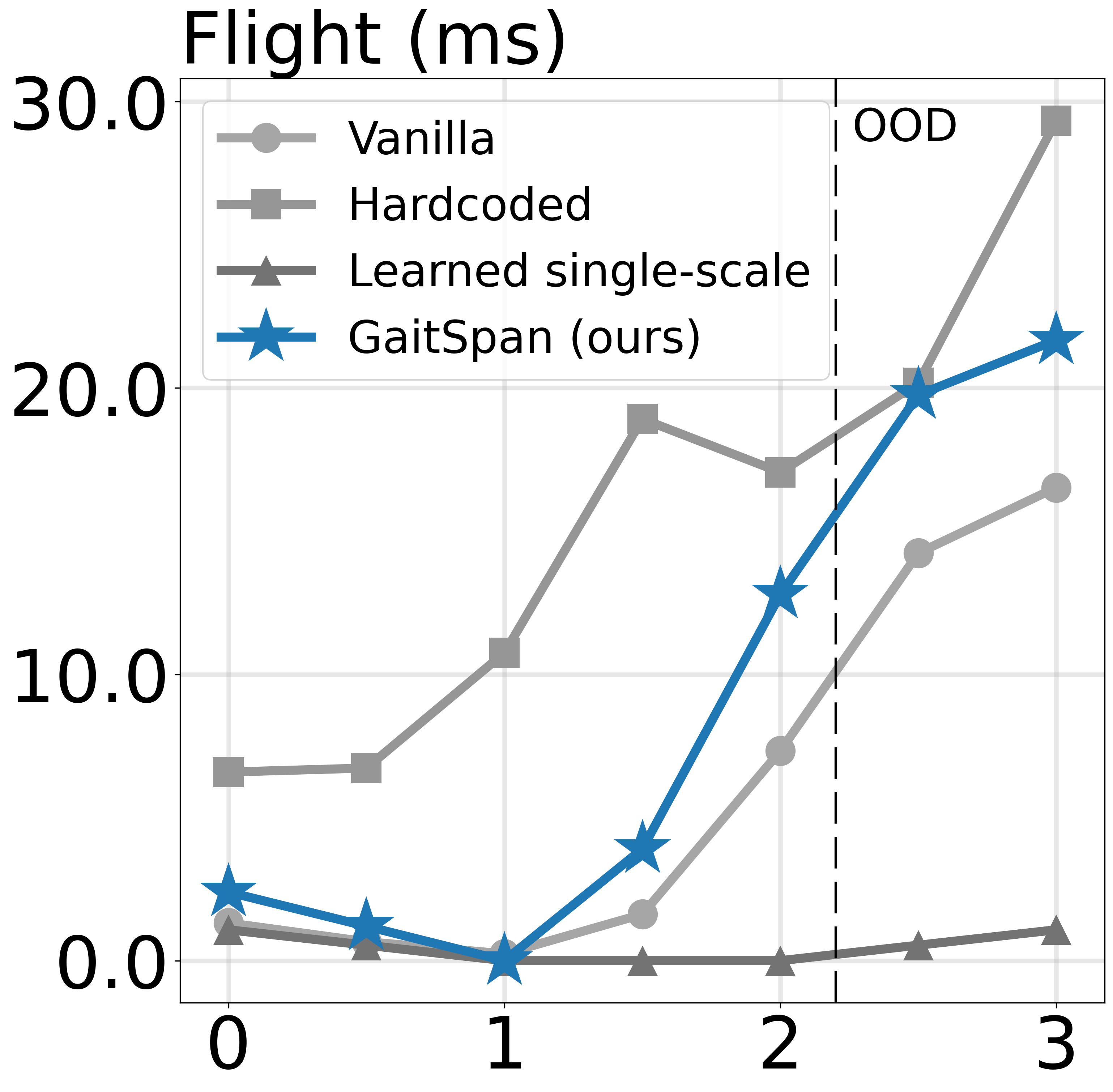} & \tablerow{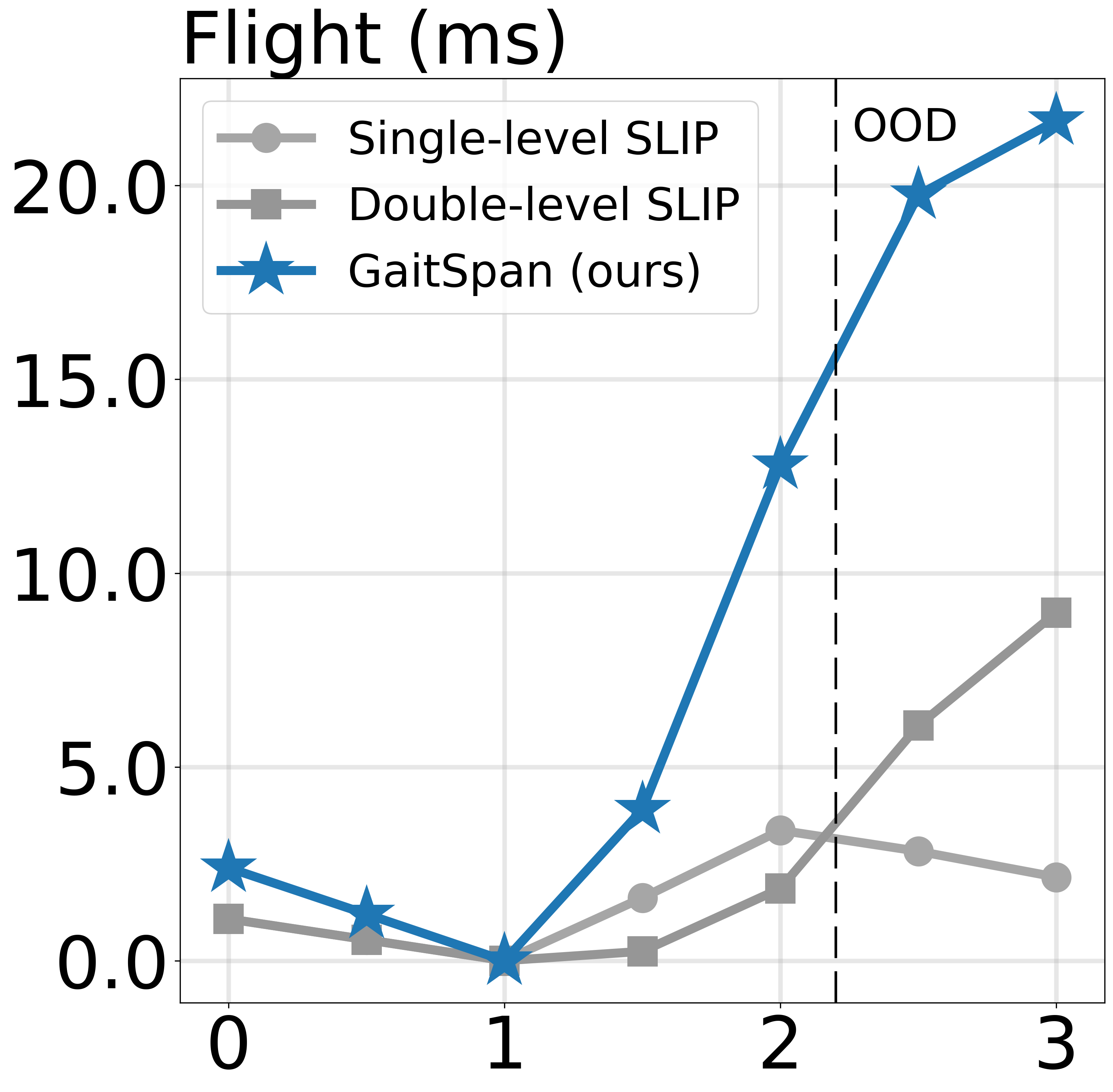} \\
\multicolumn{2}{c}{c) fractal basis} & \multicolumn{2}{c}{d) SLIP reward}
\end{tabular}
}
\caption{\textbf{Additional ablation on Unitree G1 (23 DoF).} The trends are consistent with Booster T1 in the main paper and Fig~\ref{fig:suppl_ablation_combineway} in the Supp.}
\label{fig:suppl_ablation_g1}
\end{figure*}
We evaluate whether the main components of GaitSpan remain effective across robot embodiments. The experiment repeats the component, training-strategy, wave-composition, and H-SLIP hierarchy ablations on an additional humanoid embodiment, Unitree G1 robot (23 DoF). 

Fig.~\ref{fig:suppl_ablation_g1} reveals several key findings. \textbf{1)} The overall trends are consistent with those observed on Booster T1: GaitWave and H-SLIP remain complementary, and the training strategy continues to matter. The full GaitSpan model achieves the best balance between tracking accuracy and dynamic flight emergence, while single-component variants, from-scratch training, and direct fine-tuning fail to achieve both simultaneously. \textbf{2)} Compared with Booster T1, Unitree G1 shows more significant sensitivity to dynamic shaping, making the benefit of hierarchical H-SLIP more visible. In particular, the gap between single-/double-level SLIP shaping and the full three-level H-SLIP is larger, suggesting that articulated virtual-leg hierarchy is especially important for coordinating global support and local limb dynamics on this embodiment. Together, these results indicate that GaitSpan's design generalizes across robots.

\begin{figure}[t]
\centering\small
\newcommand{\squareimage}[1]{\includegraphics[width=\textwidth]
{figs/imgs/foottrajectory/foottraj_#1.png}}
%
\resizebox{\textwidth}{!}{
\begin{tabular}{@{}l@{}}
a) Booster T1 23DoF \\
\squareimage{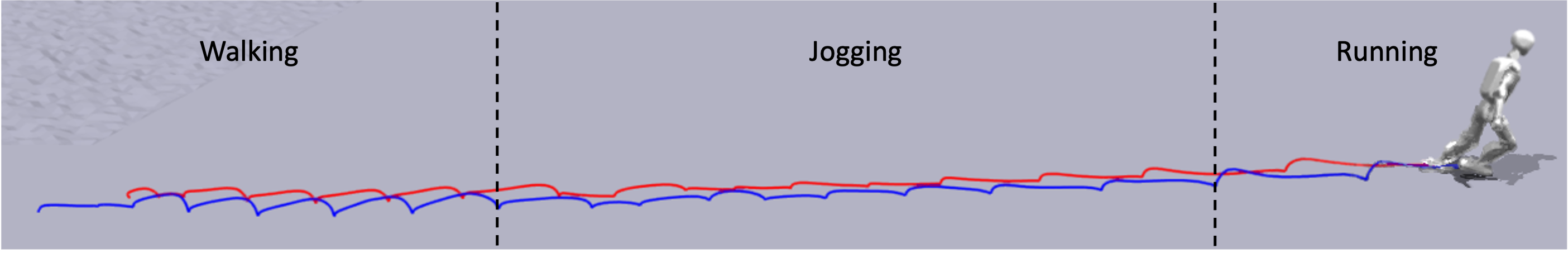} \\
b) Booster T1 29DoF \\
\squareimage{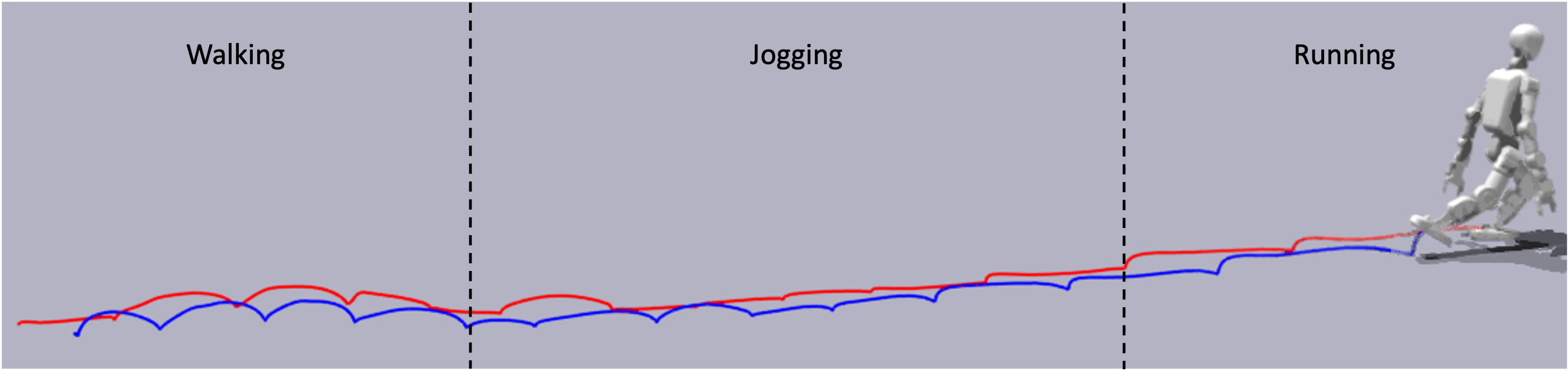} \\
c) Booster K1 22DoF \\
\squareimage{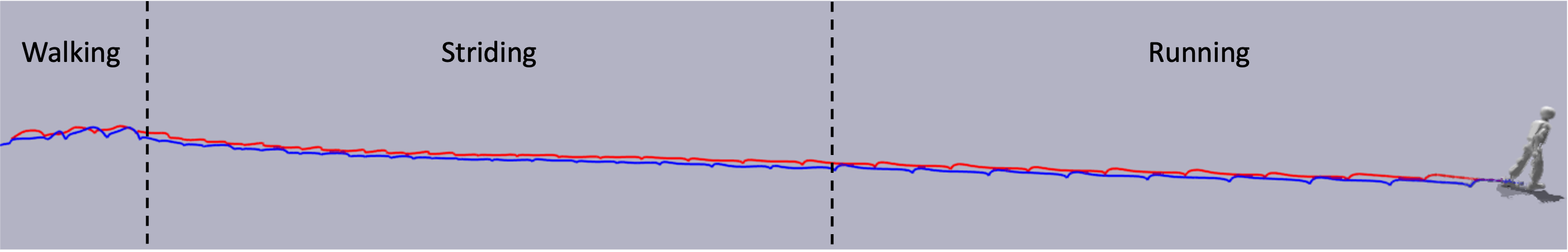} \\
d) Unitree G1 23DoF \\
\squareimage{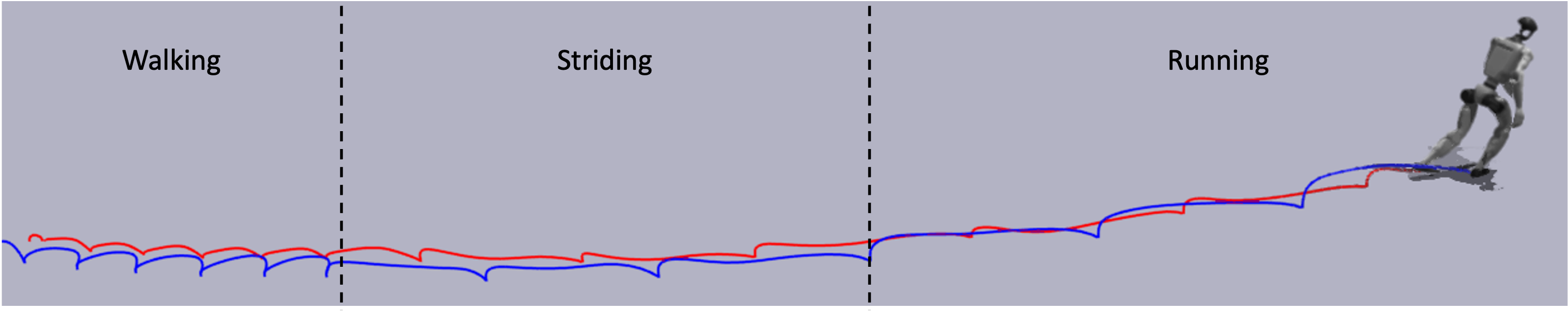} \\
e) Unitree G1 29DoF \\
\squareimage{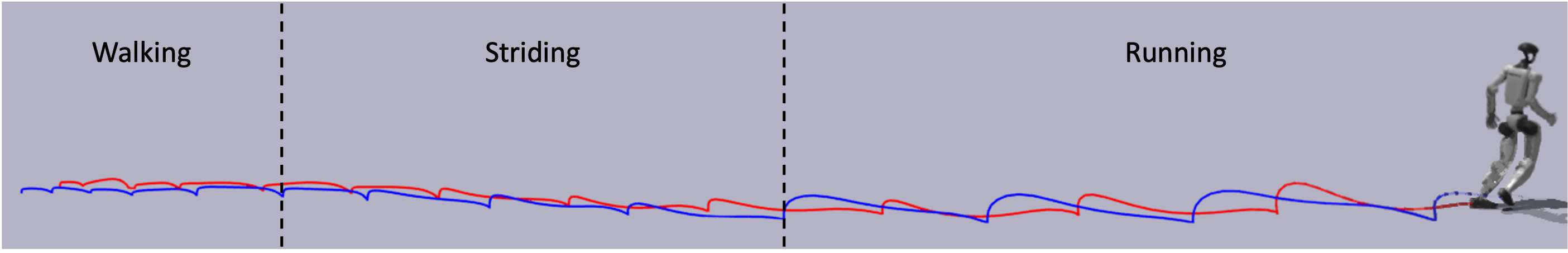} \\
\end{tabular}
}
\caption{\textbf{Foot trajectories across speed regimes and embodiments.} The command speed increases from $0$ to $2.5$\,m/s progressively.}

\label{fig:foot_trajectory}
\end{figure}

\subsubsection{Additional Emergent Behavior Analysis Across Robots}

\noindent\textbf{Foot trajectories across speed regimes.} Fig.~\ref{fig:foot_trajectory} visualizes left and right foot trajectories across five humanoid embodiments as commanded speed increases. \textbf{1)} Across robots, the trajectories all evolve from short, low-clearance walking steps to longer and more dynamic patterns in striding, jogging, and running-like regimes. This progression indicates that GaitSpan does not merely speed up a fixed walking gait; instead, it reorganizes stride length, foot clearance, and swing geometry as the locomotion regime changes. \textbf{2)} The exact trajectory shapes differ across embodiments, reflecting the flexibility of GaitSpan that enables morphology-specific gait expression under the same skill-growth framework. For example, Booster K1 is a smaller humanoid with $95cm$ height, so its emerged gait appears more compact, whereas larger embodiments such as Booster T1 and Unitree G1 exhibit more significant stride extension and swing variation at higher speeds.

\noindent\textbf{Additional Foot-Contact Results Across Embodiments.}
Fig.~\ref{fig:footcontact_additional} provides additional foot-contact visualizations across Booster T1 and Unitree G1 variants. Consistent with the main paper, GaitSpan produces increasingly dense contact--non-contact transitions as commanded speed increases, indicating a smooth progression from walking toward more dynamic gait regimes. The results also reveal embodiment-specific differences even within the same robot family. For example, Unitree G1 with 23 DoF and 29 DoF exhibits different contact timing and transition patterns under the same speed commands. The additional DoFs in the 29-DoF configuration are located in the waist and arms, providing more freedom for whole-body coordination rather than forcing adaptation to be expressed primarily through the lower body. Interestingly, this does not necessarily lead to larger visible motion amplitudes. Instead, the policy can distribute adjustments across more joints, resulting in smoother and more subtle whole-body compensation, and consequently different foot-contact organization compared with the more leg-dominant 23-DoF configuration.

\begin{figure}[t]
\centering\small
\newcommand{\squareimage}[1]{\includegraphics[width=\textwidth, height=0.34\textwidth]{#1}}

\resizebox{\textwidth}{!}{
\begin{tabular}{@{}l@{}}
\squareimage{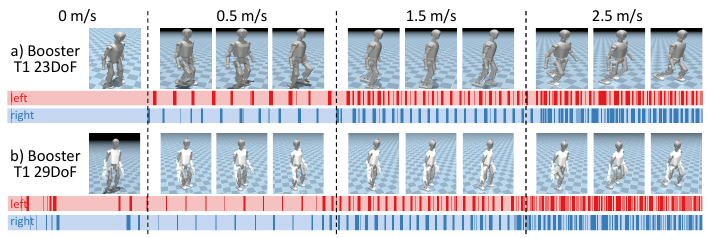} \\
\squareimage{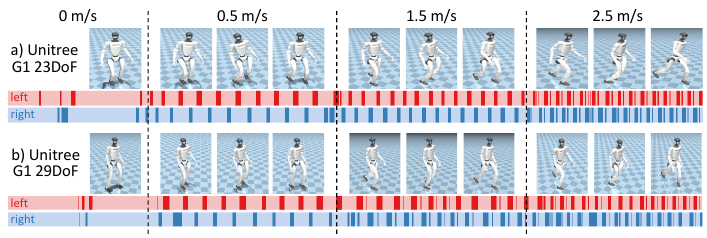}
\end{tabular}
}
\caption{\textbf{Additional foot contact plots, comparing T1/G1 23DoF vs. 29DoF}. We list the 23 DoF results again here for better comparison. }
\label{fig:footcontact_additional}
\end{figure}


\begin{figure}[h]
\newcommand{\close}{\hspace{1pt}}
\centering\small
\begin{minipage}[t]{0.59\linewidth}
\newcommand{\squareimage}[1]{\includegraphics[width=0.49\textwidth, height=0.49\textwidth]{#1}}
\vspace{0pt}
\centering\small
\resizebox{\textwidth}{!}{
\begin{tabular}{@{}c@{\close}c@{}}
\squareimage{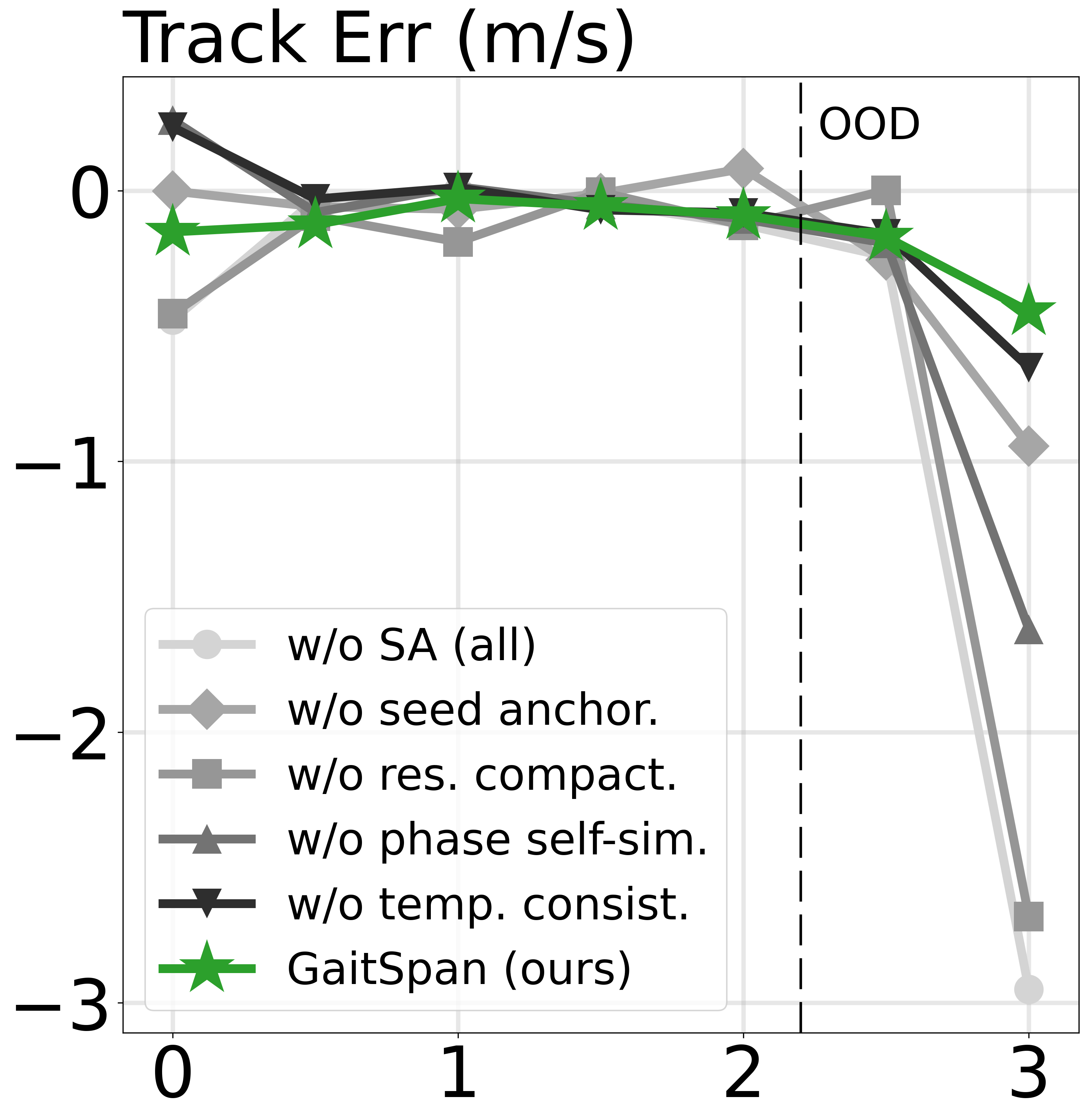} &
\squareimage{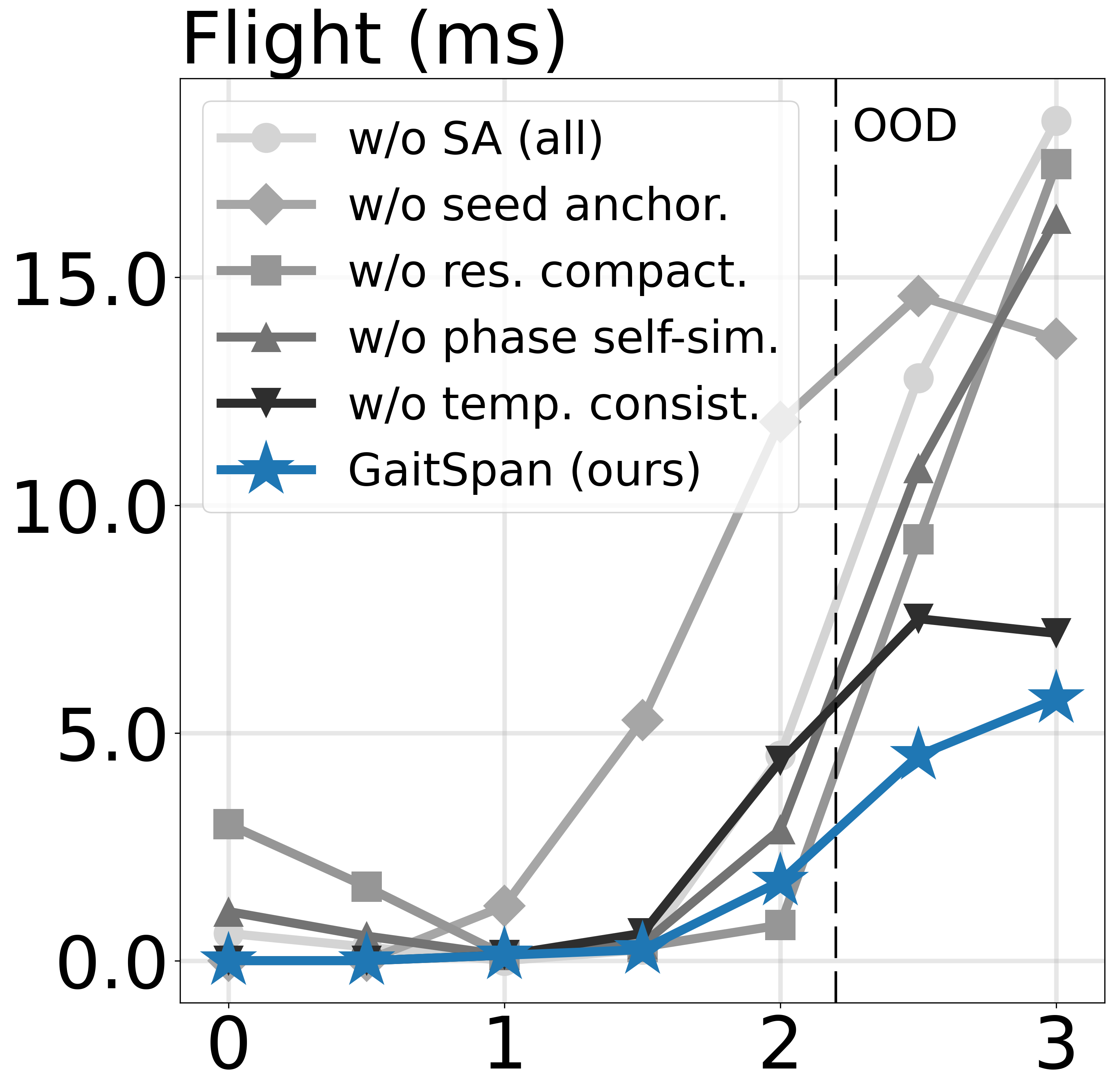}
\end{tabular}
}
\end{minipage}
\hfill
\begin{minipage}[t]{0.39\linewidth}
\newcommand{\squareimage}[1]{\includegraphics[width=0.5\textwidth, height=0.5\textwidth]{#1}}
\vspace{0pt}
\centering\small
\resizebox{\textwidth}{!}{
\begin{tabular}{@{}c@{\close}c@{\close}c@{}}
\squareimage{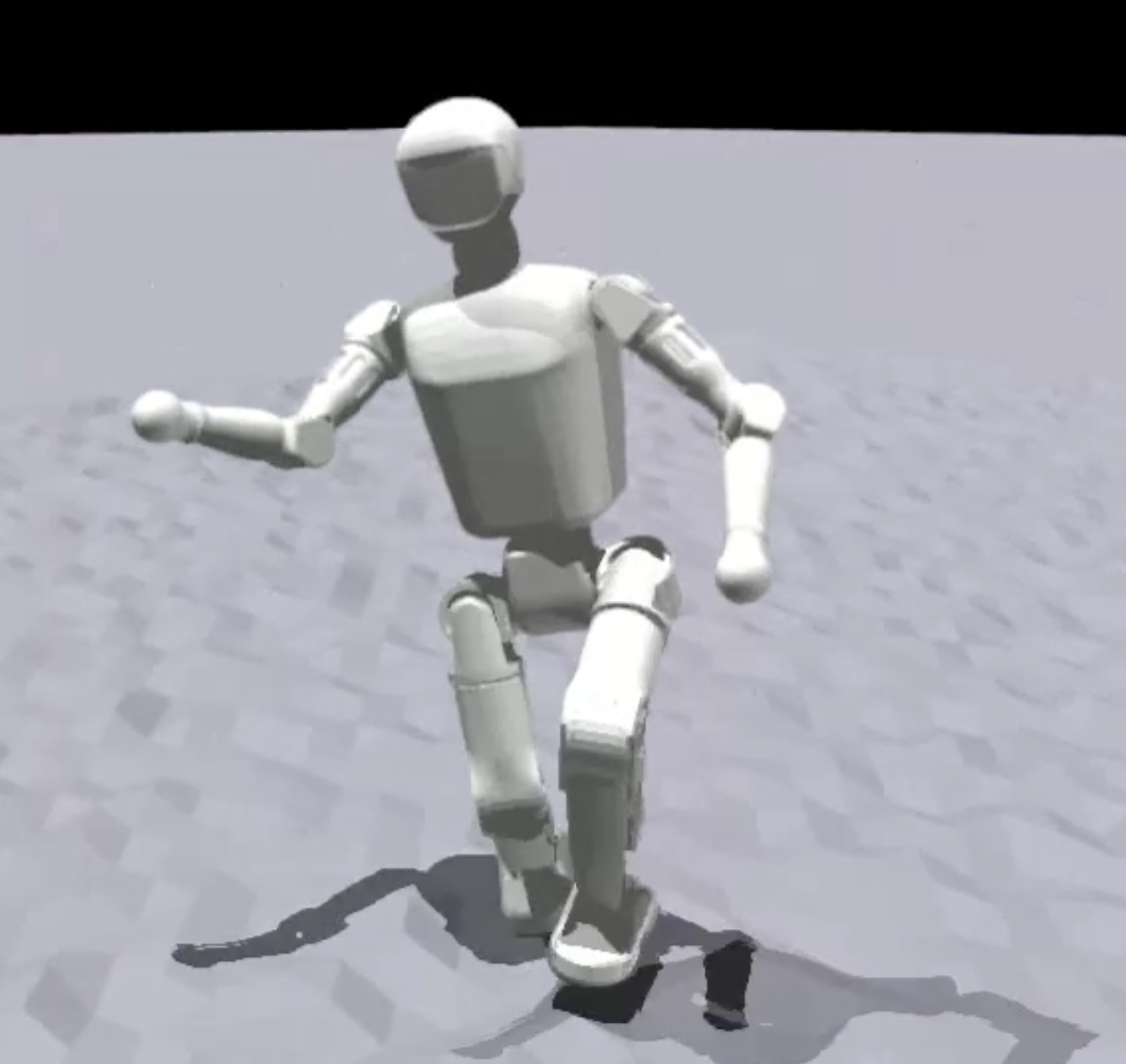} &
\squareimage{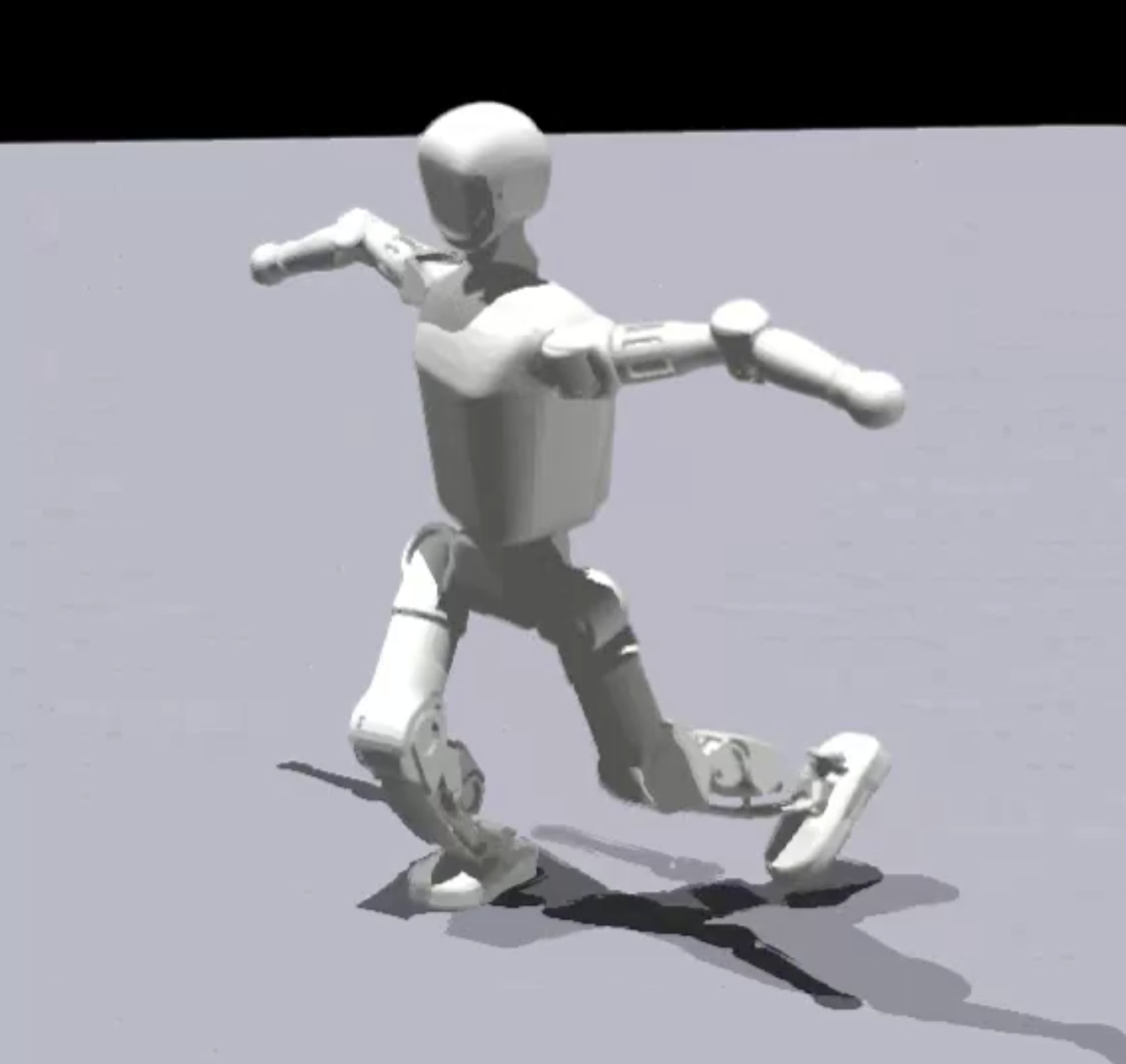} &
\squareimage{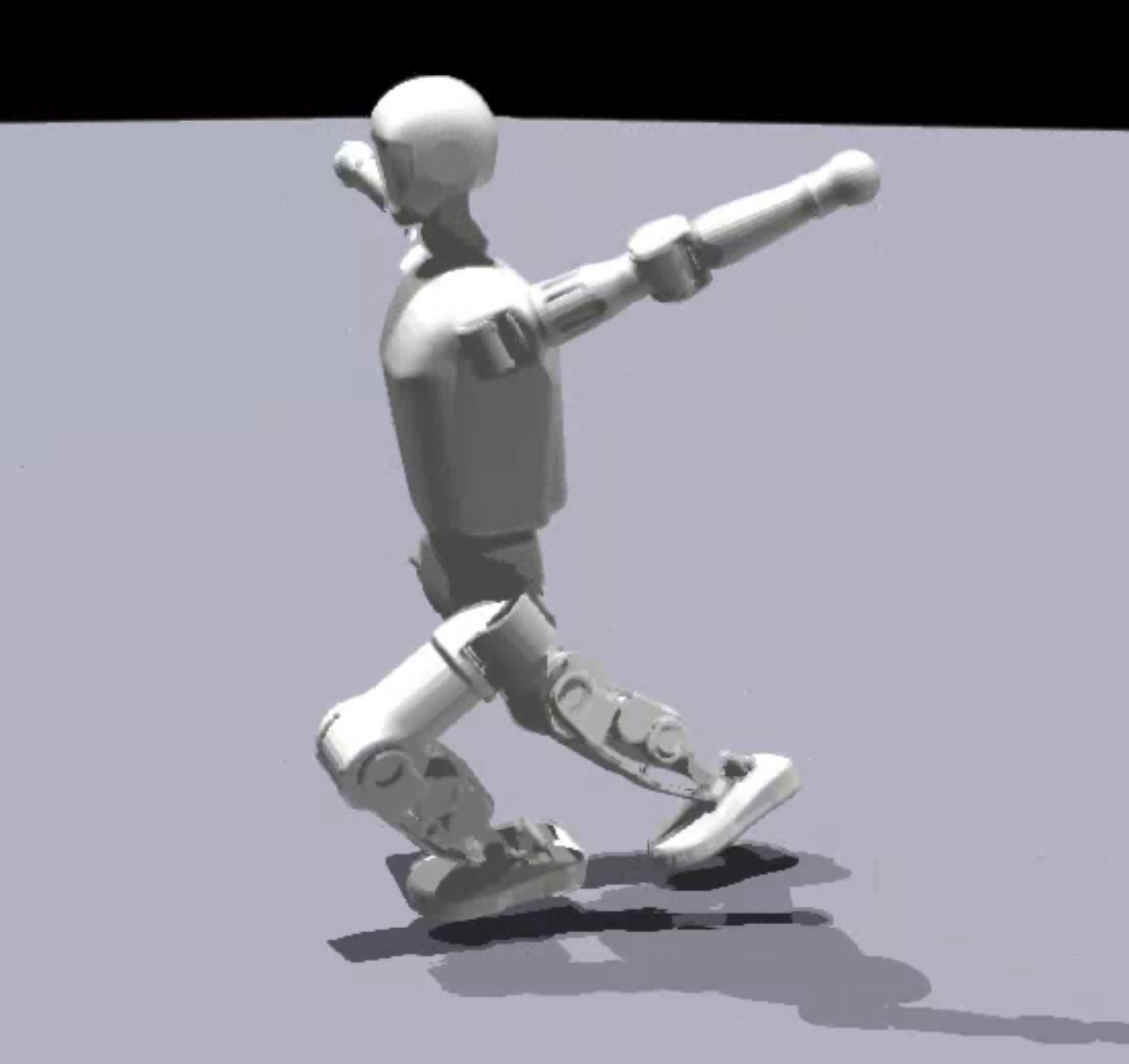} \\
\multicolumn{3}{@{}l}{a) \textit{w/o seed anchor.} at 0.5, 1.5, 2.5 m/s} \\ [2pt]
\squareimage{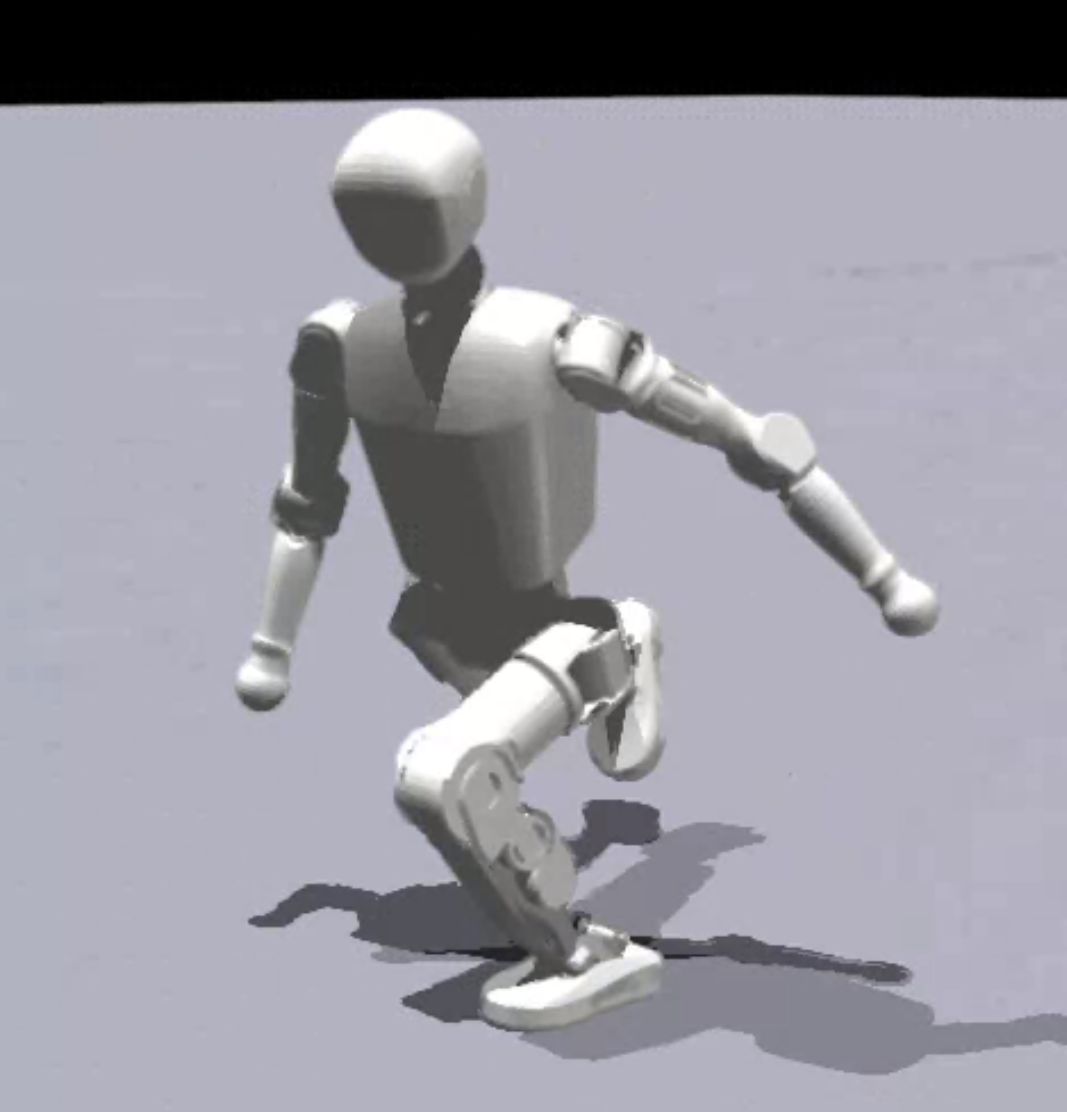} &
\squareimage{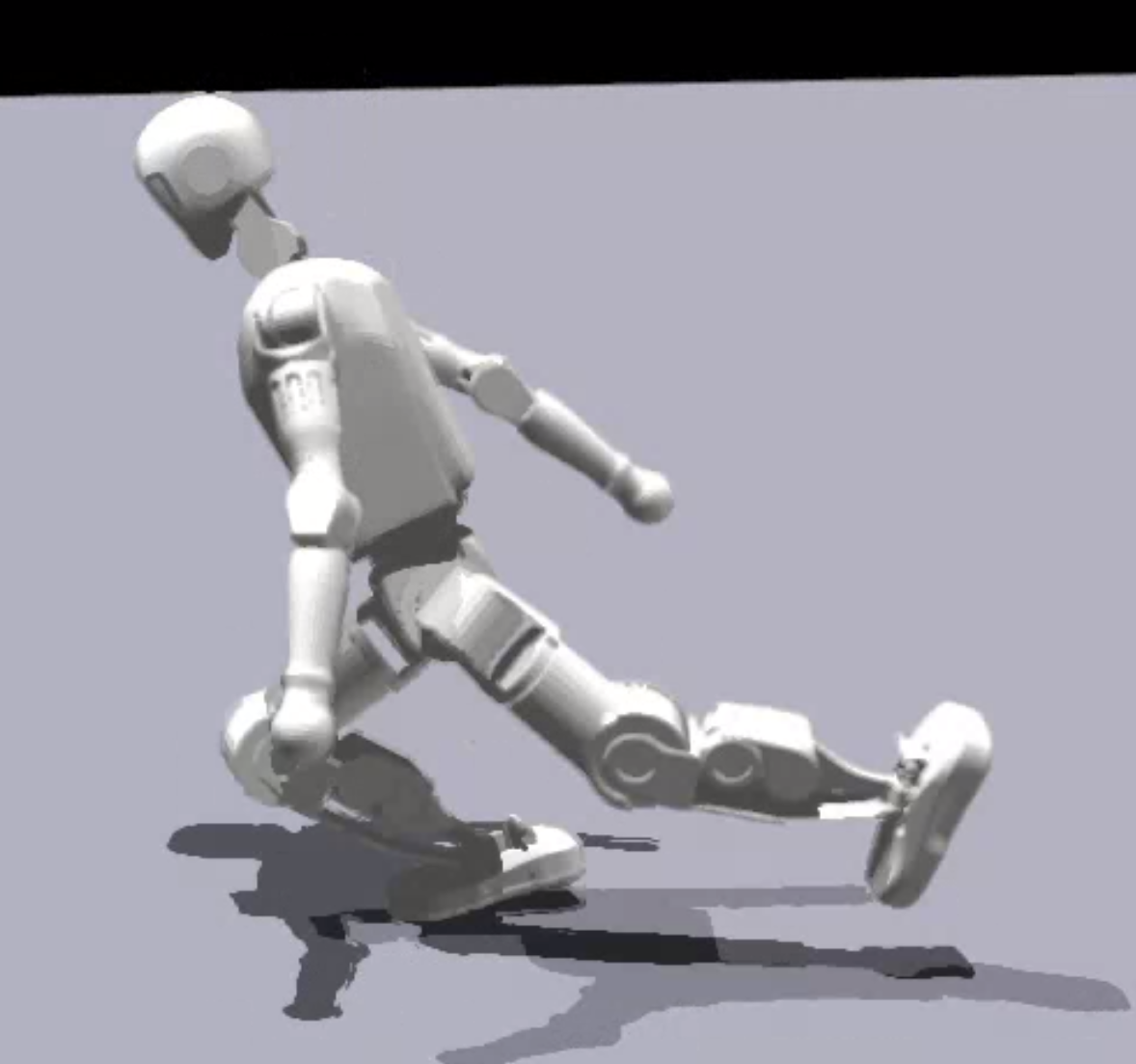} &
\squareimage{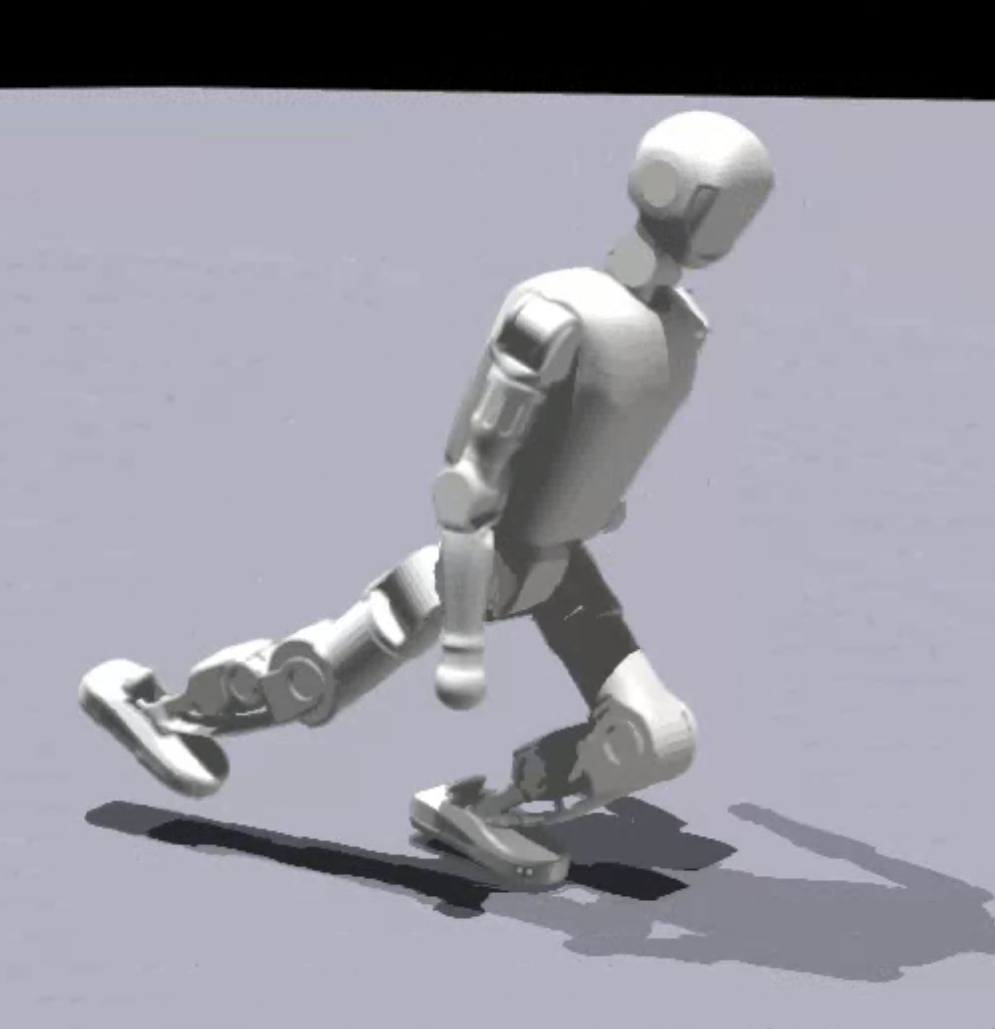} \\
\multicolumn{3}{@{}l}{b) w/o one of each three remaining SA components at 2.5 m/s}
\end{tabular}
}
\end{minipage}
\caption{\textbf{Ablation on self-anchored regularization.} Removing self-anchored terms preserves reasonable leg-level gait structure due to GaitWave and H-SLIP, but leads to unstable upper-body motion and less controlled high-speed behavior. The full model better balances tracking, flight emergence, and whole-body stability.}
\label{fig:sa_ablation}

\end{figure}

\section{Additional Implementation Details}
\label{appendix:implementation}

\subsection{Additional Method Details}
\label{appendix:method}

\subsubsection{Self-Anchored (SA) Regularization}
\label{sec:regularization}

Although GaitWave and the residual branch provide expressiveness, unconstrained expansion can cause the learned controller to disregard the foundational walking skill. We introduce self-anchored regularization to ensure that deviations from walking are deliberate and command-driven.

\noindent\textbf{Seed anchoring.}
Let
$    g^{\mathrm{seed}}_t
    =
    1 -
    g^{\mathrm{dyn}}_t
$
denote a soft walking-regime gate. At low commanded speeds, we encourage the final action to remain close to the nominal canonical walking action:
\begin{equation}
    \mathcal{L}_{\mathrm{anchor},t}
    =
    g^{\mathrm{seed}}_t
    \left\|
        \mathbf{a}_t
        -
        \mathbf{a}^{\mathrm{seed}}_t
    \right\|_2^2.
    \label{eq:anchor_loss}
\end{equation}
This regularization preserves canonical stability in the region where walking is sufficient, while smoothly relaxing the constraint as dynamic motion becomes necessary.

\noindent\textbf{Residual compactness.} We discourage unnecessarily large residual corrections through
\begin{equation}
    \mathcal{L}_{\mathrm{res},t}
    =
    \left\|
        \mathbf{a}^{\mathrm{res}}_t
    \right\|_2^2.
    \label{eq:residual_compactness}
\end{equation}
This term biases the learned controller toward explaining motion through the expanded canonical structure whenever possible, and using the residual branch only for adaptations not captured by GaitWave.

\noindent\textbf{Phase-Aware self-similarity.} To preserve coherent locomotion structure across repeated phase states, we extract a compact proprioceptive feature
\begin{equation}
    \mathbf{f}_t
    =
    \varphi(\mathbf{s}_t),
\end{equation}
where $\varphi(\cdot)$ includes joint configurations, joint velocities, gravity-aligned orientation, and optionally contact features. We discretize the locomotion phase into $N_{\phi}$ bins and maintain an exponential-moving-average prototype
$\bar{\mathbf{f}}_j$ for each phase bin $j$. For
$j_t = \operatorname{bin}(\phi_t)$, the prototype is updated as
\begin{equation}
    \bar{\mathbf{f}}_{j_t}
    \leftarrow
    (1-\mu)
    \bar{\mathbf{f}}_{j_t}
    +
    \mu
    \mathbf{f}_t,
\end{equation}
without gradient propagation through the prototype. The self-similarity loss is
\begin{equation}
    \mathcal{L}_{\mathrm{ss},t}
    =
    \left\|
        \mathbf{f}_t
        -
        \bar{\mathbf{f}}_{j_t}
    \right\|_2^2.
    \label{eq:self_similarity_loss}
\end{equation}
This term encourages recurrent locomotion structure without enforcing a predefined gait trajectory or labeled phase schedule.

\noindent\textbf{Temporal action consistency.} Finally, we encourage temporal smoothness relative to the canonical action evolution. Let
\begin{equation}
    \operatorname{sim}(\mathbf{x},\mathbf{y})
    =
    \frac{
        \mathbf{x}^{\top}\mathbf{y}
    }{
        \|\mathbf{x}\|_2
        \|\mathbf{y}\|_2
        + \epsilon
    }.
\end{equation}
We define
\begin{equation}
\begin{aligned}
    \mathcal{L}_{\mathrm{temp},t}
    =
    \operatorname{ReLU}
    \Big(
        &
        \operatorname{sim}
        \left(
            \mathbf{a}^{\mathrm{seed}}_t,
            \mathbf{a}^{\mathrm{seed}}_{t-1}
        \right)
        \\
        &-
        \operatorname{sim}
        \left(
            \mathbf{a}_t,
            \mathbf{a}_{t-1}
        \right)
    \Big).
    \label{eq:temporal_loss}
\end{aligned}
\end{equation}
This term prevents the expanded policy from becoming less temporally coherent than the foundational skill from which it grows.

\subsubsection{Ablation on SA}
\label{sec:suppl_objective}

We further ablate the auxiliary self-anchored regularization terms used in GaitSpan. As shown in Fig.~\ref{fig:sa_ablation}, removing all self-anchored regularization still produces reasonable leg motion, suggesting that the main GaitWave and H-SLIP designs already provide the core structure for gait emergence. However, the resulting motions show unstable upper-body behavior and exaggerated posture changes, which may increase risk during real-world deployment. This indicates that self-anchored regularization is not the primary source of gait growth, but serves as an important stabilizer for preserving whole-body coherence.

The individual ablations further show that different regularizers play complementary roles. Removing seed anchoring causes the policy to deviate more aggressively from the walking prior, leading to larger flight time but less controlled motion. Removing residual compactness, phase self-similarity, or temporal consistency also degrades the balance between tracking and dynamic gait emergence, especially under high-speed and OOD commands. In contrast, the full GaitSpan model maintains low tracking error while producing moderate, command-dependent flight, showing that self-anchored regularization helps the policy expand the seed skill without overwriting its stabilizing structure.


\subsection{Action Space}
At each time step, the policy receives a concatenated observation vector $s_t$ that combines proprioception, velocity command, and short-term history:
1) base angular velocity $\omega_t \in \mathbb{R}^3$,
2) gravity direction projected to the body frame $g_t \in \mathbb{R}^3$ (a compact representation of base orientation),
3) commanded velocities,
4) joint positions $q_t \in \mathbb{R}^{n_{\text{dof}}}$,
5) joint velocities $\dot q_t \in \mathbb{R}^{n_{\text{dof}}}$,
6) the previous action $a_{t-1}\in\mathbb{R}^{n_{\text{dof}}}$ (we use $1$ frame in practice),
and 7) a periodic phase encoding $(\sin\varphi_t,\cos\varphi_t)$ to facilitate learning of cyclic locomotion structure and to support the rhythm expansion described in the main paper.

Following common practice in humanoid RL, these target angles are converted into joint torques using a joint-space PD controller with zero target joint velocity:
\begin{equation}
\tau_t = K_p(\hat{q}_t - q_t) - K_d \dot{q}_t,
\label{eq:pd_control}
\end{equation}
where $K_p$ and $K_d$ are joint-wise gains. 

\subsection{General Training Setup}

We train with FastSAC in Isaac Gym (200\,Hz physics, 50\,Hz control) using 4096 parallel environments, 20\,s episodes, 10\,s command resampling, and a 0.2 standing-command probability. We partition the command speed range into three parallel resolutions (2, 4, and 8 bins) and use temperature-scaled softmax blending ($\tau = 0.08$).
An anchor regularizer keeps learned weights near $\boldsymbol{\beta} = (1.0, 0.25, 0.10, 0.05)$ over scales $(1.0, 1.25, 1.5, 2.0)$, with the stronger effect at lower velocity. High-speed exposure increases via a sampling bias whose exponent ramps linearly from 1.0 to 2.6 over the first 95\% of training, concentrating mass near 2.2\,m/s. Learning rates are $3 \times 10^{-4}$ (actor, critic, entropy), with buffer size 1024, batch size 8192, 8 updates per step, policy updates every 4 steps, $\gamma = 0.97$, Polyak $\tau = 0.125$, a 101-atom distributional critic on $[-20, 20]$, hidden sizes 512/768, symmetry augmentation, automatic entropy tuning, and bf16 training.

Complete \textbf{RL reward terms} and \textbf{domain randomizations} are provided in Tab.~\ref{tab:reward_terms} and Tab.~\ref{tab:randomization}.

\begin{table}[h]
\centering
\footnotesize
\renewcommand{\arraystretch}{1.0}
\caption{\textbf{General reward terms and weights.}
Per-step contribution is $r_t = \sum_i w_i\,\Delta t\, r_i(\cdot)$.}
\label{tab:reward_terms}
\begin{tabular}{@{}l@{\hspace{6pt}}l@{\hspace{6pt}}r@{}}
\toprule
\textbf{Reward} & \textbf{Definition} & \textbf{Weight} \\
\midrule
\multicolumn{3}{@{}l}{\textit{Command Tracking}} \\
\quad Tracking lin.\ vel.
  & $\exp(-\|\mathbf{v}_{xy}^{\mathrm{cmd}}-\mathbf{v}_{xy}\|^2/\sigma)$
  & 3.0 \\
\quad Tracking ang.\ vel.
  & $\exp(-(\omega_z^{\mathrm{cmd}}-\omega_z)^2/\sigma)$
  & 0.15 \\
\midrule
\multicolumn{3}{@{}l}{\textit{Gait \& Running}} \\
\quad Feet phase
  & $\exp(-\sum_{f\in\{L,R\}} (z_f-\hat{z}_f(\phi_f))^2/\sigma)$
  & 1.2 \\
\quad Flight time
  & $\mathrm{clamp}(t_f/t_{\max},0,1)$
  & 3.6 \\
\quad Global SLIP
  & defined in Sec. 3
  & 4.8 \\
\quad Local SLIP-Up
  & defined in Sec. 3
  & 3.6 \\
\quad Local SLIP-Low
  & defined in Sec. 3
  & 3.6 \\
\midrule
\multicolumn{3}{@{}l}{\textit{Posture \& Stability}} \\
\quad Penalty ang.\ vel.\ (pitch)
  & $\omega_y^2$
  & $-0.3$ \\
\quad Penalty orientation
  & $\|\mathbf{g}_{xy}^{\mathrm{proj}}\|^2$
  & $-0.1$ \\
\quad Penalty base height
  & $(h_{\mathrm{base}}-h^*)^2$
  & $-1.0$ \\
\quad Penalty original pose
  & $\sum_j w_j^{\mathrm{dof}}(q_j-q_j^{\mathrm{def}})^2$
  & $-0.05$ \\
\midrule
\multicolumn{3}{@{}l}{\textit{Regularization}} \\
\quad Penalty action rate
  & $\|\mathbf{a}_t-\mathbf{a}_{t-1}\|^2$
  & $-0.2$ \\
\quad Penalty close feet
  & $\mathbb{1}[d_{\perp}<d^*]$
  & $-1.0$ \\
\quad Penalty feet orientation
  & $\sum_{f\in\{L,R\}} \|\mathbf{g}_{xy,f}\|$
  & $-0.5$ \\
\quad Penalty ankle roll action
  & $\sum_{f\in\{L,R\}}|a_{ankle\_roll, f}|$
  & $-0.1$ \\
\midrule
\multicolumn{3}{@{}l}{\textit{Survival}} \\
\quad Alive
  & $1$
  & 0.2 \\
\bottomrule
\end{tabular}
\end{table}

\begin{table}[h]
\centering
\renewcommand{\arraystretch}{1.0}
\caption{\textbf{Domain randomization terms} used for all settings and embodiments.}
\label{tab:randomization}
\begin{tabular}{@{}l@{\hspace{6pt}}l@{}}
\toprule
\textbf{Randomization} & \textbf{Range} \\
\midrule
\multicolumn{2}{@{}l}{\textit{Push Disturbances}} \\
\quad Push interval (s)             & $[12,\ 20]$ \\
\quad Push $|v_{xy}|$ cap (m/s)     & $0.22$ \\
\midrule
\multicolumn{2}{@{}l}{\textit{Control \& Actuation}} \\
\quad Action delay (steps)          & $[0,\ 1]$ \\
\quad DOF default bias (rad)        & $[-0.004,\ 0.004]$ \\
\quad $k_P$ / $k_D$ scale             & $[0.9,\ 1.1]$ \\
\quad Torque RFI                    & off \\
\midrule
\multicolumn{2}{@{}l}{\textit{Physical Properties}} \\
\quad Link mass scale               & $[0.93,\ 1.07]$ \\
\quad Torso mass add (kg)           & $[-0.35,\ 1.5]$ \\
\quad Friction (abs)                & $[0.65,\ 1.1]$ \\
\quad Base CoM offset (m, xyz)      & $\pm 0.008$ \\
\midrule
\multicolumn{2}{@{}l}{\textit{Reset Initialization}} \\
\quad Reset joint pos scale         & $[0.94,\ 1.06]$ \\
\quad Reset joint vel (rad/s)       & $[-0.08,\ 0.08]$ \\
\bottomrule
\end{tabular}
\end{table}

\subsection{Robot-Specific Configurations}

We provide robot-specific configurations in Tab.~\ref{tab:robot_specific} and pose reward weights for every joint in Tab.~\ref{tab:pose_weights}.

\begin{table}[h]
\centering
\footnotesize
\renewcommand{\arraystretch}{1.0}
\caption{\textbf{Robot-specific configurations.}}
\label{tab:robot_specific}
\begin{tabular}{@{}lccccc@{}}
\toprule
\textbf{Term} & \textbf{Booster T1} & \textbf{Booster K1} & \textbf{Unitree G1} \\
\midrule
Desired base height $h^*$
  & 0.68\,m & 0.58\,m & 0.75\,m\\
Close feet threshold $d^*$
  & 0.15\,m & 0.15\,m & 0.20\,m \\
Training iterations
  & 100K & 100K & 50K \\
\bottomrule
\end{tabular}
\end{table}

\begin{table}[h]
\centering
\footnotesize
\renewcommand{\arraystretch}{1.0}
\caption{\textbf{Per-joint pose penalty weights $w_j^{\mathrm{dof}}$.}
``--'': joint not actuated.}
\label{tab:pose_weights}
\begin{tabular}{@{}l@{\hspace{4pt}}c@{\hspace{6pt}}c@{\hspace{6pt}}c@{\hspace{6pt}}c@{\hspace{6pt}}c@{}}
\toprule
\textbf{Joint} & \textbf{T1-23DoF} & \textbf{T1-29DoF} & \textbf{K1-22DoF} & \textbf{G1-23DoF} & \textbf{G1-29DoF} \\
\midrule
\multicolumn{6}{@{}l}{\textit{Head}} \\
\quad Head yaw
  & 50.0 & 50.0 & 50.0 & -- & -- \\
\quad Head pitch
  & 50.0 & 50.0 & 50.0 & -- & -- \\
\midrule
\multicolumn{6}{@{}l}{\textit{Left arm}} \\
\quad Shoulder pitch
  & 0.04 & 0.04 & 50.0 & 50.0 & 50.0 \\
\quad Shoulder roll
  & 0.12 & 0.12 & 50.0 & 50.0 & 50.0 \\
\quad Shoulder yaw
  & -- & -- & -- & 50.0 & 50.0 \\
\quad Elbow pitch
  & 0.06 & 0.06 & 50.0 & -- & -- \\
\quad Elbow yaw
  & 0.22 & 0.22 & 50.0 & -- & -- \\
\quad Elbow
  & -- & -- & -- & 50.0 & 50.0 \\
\quad Wrist pitch
  & -- & 50.0 & -- & -- & 50.0 \\
\quad Wrist yaw
  & -- & 50.0 & -- & -- & 50.0 \\
\quad Hand/wrist roll
  & -- & 50.0 & -- & 50.0 & 50.0 \\
\midrule
\multicolumn{6}{@{}l}{\textit{Right arm}} \\
\quad Shoulder pitch
  & 0.04 & 0.04 & 50.0 & 50.0 & 50.0 \\
\quad Shoulder roll
  & 0.12 & 0.12 & 50.0 & 50.0 & 50.0 \\
\quad Shoulder yaw
  & -- & -- & -- & 50.0 & 50.0 \\
\quad Elbow pitch
  & 0.06 & 0.06 & 50.0 & -- & -- \\
\quad Elbow yaw
  & 0.22 & 0.22 & 50.0 & -- & -- \\
\quad Elbow
  & -- & -- & -- & 50.0 & 50.0 \\
\quad Wrist pitch
  & -- & 50.0 & -- & -- & 50.0 \\
\quad Wrist yaw
  & -- & 50.0 & -- & -- & 50.0 \\
\quad Hand/wrist roll
  & -- & 50.0 & -- & 50.0 & 50.0 \\
\midrule
\multicolumn{6}{@{}l}{\textit{Waist}} \\
\quad Waist yaw
  & 5.0 & 5.0 & -- & 50.0 & 50.0 \\
\quad Waist roll
  & -- & -- & -- & -- & 50.0 \\
\quad Waist pitch
  & -- & -- & -- & -- & 50.0 \\
\midrule
\multicolumn{6}{@{}l}{\textit{Left leg}} \\
\quad Hip pitch
  & 0.01 & 0.01 & 0.01 & 0.01 & 0.01 \\
\quad Hip roll
  & 1.0 & 1.0 & 1.0 & 1.0 & 1.0 \\
\quad Hip yaw
  & 5.0 & 5.0 & 5.0 & 5.0 & 5.0 \\
\quad Knee
  & 0.01 & 0.01 & 0.01 & 0.01 & 0.01 \\
\quad Ankle pitch
  & 5.0 & 5.0 & 5.0 & 5.0 & 5.0 \\
\quad Ankle roll
  & 5.0 & 5.0 & 5.0 & 5.0 & 5.0 \\
\midrule
\multicolumn{6}{@{}l}{\textit{Right leg}} \\
\quad Hip pitch
  & 0.01 & 0.01 & 0.01 & 0.01 & 0.01 \\
\quad Hip roll
  & 1.0 & 1.0 & 1.0 & 1.0 & 1.0 \\
\quad Hip yaw
  & 5.0 & 5.0 & 5.0 & 5.0 & 5.0 \\
\quad Knee
  & 0.01 & 0.01 & 0.01 & 0.01 & 0.01 \\
\quad Ankle pitch
  & 5.0 & 5.0 & 5.0 & 5.0 & 5.0 \\
\quad Ankle roll
  & 5.0 & 5.0 & 5.0 & 5.0 & 5.0 \\
\bottomrule
\end{tabular}
\end{table}

\end{document}